\documentclass[10pt]{article} 
\usepackage[preprint]{tmlr}

\usepackage{amsmath,amsfonts,bm}

\def\eqref#1{equation~\ref{#1}}

\def\1{\bm{1}}

\DeclareMathAlphabet{\mathsfit}{\encodingdefault}{\sfdefault}{m}{sl}
\SetMathAlphabet{\mathsfit}{bold}{\encodingdefault}{\sfdefault}{bx}{n}

\newcommand{\E}{\mathbb{E}}

\newcommand{\R}{\mathbb{R}}

\newcommand{\Var}{\mathrm{Var}}

\usepackage{hyperref}
\usepackage{url}

\usepackage[rightcaption]{sidecap}
\usepackage{graphicx} 
\usepackage{enumitem}
\usepackage{overpic} 
\usepackage{caption}
\usepackage{tcolorbox}

\usepackage{amsmath}
\usepackage{amssymb}
\usepackage{amsthm}

\usepackage{euscript}
\usepackage{booktabs}

\newcommand{\be}{\begin{equation}}
\newcommand{\ee}{\end{equation}}
\newcommand{\benn}{\begin{equation*}}
\newcommand{\eenn}{\end{equation*}}
\newcommand{\bal}{\begin{align}}
\newcommand{\eal}{\end{align}}
\newcommand{\balnn}{\begin{align*}}
\newcommand{\ealnn}{\end{align*}}

\newtheorem{theorem}{Theorem}[section]
\newtheorem{corollary}[theorem]{Corollary}
\newtheorem{proposition}[theorem]{Proposition}
\newtheorem{lemma}[theorem]{Lemma}

\newtheorem{definition}[theorem]{Definition}

\newcommand{\N}{{\mathbb{N}}} 

\renewcommand{\vec}[1]{\mathbf{#1}} 
\newcommand{\Id}{{\mathbb{I}}} 

\newcommand*{\boldone}{\text{\usefont{U}{bbold}{m}{n}1}}

\title{Strictly proper scoring rules for node selection tasks on graphs}

\author{\name Lorenz~Hilfiker \email lorenz.hilfiker@fhnw.ch \\
    \addr University of Applied Sciences and Arts Northwestern Switzerland FHNW
    \AND
    \name Nathan~Cincera~Brack \email nathan.brack@fhnw.ch\\
    \addr University of Applied Sciences and Arts Northwestern Switzerland FHNW \\
    University of Zurich
    \AND
    \name Martin~Sterchi \email martin.sterchi@fhnw.ch\\
    \addr University of Applied Sciences and Arts Northwestern Switzerland FHNW \\
    University of Zurich
}

\def\month{MM}  
\def\year{YYYY} 
\def\openreview{\url{https://openreview.net/forum?id=XXXX}} 

\begin{document}

\maketitle

\begin{abstract}
    In node selection tasks, the goal is to identify a single unknown target node in a graph. We consider probabilistic supervised node selection, where the estimator outputs a probability distribution over the nodes of the graph and the ground-truth node is available for evaluation. Classical evaluation metrics for such tasks are either statistically ill-conceived or blind to the graph structure. To fill this gap, we introduce two strictly proper scoring rules that are explicitly graph-aware: the Resistance score, which is parameter-free, and the Diffusion score, which operates at an adjustable coarse-graining resolution. We show that the Diffusion score interpolates between the multi-category Brier score and a cluster-level weighted Brier score. The Resistance score generalises two classical scoring rules: the multi-category Brier score and the Ranked probability score. Applied to rooted trees, it also serves as the strictly proper version of the improper Hierarchical error distance used in hierarchical classification. In the same setting, we furthermore establish strict propriety of the Hierarchical cross-entropy family, a recently introduced separate class of hierarchy-aware scoring rules. Experiments on epidemic source detection and on ImageNet classification confirm that the proposed scores detect meaningful differences in estimation quality that classical metrics miss.
\end{abstract}

\section{Introduction}
\label{sec:introduction}

Graph learning encompasses a range of tasks, from supervised problems such as graph classification, node classification, or link prediction to unsupervised problems such as community detection. In this paper, we focus on the supervised task of \emph{(single) node selection}. Whereas node classification assigns class labels to nodes, potentially assigning the same label to multiple nodes, node selection consists of identifying a \emph{single} node as the output.

Many real-world applications take this form: where did an epidemic or rumour start? Which node should we target to maximise influence? Which node has failed? In other words: node selection subsumes important graph learning problems such as source detection for spreading processes like epidemics or rumours \citep{Shah:2011}, optimisation problems such as single-seed influence maximisation \citep{Kempe:2003}, or fault detection in communication networks \citep{Ma:2014}. Moreover, any edge selection task can be recast as a node selection task on the corresponding line graph. Finally, an important special case is node selection on rooted trees, which naturally captures \emph{hierarchical classification} \citep{Silla:2011}, where class labels are organised according to a hierarchical taxonomy. Applications range from computer vision \citep{Deng:2009} to natural language processing \citep{Kosmopoulos:2015} and bio-informatics \citep{Radivojac:2013}.

Most state-of-the-art approaches to node selection tasks are probabilistic (e.g., Monte Carlo methods or graph neural networks). Rather than directly predicting a single node, these methods output a probability distribution over the entire set of nodes. This raises the question of how such probabilistic estimates should be evaluated against the ground truth.

Despite the prevalence of probabilistic node selection methods, there is no consensus on how to evaluate them. Most papers report rank-based metrics such as top-$K$ accuracy or the rank of the true node (e.g., \cite{Lokhov:2014, Altarelli:2014, Antulov:2015, Paluch:2018, Sha:2021}). However, these metrics are unsatisfactory for at least two reasons: \textbf{(a)}~they do not take into account the graph structure when assessing how closely an estimate matches the observed true node, and \textbf{(b)}~they do not adequately penalise overconfident, yet incorrect, estimates. Existing alternatives such as the Error distance \citep{Antulov:2015,Jiang:2017,Paluch:2018} or the classical multi-category Brier score \citep{Brier:1950} address either \textbf{(a)} or \textbf{(b)}, but neither of them resolves both points simultaneously.

In this paper, we close this gap by proposing evaluation metrics that are both \emph{graph-aware} -- rewarding estimates that place high probability near the true node in a graph sense -- and \emph{strictly proper} -- so that the score is uniquely minimised when the estimated distribution matches the true one. We achieve this by bridging ideas from statistics, in particular the concept of \emph{kernel scores} \citep{Gneiting:2007}, with ideas from graph learning \citep{Smola:2003, Fouss:2007} and network science \citep{Devriendt:2022,Villegas:2023}.

In the special case of hierarchical classification, hierarchy-aware evaluation metrics have received considerable attention \citep{Silla:2011, Kosmopoulos:2015, Zangari:2024}. Threshold-based set-overlap metrics such as Hierarchical precision, recall and F-score \citep{Kiritchenko:2005} and mode-based metrics such as Hierarchical error distance \citep{Russakovsky:2015} are widely used, but they are not proper scores. The only exception known to us is the concurrent and independent work of \citet{Urbani:2024}, who prove propriety (but not strict propriety) of a family of log-score-type scoring rules  for node selection on rooted trees. We extend their result to strict propriety and provide a complementary Brier-type family of scoring rules applicable to general graphs.

Our main contributions are:
\begin{itemize}
\item We provide the first systematic treatment of strictly proper scoring rules for node selection tasks on general graphs that reward estimates based on their proximity to the true node in the graph. We show that existing graph-aware metrics (Error distance, Hierarchical error distance and their variations) are improper.
\item We introduce two strictly proper, graph-aware scoring rules for probabilistic node selection: the \emph{Resistance score} (parameter-free, based on effective resistance distance) and the \emph{Diffusion score} (one-parameter family, based on the diffusion kernel).
\item We show that the Resistance score generalises the multi-category Brier score (recovered on the complete graph) and the Ranked probability score (recovered on the path graph), and that, specialised to rooted trees, it is the natural way to turn the improper Hierarchical error distance into a proper score.
\end{itemize}

The paper is structured as follows. Section~\ref{sec:problem_statement} formally states the problem, Section~\ref{sec:background} provides the necessary statistical background on proper scoring rules, and Section~\ref{sec:related_work} discusses related work. In Section~\ref{sec:scores}, we introduce the Resistance score and the Diffusion score. We discuss their properties, show how they relate to classical proper scoring rules, and illustrate their behaviour on a toy example. Section~\ref{subsec:hierarchical_classification} discusses the Resistance score in the context of hierarchical classification, where it provides a strictly proper alternative to widely used but improper metrics such as Hierarchical error distance and distance@$K$. Efficient computation of our scores on large graphs is briefly discussed in Section~\ref{subsec:efficient_computation}. Section~\ref{sec:source_detection} illustrates the scores on a canonical source detection task. Section~\ref{sec:imagenet} presents an empirical evaluation in the standard hierarchical classification setting on ImageNet. We conclude with a discussion of extensions and limitations in Section~\ref{sec:discussion}.

\section{Problem Statement}
\label{sec:problem_statement}

Let $G = (V, E)$ be a simple graph with node set $V$ and edge set $E$, and let $N = |V|$. A \emph{node selection task} on $G$ consists of identifying a single unknown node $i \in V$. A probabilistic method for such a task outputs a probability distribution over $V$, i.e., an element in the $(N-1)$-dimensional probability simplex
\be
    \Delta^{N-1} = \left\lbrace \vec{p}\in\R^N \;\middle\vert\; \sum_{j'=1}^{N} p_{j'} = 1, \; p_j\geq 0 \;\forall j=1,\dots, N \right\rbrace \, ,
\ee
whose component $p_j$ denotes the estimated probability that node $j$ is the true node. This setting subsumes a variety of problems in graph learning and network science. We mention two canonical instances that will serve as examples in the paper:

\paragraph{Source detection.}
A spreading process (an epidemic, a rumour, a cascade) originates at a single unknown source node $i \in V$ and propagates across $G$. Given some information about the progression of the spread at some later time, the task is to identify its origin $i$. Many methods for this type of inverse problem are probabilistic, issuing a distribution $\vec{p}$ over $V$. The graph structure matters when evaluating this task: it is arguably better to miss the true source by a single hop than estimating a node far away from the true source, so a fair evaluation of $\vec{p}$ should take $G$ into account.

\paragraph{Hierarchical classification.}
A label taxonomy organises a set of fine-grained classes as the leaf nodes $\tilde{V} \subset V$ of a rooted tree $T = (V, E)$, with internal nodes representing progressively coarser superclasses as we move from leaf nodes towards the root. A probabilistic classifier typically outputs a distribution $\tilde{\vec{p}} \in \Delta^{\tilde{N}-1}$ over the $\tilde{N} = |\tilde{V}|$ leaf classes for a given input, which can also be viewed as a distribution $\vec{p}\in\Delta^{N-1}$ by assigning zero mass to internal nodes. The true label is a single leaf $i \in \tilde{V}$. The graph structure encodes semantic similarity: for example, in image classification, confusing a Golden Retriever with a Labrador Retriever is a qualitatively different mistake from confusing it with an elephant, and evaluation metrics should reflect this.

\medskip

In both settings, the central question is the same: given a probabilistic estimate $\vec{p} \in \Delta^{N-1}$ and an observed true node $i \in V$, how should we evaluate $\vec{p}$ against $i$ in a way that is both sensitive to the graph structure of $G$ and statistically well-justified? We can formalise this challenge as the problem of constructing a \emph{scoring rule} $S(\vec{p}, i)$ that is \emph{graph-aware} and \emph{strictly proper}.

\section{Background on Proper Scoring Rules}
\label{sec:background}

In this section we give a brief overview of the theory of proper scoring rules. For a recent in-depth review of the vast statistical literature on this topic, see for example \cite{Waghmare:2025}. We restrict our attention to scoring rules for finite outcome spaces.

Throughout this section, let $\mathcal{Y}$ be a finite nonempty set of outcomes that we identify with the integers $1,\dots, N$, and let $\vec{p} = (p_1, \dots, p_N)^\top \in \Delta^{N-1}$ denote a probability distribution over $\mathcal{Y}$, where $p_j$ is the predicted probability of outcome $j$. A \emph{scoring rule} on $\mathcal{Y}$ is a function $S: \Delta^{N-1} \times \mathcal{Y} \to \overline{\R}$ that evaluates a probabilistic prediction $\vec{p}$ against an observed outcome $i \in \mathcal{Y}$. We adopt the convention that smaller scores indicate better predictions, although the opposite choice is also common.

\begin{definition}
    A scoring rule $S: \Delta^{N-1} \times \mathcal{Y} \to \overline{\R}$ is called \emph{proper} if
    \be\label{eqn:proper}
        \E_{i\sim\vec{q}}\,S(\vec{q}, i) ~\leq~
        \E_{i\sim\vec{q}}\,S(\vec{p}, i)
    \ee
    for every $\vec{p}, \vec{q} \in \Delta^{N-1}$. It is called \emph{strictly proper} if \eqref{eqn:proper} holds as a strict inequality whenever $\vec{p} \neq \vec{q}$.
\end{definition}

Much of the literature on proper scoring rules has developed in the context of forecast evaluation, with applications such as meteorology or predicton markets in mind. The central motivation for using proper scoring rules in these fields lies in incentivising the forecaster to reveal their true beliefs. If the forecaster's true belief is $\vec{q}$ but they report $\vec{p} \neq \vec{q}$, \eqref{eqn:proper} guarantees that they cannot expect to score any better by doing so. If the scoring rule is \emph{strictly} proper, they would in fact expect to incur a strictly positive penalty for misreporting.

The same logic applies when evaluating a probabilistic estimator rather than a forecaster: a strictly proper scoring rule ensures that the best expected score is achieved if and only if the model distribution $\vec{p}$ coincides with the true data distribution $\vec{q}$. Beyond evaluation, proper scoring rules also provide principled loss functions for probabilistic estimation and machine learning, since minimising their empirical average encourages $\vec{p}$ to match $\vec{q}$.

\subsection{Canonical examples}
\label{subsec:examples-of-proper-scoring-rules}

We briefly recall three canonical examples of strictly proper scoring rules on a finite outcome space $\mathcal{Y}$, all of which will reappear in the rest of the paper. Throughout, $\vec{p}\in\Delta^{N-1}$ denotes the estimated distribution, and $i\in\mathcal{Y}$ the observed ground truth.

The \emph{Brier score} is the oldest and most classical example:
\be\label{eqn:brier-score}
    \mathrm{Br}(\vec{p}, i) ~=~
    \|\vec{p} - \vec{e}_i\|_2^2 ~=~
    \sum_{j=1}^N (p_j - \delta_{ij})^2
    ~\in~ [0, 2]~,
\ee
where $\vec{e}_i$ denotes the $i$-th standard basis vector and $\delta_{ij}$ the Kronecker-Delta. It penalises probability mass assigned to any class other than the true one, while preferring mass to be spread evenly among wrong classes rather than concentrated on a single wrong class. This is a first illustration of the general principle that proper scoring rules balance incentivising well-justified sharpness with penalising unjustified overconfidence.

The \emph{logarithmic score}, better known as log-loss or cross-entropy loss in machine learning, is
\be\label{eqn:log-score}
    S_{\log}(\vec{p}, i) ~=~ -\log p_i ~\in~ [0, \infty)~.
\ee
This is a \emph{local} scoring rule in that it depends on the predicted distribution only through the probability assigned to the true outcome. Minimising it amounts to maximum likelihood estimation.

Now consider the case of ordinal classification, where the outcome space $\mathcal{Y}$ has a natural ordering. The \emph{Ranked probability score (RPS)} \citep{Epstein:1969} is defined as
\be\label{eqn:rps}
    \mathrm{RPS}(\vec{p}, i) ~=~
    \sum_{j'=1}^N \left(\Bigl(\sum_{j=1}^{j'} p_j\Bigr) - \boldone_{i \leq j'}\right)^2~\in~ [0, N-1]~.
\ee
where the inner sum is the predicted cumulative probability up to category $j'$. The RPS is the natural discrete analog of the popular \emph{Continuous ranked probability score (CRPS)} \citep{Gneiting:2007} for univariate real outcome spaces, and both are strictly proper. While outwardly similar to the Brier score, the RPS accounts for the additional structure on $\mathcal{Y}$: probability mass far away from the true category $i$ is penalised more heavily than mass placed on a neighbouring category. This provides a first example of a proper scoring rule that takes into account a \emph{relationship} between the categories. The main contribution of the present paper is to construct scoring rules for categorical variables with \emph{arbitrary} binary relationships between the categories, so that $\mathcal{Y}$ becomes the node set of a graph of size $N$. As we have seen, such outcome spaces arise naturally in many applications.

\subsection{Kernel scores}
\label{sec:kernel-scores}

We now focus on the class of proper scoring rules known as \emph{kernel scores}, as they provide the foundation for our proposed graph-aware scoring rules.

Let $k:\mathcal{Y}\times\mathcal{Y}\to\mathbb{R}$ be a symmetric function, called the \emph{kernel}. Let $K=[k(i,j)]_{i,j=1}^N$ denote the corresponding kernel matrix, and write $\vec{1}^\perp := \{\vec{x}\in\R^N \mid \sum_{j=1}^N x_j = 0\}$ for the orthogonal complement of the $N$-dimensional vector of ones $\vec{1} = (1,...,1)^\top$. Define the associated quadratic form by
\begin{equation}
    b_k(\vec{x}) := \vec{x}^{\top}K\vec{x} \qquad\text{for }\vec{x}\in\mathbb{R}^N.
\end{equation}
The following definitions are well-known:

\begin{definition}
\label{def:kernel-definiteness}
    The kernel $k$ is called
    \begin{itemize}[topsep=0pt, itemsep=1pt, parsep=0pt, partopsep=0pt]
        \item \emph{positive definite (p.d.)} if $b_k(\vec{x})\geq0$ for every $\vec{x}\in\mathbb{R}^N$,
        \item \emph{negative definite (n.d.)} if $b_k(\vec{x})\leq0$ for every $\vec{x}\in\mathbb{R}^N$,
        \item \emph{conditionally positive definite (c.p.d.)} if $b_k(\vec{x})\geq0$ for every $\vec{x}\in\vec{1}^\perp$,
        \item \emph{conditionally negative definite (c.n.d.)} if $b_k(\vec{x})\leq0$ for every $\vec{x}\in\vec{1}^\perp$.
    \end{itemize}
    Moreover, $k$ is called \emph{strictly p.d.}, \emph{strictly n.d.}, \emph{strictly c.p.d.}, or \emph{strictly c.n.d.}, respectively, if the corresponding inequality is strict whenever $\vec{x}\neq\vec{0}$ in the respective domain.
\end{definition}

Evidently, c.p.d. (c.n.d.) kernels include all p.d. (n.d.) kernels. Moreover, there are subtle connections between p.d. (n.d.) and c.n.d. (c.p.d.) kernels: the two notions are dual to each other in a non-trivial way \citep{Berg:2012:book}. For the purposes of this paper, however, we only need the simple observation that if $k$ is p.d., then $-k$ is n.d., and in particular c.n.d.

For a c.n.d.\ kernel $k: \mathcal{Y} \times \mathcal{Y} \to \mathbb{R}$, the corresponding \emph{kernel score} is defined as
\be\label{eqn:kernel-score}
    S_k(\vec{p}, i) ~=~ \E_{J \sim \vec{p}}\bigl[k(J, i)\bigr] - \frac{1}{2}\,\E_{J, J' \sim \vec{p}}\bigl[k(J, J')\bigr]~,
\ee
This expectation-based form is, in fact, more general and applies to arbitrary outcome spaces. In our finite setting, it can be written in the following simple vector form:
\be\label{eqn:kernel-score-finite}
    S_k(\vec{p}, i) ~=~ \vec{p}^\top K \vec{e}_i - \frac{1}{2}\vec{p}^\top K \vec{p}~.
\ee
This is a proper scoring rule if and only if $k$ is c.n.d., and strictly proper if and only if $k$ is strictly c.n.d.\ \citep{Gneiting:2007}. This is a significant simplification compared to infinite outcome spaces, where strict positive definiteness of the underlying p.d.\ kernel is still sufficient but no longer necessary for strict propriety of the kernel score \citep{Ziegel:2024}.

The first and second term in~\eqref{eqn:kernel-score-finite} have a clear interpretation: the first is a calibration term measuring the extent to which $\vec{p}$ is miscalibrated with respect to the true outcome $i$, while the second is an entropy term quantifying the spread of $\vec{p}$ independently of $i$. Their opposing signs mean that the kernel score balances calibration against sharpness. As mentioned above, this is a general feature of proper scoring rules.

There is an alternative convention for kernel scores which involves an additional term \citep{Ziegel:2024},
\be\label{eqn:kernel-score-finite-+}
    S_k^+(\vec{p}, i) ~=~ S_k(\vec{p}, i) - \tfrac{1}{2}\,K_{ii}~.
\ee
The scoring rules $S_k$ and $S_k^+$ are \emph{strongly equivalent} in that they differ only by a term that does not depend on $\vec{p}$. Strong equivalence preserves propriety. We prefer the $S_k^+$ convention because it yields the clean quadratic form
\be\label{eqn:kernel-score-finite-quadraticform-+}
    S_k^+(\vec{p},i) ~=~ -\tfrac{1}{2}(\vec{p}-\vec{e}_i)^\top K (\vec{p}-\vec{e}_i)~,
\ee
which is always non-negative: since $\vec{p}-\vec{e}_i\in\vec{1}^\perp$ for all $\vec{p}\in\Delta^{N-1}$ and $i\in\mathcal{Y}$, c.n.d.\ of $K$ implies $S_k^+(\vec{p},i)\geq 0$. Note, that if $-k$ is strictly p.d., then $S_k^+$ can be seen as a special case of the squared maximum mean discrepancy \citep{Gretton:2012} between $\vec{p}$ and a point mass at $i$. Moreover, in the case where $\mathcal{Y}$ is the node set of a graph (see Section \ref{sec:scores}), it is a type of \emph{node vector distance} \citep{Coscia:2023} between $\vec{p}$ and $\vec{e}_i$.

As it turns out, two of the scores introduced in the previous section are, in fact, kernel scores. The Brier score corresponds to $k(j,j') = 1-\delta_{jj'}$, while the RPS corresponds to $k(j, j') = |j - j'|$ \citep{Gneiting:2007}. For details, see Appendix~\ref{app:rps}.

\section{Related Work}
\label{sec:related_work}

\paragraph{Evaluation metrics for node selection tasks.}

For non-probabilistic methods, the standard aggregate evaluation metric is accuracy (or, equivalently, the error rate). Probabilistic, likelihood-based or ranking-based methods typically report the (inverse) rank of the true node or top-$K$ accuracy \citep{Lokhov:2014, Altarelli:2014, Antulov:2015, Sha:2021}. A popular alternative is the Error distance \citep{Antulov:2015,Zhu:2016,Jiang:2017,Paluch:2018}, also known as Graph error distance or Distance error, which measures the shortest-path distance between the top-$1$ prediction and the true node. Error distance is graph-aware but improper (see Section \ref{subsec:traditional_scores}). \citet{Paluch:2020} propose the $\alpha$-credible set size (CSS$_\alpha$), i.e., the number  of nodes needed to cover a given fraction of predicted probability mass, as a complementary sharpness measure. The use of CSS$_\alpha$ implicitly acknowledges that rank and accuracy fail to capture important aspects of the full distribution, yet it does not constitute a principled  solution: it is not a proper scoring rule and does not account for graph structure.

\paragraph{Proper scoring rules and kernel scores.}

The statistical literature on proper scoring rules has largely focused on standard outcome spaces. \citet{Gneiting:2007} give a comprehensive treatment of proper scoring rules and kernel scores on Euclidean spaces. They establish the connection between c.n.d.\ kernels and strictly proper kernel scores and introduce energy scores and CRPS as canonical examples. Notable extensions to non-standard outcome spaces include the circular CRPS of \citet{Grimit:2006}, the adaptation of the Hyvärinen score to Riemannian manifolds \citep{Mardia:2016} and to $\{-1,1\}^N$ \citep{Hyvarinen:2007}, and proper scoring rules for top-$k$ list predictions \citep{Resin:2023}, where the prediction consists of the $k$ most likely classes together with associated confidence levels.

\paragraph{Proper scoring rules on node sets of graphs.}

The only prior work to study proper scoring rules specifically on the node set of a graph is \citet{Dawid:2012}, who characterise proper scoring rules which are local in the sense that the score only depends on the total probability assigned to a neighbourhood of the true node. They show for an important class of such scoring rules (namely those that only depend on probability ratios rather than absolute probabilities) that they can be decomposed as a superposition of the logarithmic score and proper scoring rules on maximal cliques. While elegant, this result does not provide scoring rules that evaluate the entire probability distribution according to its proximity to the true node, which is the primary goal of the present paper.

\paragraph{Kernels and distributions on node sets of graphs.} A number of machine learning papers explicitly consider p.d.\ kernels on the node set of graphs. The seminal work of \citet{Smola:2003} constructs kernels as functions of the graph Laplacian, defining diffusion kernels and random walk kernels as special cases. \citet{Fouss:2007} introduce the resistance or commute-time kernel, but the convergence results of \citet{vonLuxburg:2014} show that resistance distance has limited discrimination power on large graphs unless they are sufficiently sparse. \citet{Fouss:2012} review and compare a variety of Laplacian-based kernels on node sets of graphs. \citet{Borovitskiy:2021} generalise Matérn kernels to node sets of graphs with a construction that also makes use of the graph Laplacian. Meanwhile, \citet{Devriendt:2022} study probability distributions over node sets directly, introducing a notion of variance based on the effective resistance distance. All three of these papers use the graph Laplacian as the central object encoding proximity between nodes, and their constructions inform the scoring rules we propose in Section~\ref{sec:scores}. \citet{Coscia:2023} show how Laplacian-based node vector distances can be computed efficiently using Laplacian solvers. Their framework applies directly to computing the Resistance score  (Section~\ref{subsec:efficient_computation}). The Diffusion score introduced in Section~\ref{subsec:diffusion_score} is also informed by the Laplacian renormalisation group of \citet{Villegas:2023}, which uses the diffusion kernel to coarse-grain graphs at different scales.

\paragraph{Evaluation metrics for hierarchical classification.}

There is a large literature on hierarchy-aware evaluation metrics \citep{Silla:2011, Kosmopoulos:2015, Zangari:2024}. Most of these metrics operate on point estimates rather than probability distributions over nodes, and some of them apply to estimated node subsets rather than our single node selection setting. Hierarchical precision, recall and F-score \citep{Kiritchenko:2005} measure set overlap between the predicted subset of nodes and a true subset of nodes. Tree-distance metrics \citep{Ramaswamy:2015, Cao:2024} score a single estimated node against the true node. Metrics based on the lowest common ancestor (LCA), including the semantic similarity of \citet{WuPalmer:1994} and the Hierarchical error distance \citep{Russakovsky:2015} and its generalisation to distance@$K$ \citep{Bertinetto:2020, Karthik:2021}, depend on the distribution only through its mode or its ranking. Separately, \citet{Valmadre:2022} evaluates hierarchy-aware sensitivity and specificity by sweeping an inference threshold, and \citet{Karthik:2021} and \citet{Plaud:2025} derive post-hoc Bayes-optimal point estimates from a distributional estimate under hierarchical costs. Neither line of work addresses the evaluation of the distribution itself. To the best of our knowledge, the only proper scoring rules for hierarchical node selection are the hierarchical cross-entropy (HXE) of \citet{Bertinetto:2020} and its generalisations, whose propriety (but not strict propriety) was recognised and proved by \citet{Urbani:2024}.

\section{Strictly proper scoring rules for node selection}
\label{sec:scores}

We now consider scoring rules for the case where the set of outcomes is the node set of a graph. As in Section \ref{sec:problem_statement}, we assume $G=(V,E)$ to be a simple graph of size $N=|V|$, and we set $\mathcal{Y}=V$, identifying the nodes $V$ with integers $1,\dots, N$.

Let $A\in\{0,1\}^{N\times N}$ be the adjacency matrix of $G$. Denote by $d_j=\sum_{j'=1}^N A_{jj'}$ the degree of node $j$, and let $D\in\R^{N\times N}$ be the degree matrix, i.e., the diagonal matrix with entries $D_{jj'} = d_j\delta_{jj'}$. The \emph{graph Laplacian} is defined as $L = D - A$, with eigenvalues $0=\lambda_1 \leq \lambda_2 \leq \dots \leq \lambda_N$ and associated orthonormal eigenbasis $\vec{v}_1, \vec{v}_2, \dots, \vec{v}_N \in \R^N$, where $\vec{v}_1 = \frac{1}{\sqrt{N}}\vec{1}$. We write $\Id$ for the $N\times N$ identity matrix and $\vec{e}_j \in \{0,1\}^N$ for the $j$-th standard basis vector, defined by $(\vec{e}_i)_j = \delta_{ij}$. It will be repeatedly useful to recall the fact that $\vec{p} - \vec{e}_i \in \vec{1}^\perp$ for all $\vec{p} \in \Delta^{N-1}$ and $i \in V$.

We will sometimes refer to special graphs: $K_N$ denotes the \emph{complete graph} on $N$ nodes, with Laplacian $L = N\Id - \vec{1}\vec{1}^\top$ and non-zero eigenvalues $\lambda_2 = \dots = \lambda_N = N$. $P_N$ denotes the \emph{path graph} with nodes $1,\dots, N$ and edges $(j, j+1)$ for $j = 1, \dots, N-1$. Moreover, $C_N$ denotes the \emph{cycle graph}, which is the same as $P_N$ but with an additional edge $(N, 1)$ to close the cycle.

We construct our scoring rules within the kernel score framework of Section~\ref{sec:kernel-scores}. Specifically, we design scores of the form displayed in \eqref{eqn:kernel-score-finite-+} with kernels derived from the graph Laplacian $L$. In each case, it suffices to find a (strictly) c.n.d.\ kernel matrix $K$ on $V$; the score $S_k^+(\vec{p},i)$ is then automatically non-negative and (strictly) proper.

Throughout this section, we adopt the convention of the kernel literature, consistent with our Definition \ref{def:kernel-definiteness} (and aligned with the definition of proper and strictly proper scoring rules), that a symmetric matrix $M \in \R^{N \times N}$ is called \emph{positive definite} (rather than positive semi-definite) if $\vec{x}^\top M \vec{x} \geq 0$ for all $\vec{x} \in \R^N$, and \emph{strictly positive definite} (rather than just positive definite) if the inequality is strict for all $\vec{x} \neq \vec{0}$. Under this convention, the graph Laplacian $L$ is p.d., and its Moore--Penrose pseudoinverse $L^\dagger$ is p.d.\ as well: if $L = \sum_{k=1}^N \lambda_k \vec{v}_k \vec{v}_k^\top$ is the eigendecomposition of $L$, then $L^\dagger = \sum_{k:\,\lambda_k > 0} \lambda_k^{-1} \vec{v}_k \vec{v}_k^\top$, which has exclusively non-negative eigenvalues.

\subsection{A first Laplacian score}
\label{subsec:laplace_stepping_stone}

The graph Laplacian $L$ is a natural candidate kernel matrix, as it encodes one-hop proximity between nodes. The following lemma ensures that it can indeed be used to construct a proper kernel score:

\begin{lemma}
\label{lem:neglap_is_cnd}
    The negative Laplacian $-L$ is conditionally negative definite. If the graph is connected, then $-L$ is strictly c.n.d.
\end{lemma}
\begin{proof}
    $L$ is positive definite, thus $-L$ is trivially c.n.d.\ If the graph is connected, then the null space of $-L$ is one-dimensional and is spanned by $\vec{v}_1=\vec{1}$. Thus, any nonzero $\vec{x}\in\vec{1}^\perp$ is not in the null space, and since $-L$ is negative definite, this implies that $-\vec{x}^\top L \vec{x}  < 0$.
\end{proof}

Using~\eqref{eqn:kernel-score-finite-+} with $K = -L$, the corresponding score reads
\be\label{eqn:s_lap}
    S_{-L}^+(\vec{p},i)
    ~=~
    \tfrac{1}{2}\vec{p}^\top L\vec{p}
    - \vec{e}_i^\top L\vec{p}
    + \tfrac{d_i}{2} \, .
\ee
By Lemma~\ref{lem:neglap_is_cnd}, the kernel score $S_{-L}^+$ is a strictly proper scoring rule on any connected graph. This score rewards probability mass concentrated on neighbours of $i$ (via the second term) and penalises globally rough distributions (via the first term). In this sense, the score is reminiscent of the Hyvärinen score \citep{Hyvarinen:2005} for distributions on $\R^n$. (For a modern introduction to this score we refer to the review by \cite{Waghmare:2025}). However, since the Laplacian is zero everywhere except on the diagonal and the first off-diagonal, the calibration term of the score sees only the one-hop neighbourhood of the true node: any probability mass at shortest path distance two or more from the true node is treated identically by $S_{-L}^+$. Thus, $S_{-L}^+$ is at best a first order refinement of the Brier score. This motivates looking for a kernel that encodes the global graph structure rather than just the local adjacency structure.

Before we do just that, let us make the following observation. On the complete graph $K_N$, all local neighbourhoods are identical with $V$. The incentives of the score $S_{-L}^+$ thus transform exactly into those of the Brier score. More formally, $S_{-L}^+$ is a generalisation of the Brier score in the following sense:

\begin{proposition}\label{prop:lap_score_to_brier}
    On the complete graph $K_N$, $S_{-L}^+$ is strongly equivalent to the
    multi-category Brier score.
\end{proposition}
\begin{proof}
    Consider the complete graph $K_N$ and its Laplacian $L=N\Id - \vec{1}\vec{1}^\top$. The Laplace score in its product form (\eqref{eqn:kernel-score-finite-quadraticform-+}) is then: $S_{-L}(\vec{p},i) = \frac{1}{2}(\vec{p}-\vec{e}_i)^\top (N\Id -\vec{1}\vec{1}^\top) (\vec{p}-\vec{e}_i)$. Now observe that $\vec{1}^\top (\vec{p}-\vec{e}_i) = 0$. Hence, the score reduces to $S_{-L}(\vec{p},i) = \frac{N}{2}(\vec{p}-\vec{e}_i)^\top (\vec{p}-\vec{e}_i) = \frac{N}{2}\mathrm{Br}(\vec{p}, i)$.
\end{proof}

\subsection{The Resistance score}
\label{subsec:resistance_score}

A natural upgrade to the one-hop Laplacian kernel score is to construct a scoring rule based on a notion of distance between nodes. \emph{Effective resistance distance} (\cite{Klein:1993}) is one of the distances that can be defined on the nodes of a graph. For nodes $i, j \in V$, it is defined as
\be\label{eqn:omega1}
    \omega_{jj'} ~:=~
    (\vec{e}_j - \vec{e}_{j'})^\top L^\dagger (\vec{e}_j - \vec{e}_{j'}) \, ,
\ee
where $L^\dagger$ denotes the Moore-Penrose pseudoinverse of $L$. Note that $\omega_{jj'}$ is symmetric and vanishes if and only if $j=j'$. Moreover, all resistance distances are non-negative. This is an immediate consequence of \eqref{eqn:omega1}, because $L^\dagger$ is positive definite. An alternative way to write the resistance distance is as follows:
\be\label{eqn:omega2}
    \omega_{jj'} ~=~ L^\dagger_{jj} + L^\dagger_{j'j'} - 2L^\dagger_{jj'} \, .
\ee
As the name betrays, the origins of this distance can be found in physics: the effective resistance distance between two nodes is equivalent to the electric resistance between them if the graph is viewed as an electric network with each edge representing a resistor of unit weight.

Let $\Omega = [\omega_{ij}]_{j,j'=1}^N$ denote the matrix of all pairwise resistance distances.

\begin{lemma}
\label{lem:omega_is_neglapdagger_on_subspace}
    If $\vec{x} \in \vec{1}^\perp$, then $\vec{x}^\top\Omega\vec{x} = -2\vec{x}^\top L^\dagger\vec{x}$.
\end{lemma}
\begin{proof}
    With $\vec{d} = \mathrm{diag} (L^\dagger)$ we can write $\Omega = \vec{d} \vec{1}^\top + \vec{1} \vec{d}^\top - 2L^\dagger$. The matrix $\vec{d}\vec{1}^\top$ is rank 1 with null space $\vec{1}^\perp$, from which follows that $\vec{x}^\top (\vec{d} \vec{1}^\top + \vec{1} \vec{d}^\top) \vec{x} = 0$ for all $\vec{x}\in\vec{1}^\perp$.
\end{proof}
\begin{lemma}
\label{lem:omega_is_cnd}
    The matrix $\Omega$ is conditionally negative definite. If the graph is connected, then $\Omega$ is strictly conditionally negative definite.
\end{lemma}
\begin{proof}
    By Lemma~\ref{lem:omega_is_neglapdagger_on_subspace}, for every $\vec{x}\in\vec{1}^\perp$, we have $\vec{x}^\top\Omega\vec{x} = -2\vec{x}^\top L^\dagger\vec{x} \leq 0$, since $L^\dagger$ is positive definite. Hence, $\Omega$ is c.n.d.\ If the graph is connected, then the null space of $L^\dagger$ is spanned by $\vec{1}$. Thus, every nonzero $\vec{x}\in\vec{1}^\perp$ lies outside the null space, and since $L^\dagger$ is positive definite, we have $\vec{x}^\top L^\dagger\vec{x}>0$. Consequently, $\vec{x}^\top\Omega\vec{x} = -2\vec{x}^\top L^\dagger\vec{x}<0$ for every nonzero $\vec{x}\in\vec{1}^\perp$, proving that $\Omega$ is strictly c.n.d.
\end{proof}

Lemma~\ref{lem:omega_is_cnd} is a special case of a classical result: since $L^\dagger$ is a p.d.\ kernel, it induces a Hilbert space embedding of $V$, and the squared Euclidean distance between nodes $j$ and $j'$ in this embedding space is precisely $\omega_{jj'}$. By Schoenberg's theorem (see, e.g., \cite{Berg:2012:book}), the squared distance in a Hilbert space is always c.n.d., which yields Lemma \ref{lem:omega_is_cnd} as a corollary. The use of $L^\dagger$ as a p.d.\ kernel on the node set of a graph appears explicitly in \citet{Fouss:2007}.

\begin{definition}
    The \emph{Resistance score} $S_{\mathrm{res}}: \Delta^{N-1} \times V \to \mathbb{R}_{\geq 0}$ is the kernel score $S_\Omega^+$ associated with $\Omega$:
    \be\label{eqn:s_omega2}
        S_{\mathrm{res}}(\vec{p},i) ~:=~ \vec{e}_i^\top\Omega\vec{p} - \tfrac{1}{2}\vec{p}^\top\Omega\vec{p}
    \ee
    Since $\omega_{jj}=0$, it coincides with $S_\Omega$.
\end{definition}

From Lemma \ref{lem:omega_is_cnd}  follows:

\begin{proposition}
    The Resistance score is a proper scoring rule. If the graph is connected, it is strictly proper.
\end{proposition}

By Lemma~\ref{lem:omega_is_neglapdagger_on_subspace} and~\eqref{eqn:kernel-score-finite-quadraticform-+}, the Resistance score admits the alternative representation
\be\label{eqn:s_omega1}
    S_{\mathrm{res}}(\vec{p},i) ~=~  (\vec{p}-\vec{e}_i)^\top L^\dagger (\vec{p}-\vec{e}_i) \, ,
\ee
which shows that it measures the node vector distance between $\vec{p}$ and $\vec{e}_i$ induced by the p.d.\ kernel $L^\dagger$ \citep{Coscia:2023}. Using the notation of \citet{Devriendt:2022}, the Resistance score can also be written as
\be
    S_{\mathrm{res}}(\vec{p},i) ~=~ [\Omega\vec{p}]_i - \mathrm{Var}_\omega(\vec{p}) \, ,
\ee
where $\mathrm{Var}_\omega(\vec{p}) = \frac{1}{2}\vec{p}^\top\Omega\vec{p}$ is the graph variance of $\vec{p}$. The first term penalises probability mass in proportion to its resistance distance from the true node. The second term acts as a regulariser that discourages excessive concentration of mass in a small region of the graph.

\paragraph{Connections to classical proper scoring rules.}
\label{subsubsec:resistance_connections}

Like the Laplace score, the Resistance score reduces to the Brier score when the graph structure becomes irrelevant:

\begin{proposition}
\label{prop:resistance_score_to_brier}
    On the complete graph $K_N$, the Resistance score is strongly equivalent to
    the multi-category Brier score.
\end{proposition}

\begin{proof}
    On the complete graph we find that $L^\dagger ~=~ \sum_{j=2}^N \frac{1}{\lambda_j}\vec{v}_j \vec{v}_j^\top ~=~ \frac{1}{N}\sum_{j=2}^N \vec{v}_j \vec{v}_j^\top ~=~ \frac{1}{N} (\Id - \vec{v}_1 \vec{v}_1^\top)~=~ \frac{1}{N^2}L$, where we have used the property of orthonormal bases that $\Id = \sum_{j=1}^N \vec{v}_j\vec{v}_{j}^\top$. As a consequence, the Resistance score in the form of \eqref{eqn:s_omega1} reduces (up to a constant) to the Laplacian score of \eqref{eqn:s_lap}, $S_{\mathrm{res}}(\vec{p}, i) = \frac{2}{N^2}S_{-L}(\vec{p}, i)$. But, as was stated in Proposition \ref{prop:lap_score_to_brier}, this is equivalent to the Brier score, specifically $S_{\mathrm{res}}(\vec{p}, i) = \frac{1}{N}\mathrm{Br}(\vec{p}, i)$.
\end{proof}

To state the next connection, we first recall a fundamental property of resistance distance, which will also be used in Section~\ref{subsec:hierarchical_classification}.

\begin{proposition}[\cite{Klein:1993}]
\label{prop:resistance_equals_shortest_path_on_trees}
    On any tree, the effective resistance distance $\omega_{jj'}$ equals the shortest-path distance $d(j,j')$ for all nodes $j,j'\in V$.
\end{proposition}

With Proposition~\ref{prop:resistance_equals_shortest_path_on_trees} at our disposal, the following result is immediate:

\begin{proposition}
\label{prop:resistance_score_to_rps}
    On the path graph $P_N$, the Resistance score equals the Ranked probability score~(RPS) for $N$ ordinal categories.
\end{proposition}
\begin{proof}
    As noted in Section~\ref{sec:kernel-scores}, the RPS is a kernel score on the integers $1,\dots, N$ with associated kernel $k(j,j') = |j-j'|$ (a proof of this well-known result is supplied in Appendix~\ref{app:rps}). Identifying these integers with the nodes of the path graph, the kernel $k$ equals the shortest-path metric on $P_N$. Since $P_N$ is a tree, Proposition~\ref{prop:resistance_equals_shortest_path_on_trees} gives $\omega_{jj'} = |j-j'|$, so the two kernel scores coincide: $S_{\mathrm{res}}(\vec{p},i) = \mathrm{RPS}(\vec{p},i)$.
\end{proof}

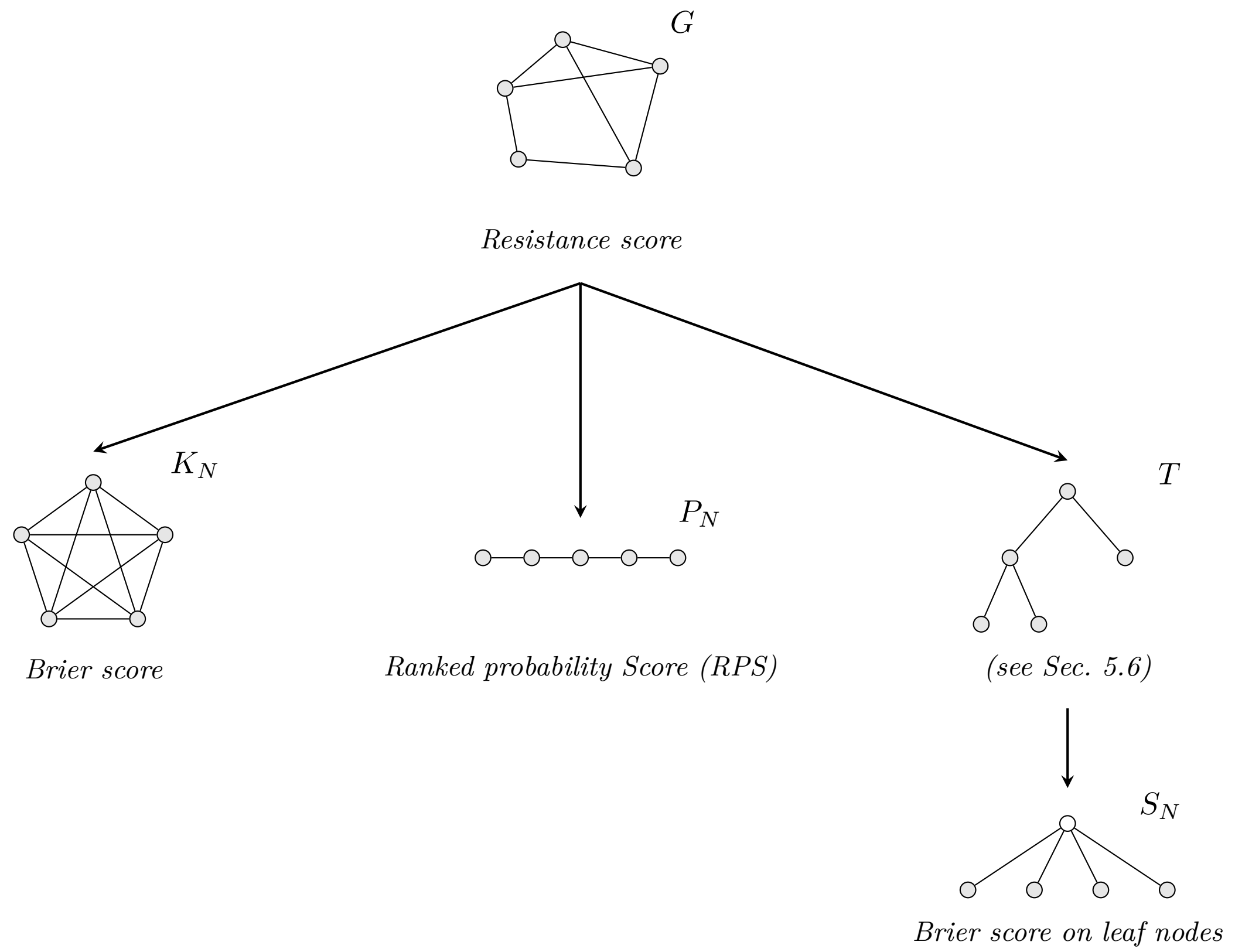
\begin{figure}[t]
\centering
\begin{tikzpicture}[
    every node/.style={},
    gnode/.style={
        circle,
        draw=black,
        fill=black!10,
        inner sep=0pt,
        minimum size=5pt
    },
    rnode/.style={
        circle,
        draw=black,
        fill=white,
        inner sep=0pt,
        minimum size=5pt
    },
    glabel/.style={
        font=\normalsize\itshape
    },
    slabel/.style={
        font=\small\itshape,
        text width=6cm,
        align=center
    },
    garrow/.style={
        ->,
        >=stealth,
        thick
    },
    sarrow/.style={
        ->,
        >=stealth,
        thick
    }
]

\def\Gx{6.5}
\def\Gy{5.0}
\def\Gr{0.85}

\node[gnode] (G1) at (\Gx-0.20, \Gy+0.85)  {};
\node[gnode] (G2) at (\Gx+0.90, \Gy+0.55)  {};
\node[gnode] (G3) at (\Gx+0.60, \Gy-0.60)  {};
\node[gnode] (G4) at (\Gx-0.70, \Gy-0.50)  {};
\node[gnode] (G5) at (\Gx-0.85, \Gy+0.30)  {};

\draw (G1)--(G2); \draw (G2)--(G3); \draw (G3)--(G4); \draw (G4)--(G5);
\draw (G1)--(G5); \draw (G1)--(G3); \draw (G2)--(G5);

\node[glabel] at (\Gx+1.15, \Gy+1.05) {$G$};

\def\Kx{1.0}
\def\Ky{0.0}
\def\Kr{0.85}

\node[gnode] (K1) at (\Kx+0.00, \Ky+\Kr)   {};
\node[gnode] (K2) at (\Kx+0.81, \Ky+0.26)  {};
\node[gnode] (K3) at (\Kx+0.50, \Ky-0.69)  {};
\node[gnode] (K4) at (\Kx-0.50, \Ky-0.69)  {};
\node[gnode] (K5) at (\Kx-0.81, \Ky+0.26)  {};

\foreach \i in {1,...,5} {
    \foreach \j in {1,...,5} {
        \pgfmathparse{int(\i<\j)}
        \ifnum\pgfmathresult=1
            \draw (K\i)--(K\j);
        \fi
    }
}

\node[glabel] at (\Kx+1.15, \Ky+1.05) {$K_N$};
\node[slabel] at (\Kx, \Ky-1.25) {\textit{Brier score}};

\def\Px{6.5}
\def\Py{0.0}
\def\Pstep{0.55}

\node[gnode] (P1) at (\Px-2*\Pstep, \Py) {};
\node[gnode] (P2) at (\Px-1*\Pstep, \Py) {};
\node[gnode] (P3) at (\Px+0*\Pstep, \Py) {};
\node[gnode] (P4) at (\Px+1*\Pstep, \Py) {};
\node[gnode] (P5) at (\Px+2*\Pstep, \Py) {};

\draw (P1)--(P2)--(P3)--(P4)--(P5);

\node[glabel] at (\Px+1.35, \Py+0.50) {$P_N$};
\node[slabel] at (\Px, \Py-1.25) {\textit{Ranked probability Score (RPS)}};

\def\Tx{12.0}
\def\Ty{0.0}
\def\Tvgap{0.75}
\def\Thgap{0.65}

\node[gnode] (Tr)  at (\Tx,             \Ty+\Tvgap)   {};   
\node[gnode] (Tl)  at (\Tx-\Thgap,     \Ty)           {};   
\node[gnode] (Trr) at (\Tx+\Thgap,     \Ty)           {};   
\node[gnode] (Tll) at (\Tx-1.5*\Thgap, \Ty-\Tvgap)   {};   
\node[gnode] (Tlr) at (\Tx-0.5*\Thgap, \Ty-\Tvgap)   {};   

\draw (Tr)--(Tl); \draw (Tr)--(Trr);
\draw (Tl)--(Tll); \draw (Tl)--(Tlr);

\node[glabel] at (\Tx+1.15, \Ty+\Tvgap+0.20) {$T$};
\node[slabel] at (\Tx, \Ty-1.25)
    {\textit{(see Sec.~\ref{subsec:hierarchical_classification})}};

\def\Sx{\Tx}      
\def\Sy{-3.0}     
\def\Sr{0.75}     

\node[rnode] (Sr0) at (\Sx,            \Sy)       {};   

\node[gnode] (Sl1) at (\Sx-1.5*\Sr,   \Sy-\Sr)   {};   
\node[gnode] (Sl2) at (\Sx-0.5*\Sr,   \Sy-\Sr)   {};   
\node[gnode] (Sl3) at (\Sx+0.5*\Sr,   \Sy-\Sr)   {};   
\node[gnode] (Sl4) at (\Sx+1.5*\Sr,   \Sy-\Sr)   {};   

\draw (Sr0)--(Sl1); \draw (Sr0)--(Sl2);
\draw (Sr0)--(Sl3); \draw (Sr0)--(Sl4);

\node[glabel] at (\Sx+1.05, \Sy+0.20) {$S_N$};
\node[slabel] at (\Sx, \Sy-\Sr-0.50)
    {\textit{Brier score on leaf nodes}};

\def\Gbase{3.1}

\draw[garrow] (\Gx, \Gbase) -- (\Kx, \Ky+\Kr+0.35);
\draw[garrow] (\Gx, \Gbase) -- (\Px, \Py+0.45);
\draw[garrow] (\Gx, \Gbase) -- (\Tx, \Ty+\Tvgap+0.35);

\draw[sarrow] (\Tx, \Ty-1.7) -- (\Tx, \Sy+0.4);

\node[slabel] at (\Gx, \Gy-\Gr-0.55)
    {\textit{Resistance score}};

\end{tikzpicture}
\caption{%
    The Resistance score on a general graph $G$ specialises to the
    Brier score on the complete graph $K_N$ (Proposition \ref{prop:resistance_score_to_brier}), to the Ranked probability
    score (RPS) on the path graph $P_N$ (Proposition \ref{prop:resistance_score_to_rps}), and to a hierarchy-aware
    score on a rooted tree $T$ (Section~\ref{subsec:hierarchical_classification}).
    On a depth-1 tree (star graph $S_N$), the latter further
    specialises to the Brier score on the leaf nodes
    (Proposition~\ref{prop:star_graph_reduction}).
    Arrows indicate specialisation.
}
\label{fig:graph_specialisations}
\end{figure}

In summary, Propositions~\ref{prop:resistance_score_to_brier} and~\ref{prop:resistance_score_to_rps} show that the Resistance score builds a bridge between two of the most widely used proper scoring rules: the Brier score (recovered on $K_N$) and the RPS (recovered on $P_N$). To our knowledge, this connection is new in the literature. More generally, the Resistance score can be viewed as an extension of the generalised energy score of~\citet{Gneiting:2007} with $\beta=1$, with the Euclidean distance replaced by the effective resistance metric. The special case when $G$ is a rooted tree is discussed in Section \ref{subsec:hierarchical_classification}. The different special cases of the Resistance score are summarised in Figure \ref{fig:graph_specialisations}.

%
%

As an aside, on the cycle graph $C_N$ we have
$\omega_{jj'} = d_{jj'}(N-d_{jj'})/N$ where
$d_{jj'}=\min(|j-j'|, N-|j-j'|)$. In the continuum limit $N\to\infty$, the Resistance score on $C_N$ therefore does \emph{not} converge to the \emph{circular CRPS} defined in \cite{Grimit:2006}. The latter is linear in the angular distance, whereas the limit of the Resistance score is quadratic in the angular distance.

\paragraph{Large graphs.} An interesting consideration concerns the behaviour of the resistance score on large graphs. On many large graphs, \citet{vonLuxburg:2014} show that effective resistance distance converges asymptotically to a function only of node degree:
\be
\omega_{jj'} \xrightarrow{N\to\infty} \frac{1}{d_j} + \frac{1}{d_{j'}}\, , 
\ee
or equivalently $\Omega \approx \vec{w}\vec{1}^\top + \vec{1}\vec{w}^\top$ where $w_j = 1/d_j$. This effect does not occur on sufficiently sparse graphs, where the resistance distance remains asymptotically aware of the full graph topology. The convergence result makes resistance distance unattractive as a similarity or distance measure on large graphs that are not sufficiently sparse. However, the leading term $\vec{w}\vec{1}^\top + \vec{1}\vec{w}^\top$ contributes nothing to the Resistance score: for any $\vec{x} = \vec{p} - \vec{e}_i \in \vec{1}^\perp$,
\be
\vec{x}^\top(\vec{w}\vec{1}^\top + \vec{1}\vec{w}^\top)\vec{x} = (\vec{x}^\top\vec{w})(\vec{1}^\top\vec{x}) + (\vec{x}^\top\vec{1})(\vec{w}^\top\vec{x}) = 0\, ,
\ee
since $\vec{x}^\top\vec{1} = \vec{1}^\top\vec{x}=0$. The Resistance score therefore depends on the second-order term $\Omega - (\vec{w}\vec{1}^\top + \vec{1}\vec{w}^\top)$ in the asymptotic expansion, and \citeauthor{vonLuxburg:2014}'s argument does not obviously invalidate it even on denser graphs. Whether this residual retains meaningful global topological information is an open question. If it reduces to quantities that depend on local neighbourhood structure only, the Resistance score would reflect only local graph properties in that regime. The Diffusion score, introduced in the next section, provides a more robustly topology-sensitive alternative at the cost of introducing a free parameter.

\subsection{The Diffusion score}
\label{subsec:diffusion_score}

Another Laplacian-based kernel on $G$ is the \emph{heat kernel} or \emph{diffusion kernel} \citep{Smola:2003},
\be\label{eqn:diffusion-kernel}
    H^\tau ~:=~ e^{-\frac{\tau}{2}L}
    ~=~ \sum_{j=1}^N
        e^{-\frac{\tau}{2}\lambda_j}\vec{v}_j\vec{v}_j^\top \, ,
    \quad \tau \geq 0 \, ,
\ee
where the parameter $\tau$ controls the diffusion time.

\begin{definition}
    We define the $\tau$-\emph{Diffusion score} $S_{\mathrm{diff}}^\tau: \Delta^{N-1}\times V \to \R_{\geq 0}$ by
    \be\label{eqn:s_tau}
        S_{\mathrm{diff}}^{\tau}(\vec{p},i) :=  (\vec{p}-\vec{e}_i)^\top H^\tau (\vec{p}-\vec{e}_i) \, .
    \ee
\end{definition}

The diffusion kernel is strictly positive definite no matter whether the underlying graph is connected or not. Therefore, $-H^\tau$ is strictly c.n.d., and we have:
\begin{proposition}
    The $\tau$-Diffusion score is a strictly proper scoring rule on any simple graph.
\end{proposition}

As for the Resistance score, we find:
\begin{proposition}
\label{prop:diffusion_score_to_brier}
    On the complete graph $K_N$ and for any $\tau\geq 0$, the $\tau$-Diffusion score is strongly equivalent to the multi-category Brier score.
\end{proposition}
\begin{proof}
    On the complete graph, the diffusion kernel of equation \ref{eqn:diffusion-kernel} simplifies to $H^\tau = \vec{v}_1\vec{v}_1^\top + e^{-\frac{\tau}{2}N}\sum_{j=2}^N\vec{v}_j\vec{v}_j^\top$. Recalling that $\sum_{j=1}^N\vec{v}_j\vec{v}_j^\top = \Id$ and $\vec{v}_1 = \frac{1}{\sqrt{N}}\vec{1}$, we obtain $H^\tau = \frac{1}{N}(1-e^{-\frac{\tau}{2}N})\vec{1}\vec{1}^\top + e^{-\frac{\tau}{2}N}\Id$. Observing again the fact that $\vec{1}^\top (\vec{p}-\vec{e}_i) = 0$, it follows that $S_{\mathrm{diff}}^\tau(\vec{p},i) = e^{-\frac{\tau}{2}N}(\vec{p}-\vec{e}_i)^\top (\vec{p}-\vec{e}_i) = e^{-\frac{\tau}{2}N}\mathrm{Br}(\vec{p},i)$.
\end{proof}

The interesting feature of the Diffusion score is its behaviour with respect to varying $\tau$, irrespective of the graph structure. It is most clearly illustrated on a disconnected graph:

\begin{proposition}
\label{prop:diff_score_asymptotics}
     Let $G$ be a graph with $K\geq 1$ connected components $C_1, \dots, C_K$.
     \begin{enumerate}
         \item 
         For small $\tau$, the $\tau$-diffusion score has the expansion
         \be S_{\mathrm{diff}}^\tau(\vec{p},i) = \mathrm{Br}(\vec{p}, i) - \tau S_{-L}^+(\vec{p}, i) + \mathcal{O}(\tau^2) \, .\ee
         \item As $\tau\to\infty$, the $\tau$-diffusion score asymptotically converges to
         \be S_{\mathrm{diff}}^\infty(\vec{p}, i) = \sum_{k=1}^{K} \frac{(P(C_k)-\boldone_{i\in C_k})^2}{|C_k|}\, ,\ee where $P(C)$ is the total probability mass placed on a subset of nodes $C\subseteq V$ by the distribution $\vec{p}$.
     \end{enumerate} 
\end{proposition}

\begin{proof}
1. For small $\tau$, we have the first-order Taylor expansion $H^\tau = \Id -\frac{\tau}{2}L+\mathcal{O}(\tau^2)$ and thus $S_{\mathrm{diff}}^\tau(\vec{p},i) = (\vec{p}-\vec{e}_i)^\top (\vec{p}-\vec{e}_i) - \frac{\tau}{2}(\vec{p}-\vec{e}_i)^\top L(\vec{p}-\vec{e}_i) + \mathcal{O}(\tau^2)= \mathrm{Br}(\vec{p}, i) - \tau S_{-L}^+(\vec{p}, i) + \mathcal{O}(\tau^2)$.

2. Consider the eigenspace of $L$ corresponding to eigenvalue $0$. It is $K$-dimensional and spanned by the orthonormal basis $\frac{1}{\sqrt{|C_1|}}\vec{1}_{C_1}, \dots, \frac{1}{\sqrt{|C_{K}|}}\vec{1}_{C_{K}}$, where $\mathbf{1}_C := \sum_{j\in C} \mathbf{e}_j$ for $C\subseteq V$. For $\tau\to \infty$ we thus obtain the singular value decomposition $H^\tau \to H^\infty = \sum_{k=1}^{K}\frac{1}{|C_k|}\vec{1}_{C_k}\vec{1}_{C_k}^\top $. Plugging this into \eqref{eqn:s_tau} yields $S_{\mathrm{diff}}^\infty(\vec{p}, i)= \sum_{k=1}^K \frac{1}{|C_k|} (\vec{p}-\vec{e}_i)^\top \vec{1}_{C_k}\vec{1}_{C_k}^\top(\vec{p}-\vec{e}_i)$. From $P(C) = \vec{1}_C^\top \vec{p}$ and $\vec{1}_C^\top\vec{e}_i  = \boldone_{i\in C}$, it follows that $(\vec{p}-\vec{e}_i)^\top \vec{1}_{C_k}\vec{1}_{C_k}^\top(\vec{p}-\vec{e}_i) = (P(C_k)-\boldone_{i\in C_k})^2$, which completes the proof.
\end{proof}

The limit $\tau\to\infty$ corresponds to a weighted version of the Brier score, where the categories are connected components rather than individual nodes, and the squared error for each component contributes with a weight proportional to the inverse of its size. In this limit, $\vec{p}$ scores well if it places the correct total mass $P(C_k)$ on component $C_k$ rather than necessarily getting the individual node probabilities $p_j$ right. Note that the diffusion score is no longer strictly proper in this limit, as any redistribution of mass within a connected component leaves the value of $S_{\mathrm{diff}}^\infty(\vec{p}, i)$ unaffected. Moreover, compared to the regular Brier score, $S_{\mathrm{diff}}^\infty(\vec{p}, i)$ down-weights larger connected components, preventing them from dominating the total score simply by virtue of their size. Consequently, there is no incentive to hedge by placing disproportionate probability mass on nodes belonging to smaller connected components.

Proposition \ref{prop:diff_score_asymptotics} provides us more generally with an interpretation of the Diffusion score as a function of $\tau$. For $\tau=0$ it reduces to the Brier score, which favours concentration of $\vec{p}$ exactly on node $i$, but otherwise prefers a uniform distribution. For small but non-zero $\tau$, the Laplacian score discussed in Section \ref{subsec:laplace_stepping_stone} enters the expansion with a \emph{negative} sign, reflecting a shift in preference: probability mass on the neighboring nodes is now encouraged, not just on the correct node itself. For larger $\tau$, the score increasingly favours concentration of $\vec{p}$ on the connected component which contains the true node $i$, but otherwise prefers a more uniform distribution of the probability mass between the different components. When using the diffusion score for large $\tau$, we take a deliberate decision to focus on predicting the correct connected component rather than exactly the right node. In this sense, the loss of strict propriety in the limit $\tau\to\infty$ is an expected feature of this score.

This idea extends further to the case where $G$ is connected: for large $\tau$ the asymptotic expression $S_{\mathrm{diff}}^\infty(\vec{p},i)$ remains approximately valid if we interpret $C_1\dots, C_K$ as the main clusters of $G$. The diffusion score for large $\tau$ thus favours predictions that are concentrated on the correct main cluster, without much consideration for the way the probability is distributed \emph{within} the cluster. But as \cite{Villegas:2023} demonstrate in their recent work on Laplacian coarse-graining, the diffusion kernel $H^\tau$ induces a natural scale-dependent clustering on the graph. In this interpretation, the number $K$ of clusters is itself dependent on $\tau$: the case $\tau=0$ represents maximal granularity in which each node is its own cluster ($K(\tau=0)=N$), while in the case $\tau\to\infty$ the entire (connected) graph becomes a single cluster ($\lim_{\tau\to\infty}K(\tau)=1$). In other words, the parameter $\tau$ in the Diffusion score acts as a control on the spectral scale on which to punish poor calibration. The scale-dependent coarse-graining of \citet{Villegas:2023} may provide a useful framework for a principled selection of $\tau$.

One detail is worth pointing out: for connected graphs, the Diffusion score as defined in \eqref{eqn:s_tau} converges to zero as $\tau\to\infty$, since the diffusion kernel converges to $\frac{1}{N}\vec{1}\vec{1}^\top$ which annihilates $\vec{p}-\vec{e}_i$. To compensate for this decay, let $\mu>0$ and set $H_\mu^\tau := e^{-\frac{\tau}{2}(L-\mu\Id)}$ which shifts the spectrum of the Laplacian by $\mu$. We can define the $\mu$-regularised $\tau$-\emph{Diffusion score} by
    \be\label{eqn:s_tau_regularised}
        S_{\mathrm{diff}}^{\tau,\mu}(\vec{p},i)
        := (\vec{p}-\vec{e}_i)^\top H_{\mu}^\tau(\vec{p}-\vec{e}_i) \, .
    \ee
Since $H_\mu^\tau = e^{\frac{\tau\mu}{2}}H^\tau$, the regularised score is simply $S_{\mathrm{diff}}^{\tau,\mu}(\vec{p},i) = e^{\frac{\tau\mu}{2}}S_{\mathrm{diff}}^\tau(\vec{p},i)$. Thus, for fixed $\tau$, regularisation amounts only to an overall rescaling and does not change the relative scores or their minimiser. In other words, the $\mu$-regularised $\tau$-Diffusion score is strongly equivalent to the original $\tau$-Diffusion score. For connected graphs, a natural a priori choice is $\mu=\lambda_2$, the Fiedler eigenvalue. In this case, the contribution of the slowest non-constant diffusion mode $\vec{v}_2$ is no longer exponentially damped.

\subsection{Traditional scores for node selection}
\label{subsec:traditional_scores}

\newcommand{\cmark}{$\checkmark$}
\newcommand{\xmark}{$\times$}
\begin{table}[b]
\centering
\caption{Evaluation metrics for node selection on general graphs.}
\label{tab:score_comparison_general}
\small
\begin{tabular}{lcccc}
\toprule
\textbf{Score} & \textbf{Graph-aware} & \textbf{Proper} & \textbf{Strictly proper} & \textbf{Parameters} \\
\midrule
Rank score                   & \xmark & \cmark & \xmark & — \\
Top-$K$ error                & \xmark & \cmark & \xmark & $K=1,2,3,\dots$ \\
Brier score                  & \xmark & \cmark & \cmark & — \\
Log-score                    & \xmark & \cmark & \cmark & — \\
Error distance               & \cmark & \xmark & \xmark & — \\
\textbf{Resistance score}    & \cmark & \cmark & \cmark & — \\
\textbf{Diffusion score}     & \cmark & \cmark & \cmark & $\tau > 0$ \\
\bottomrule
\end{tabular}
\end{table}

Before comparing our scoring rules on examples, we briefly review some of the scoring rules traditionally used for node selection problems. One class are scores based on a ranking of nodes by their assigned probability mass. Such a ranking of $V$ by $\vec{p}$ can be represented by a permutation $\sigma_{\vec{p}}\in \mathfrak{S}_N$, such that $p_{\sigma_{\vec{p}}^{-1}(1)}\geq \dots\geq p_{\sigma_{\vec{p}}^{-1}(N)}$. Note that $\sigma_{\vec{p}}$ is not unique if several entries of $\vec{p}$ are tied, but the choice of permutation in this case is irrelevant for the arguments that follow.

\begin{definition}
    Let $\sigma_{\vec{p}}\in \mathfrak{S}_N$ be a ranking of $V$ by $\vec{p}$.
    \begin{enumerate}
        \item The \emph{rank score} is defined as
        \be
            S_{\mathrm{rank}}(\vec{p}, i) = \sigma_{\vec{p}}(i) \qquad \in \lbrace 1,\dots , N\rbrace
        \ee
        \item The \emph{top-$K$ error} is defined as
        \be
            S_{\mathrm{top}K}(\vec{p}, i) = \boldone_{\sigma_{\vec{p}(i)} > K} \qquad \in \lbrace 0, 1\rbrace
        \ee
    \end{enumerate}
\end{definition}

As is the case for our graph-aware scores, smaller values indicate better performance. Alternatively, often one considers the \emph{inverse rank score} $\sigma_{\vec{p}}(i)^{-1}$ or the \emph{top-$K$ hit} $\boldone_{\sigma_{\vec{p}}(i)\leq K}$ instead. In Appendix~\ref{app:impropriety_proofs}, we show the following reassuring fact:

\begin{proposition}
    \label{prop:rank_based_scores_are_proper}
    Both the rank score and the top-$K$ error are proper scoring rules.
\end{proposition}

By the same token, the inverse rank score and the top-$K$ hit are proper scoring rules according to the opposite sign convention (higher values are better). However, all of these rank-based scoring rules completely ignore the graph-structure. Moreover, they are very far from strictly proper: altering $\vec{p}$ does not change the score as long as the ranking of the entries is preserved. A further weakness, specific to the top-$K$ error, is that it is binary and therefore meaningful only in aggregate, i.e., as an error rate over multiple examples.

Another often used evaluation metric (see, e.g., \citet{Antulov:2015,Zhu:2016,Jiang:2017,Paluch:2018}) is the Error distance. Assuming that $\vec{p}$ is uni-modal, consider its mode (i.e., the top-$1$ prediction) denoted by $m_{\vec{p}} = \arg\max_{j\in V} p_j$.

\begin{definition}
    \label{def:error_distance}
    The Error distance is defined as
    \be
        S_{\mathrm{ed}}(\vec{p},i) = d(m_{\vec{p}},i) \qquad \in \lbrace 1,\dots , ~\mathrm{diam}(G)\rbrace
    \ee
    where $d:V\times V\to \N$ denotes the shortest-path distance on $G$ and $\mathrm{diam}(G)$ the diameter of $G$.
\end{definition}

If the distribution is multi-modal, any tie-breaking mechanism or average over modes may be employed without affecting the following discussion.

In contrast to the rank-based scores, error distance does take the graph structure into account. However, by focusing solely on the mode of $\vec{p}$, the Error distance essentially ignores the probabilistic nature of the prediction or estimate it is supposed to evaluate. It is therefore not surprising that we find the following negative result:

\begin{proposition}
    \label{prop:error_distance_is_improper}
    The Error distance is not a proper scoring rule.
\end{proposition}

A simple counter-example is given in the Appendix~\ref{app:impropriety_proofs}. The Error distance thus lacks statistical justification and we do not recommend using it. Furthermore, we would like to point out that resorting to a $\vec{p}$-averaged version of the Error distance does not resolve the issue either, despite taking into account the full distribution:

\begin{proposition}             \label{prop:integrated_error_distance_is_improper}
    The average Error distance $S_{\mathrm{aed}}(\vec{p},i) = \sum_{j=1}^N p_j d(j,i)$ is not a proper scoring rule.
\end{proposition}

Again, a counter-example is provided in Appendix~\ref{app:impropriety_proofs}. Intuitively, however, the reason is clear: this score unduly rewards overconfident estimates.

In the literature on hierarchical classification, a closely related metric is the Hierarchical error distance, which we discuss in Section~\ref{subsec:hierarchical_classification} together with its impropriety. The counterexample of Appendix~\ref{app:impropriety_proofs} applies to it as well, using the same toy tree.

Table~\ref{tab:score_comparison_general} summarises the properties of the traditional evaluation metrics for node selection tasks alongside the proposed scores.

\subsection{Comparison of scores: a toy example}
\label{subsec:toy_comparison}

We illustrate the differences between the proposed scores and classical alternatives with a concrete example.

\begin{figure}[htb]
  \centering
  \includegraphics[width=0.49\textwidth]{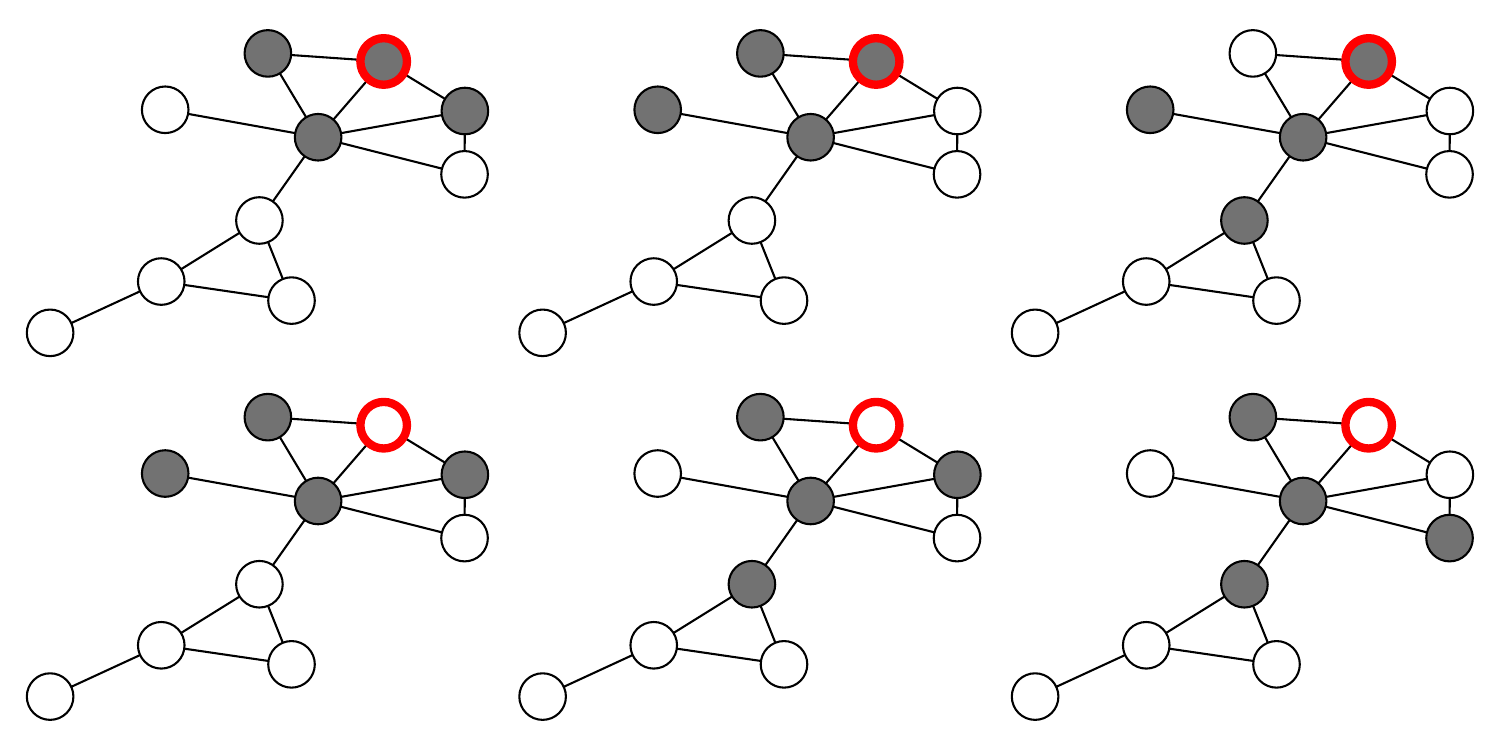}
  \includegraphics[width=0.5\textwidth]{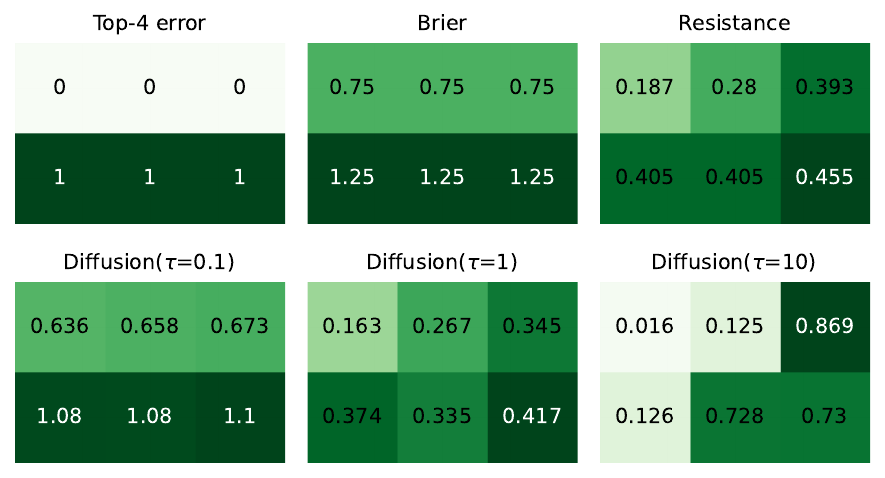}
  \caption{%
    \textbf{Left:}~Six different probabilistic node estimates on the same toy graph. The red circle indicates the true node~$i$; the gray shading indicates the probabilistic estimate $\vec{p}$ of~$i$, with probability mass~$0.25$ assigned to each gray node.
    \textbf{Right:}~Evaluation of these estimates by different scores. For each score, values are reported for the six estimates shown on the left, in matching order. Lower scores indicate better performance. The $\tau$-Diffusion score uses the regularisation with $\mu=\lambda_3$.
  }
  \label{fig:toy_graphs}
\end{figure}

The left panel of Figure~\ref{fig:toy_graphs} shows a fixed graph of ten nodes with a designated true node. We consider six probability distributions, each placing equal mass on four nodes, so that the support always consists of exactly four nodes. The top row of distributions includes the true node in the support, while the bottom row does not. Within each row, the three distributions differ in the distance of their mass from the true node: the leftmost concentrates mass near the true node, whereas the middle and rightmost place mass farther away.

\emph{Top-$K$ error.}
Since all four-element supports either include or exclude the true node, the top-$4$ error gives $0$ to all three top-row distributions and $1$ to all three bottom-row distributions. As a graph-blind metric, it is unable to distinguish any further: within each row the three distributions are seen as equally valid, despite differing substantially in their proximity to the true node.

\emph{Brier score.}
The Brier score is also unaware of the graph structure. It depends only on the probability mass at the true node and on the degree of uniformity of the distribution away from the true node. In our constrained toy example with four equal masses, the latter is fully determined by the former. Consequently, the Brier score also assigns lower scores to the top-row distributions than to the bottom-row ones, and fails to see any difference between the distributions within a row.

\emph{Resistance score.}
By contrast, the Resistance score is graph-aware. Like the two classical scores, it prefers the distributions in the top row to those in the bottom row. However, it also discriminates within each row, correctly identifying the distributions on the left as superior to those on the right.

\emph{Diffusion score.}
At small $\tau = 0.1$, the Diffusion score approximates the Brier score (Proposition~\ref{prop:diff_score_asymptotics}). Indeed, the scores within each row are nearly equal, with only small differences reflecting the graph structure. At $\tau = 1$ the graph-awareness increases and the pattern in this case is not unlike that of the Resistance score. At large $\tau = 10$, however, the score converges to a cluster-level Brier score (Proposition~\ref{prop:diff_score_asymptotics}) that reflects the macroscopic structure of the graph. It primarily distinguishes the two main clusters, favouring estimates that place all probability mass in the upper one, which contains the true node: top-left, top-middle, and bottom-left score substantially better than the remaining estimates, which place a quarter of their mass in the lower cluster. Within each of these two groups, the score further favours estimates that concentrate more of their probability mass in the dense upper-right sub-cluster containing the true node: top-left over top-middle and bottom-left in the first group, and bottom-middle and bottom-right over top-right in the second. In summary: the coarse-scale structure dominates the score, with finer-scale structure providing a secondary distinction. These examples illustrate how the parameter $\tau$ controls the spatial scale at which the Diffusion score operates.

\subsection{Scoring rules on rooted trees: the case of hierarchical classification}
\label{subsec:hierarchical_classification}

Hierarchical classification is a node selection task in which the graph is a rooted taxonomy tree $T = (V, E)$ whose leaf nodes $\tilde{V} \subset V$ correspond to the most fine-grained categories. In this section we apply the scores of the preceding sections to this setting and discuss their relationship to existing approaches.

Following~\citet{Silla:2011}, we distinguish two settings depending on where probability mass may be placed. In \emph{mandatory leaf-node prediction} (MLP), the standard setting in image classification, the classifier assigns all probability mass to leaf nodes: $\tilde{\vec{p}} \in \Delta^{\tilde{N}-1}$ is a distribution over $\tilde{V}$, embedded into $\Delta^{N-1}$ by assigning zero mass to internal nodes. In \emph{non-mandatory leaf-node prediction} (NMLP), mass may be placed on any node of $V$, including internal nodes.

\paragraph{LCA-height and Hierarchical error distance.}
 
A widely used hierarchy-aware evaluation metric is \emph{Hierarchical error distance}, introduced alongside the ImageNet benchmark \citep{Russakovsky:2015} and since adopted broadly.
 
\begin{definition}
\label{def:lca_height}
    Let $T = (V, E)$ be a rooted tree. For $i, j \in V$, let $\mathrm{LCA}(i, j)$ denote their lowest common ancestor and $\mathrm{height}(j)$ the length of the longest downward path from $j$ to any leaf in its subtree (leaves have height $0$). The \emph{Hierarchical error distance} is
    \be
        S_{\mathrm{LCA}}(\vec{p}, i)
        ~:=~
        \mathrm{height}\!\left(\mathrm{LCA}(m_{\vec{p}},\, i)\right),
    \ee
    where $m_{\vec{p}} = \arg\max_j p_j$ is the mode (i.e., top-$1$ prediction) of $\vec{p}$.
\end{definition}
 
Like the Error distance (Definition~\ref{def:error_distance}), of which it is the natural hierarchical analogue, $S_{\mathrm{LCA}}$ depends on $\vec{p}$ only through its mode $m_{\vec{p}}$ and is therefore not a proper scoring rule:
 
\begin{proposition}
\label{prop:lca_height_is_improper}
    $S_{\mathrm{LCA}}$ is not a proper scoring rule.
\end{proposition}

We refer to Appendix~\ref{app:impropriety_proofs} for a counterexample using the same tree and distributions as those for the Error distance (Propositions~\ref{prop:error_distance_is_improper} and~\ref{prop:integrated_error_distance_is_improper}).

Several papers \citep{Russakovsky:2015,Bertinetto:2020,Karthik:2021} also use the average Hierarchical error distance from each of the top-$K$ classes. This metric is denoted distance@$K$ in \citet{Karthik:2021} and reduces to $S_{\mathrm{LCA}}$ for $K=1$. However, just like in the case of the average Error distance, simple averaging does not make the fundamental deficiencies of the metric go away. Indeed, \citet{Karthik:2021} point out that distance@$K$ can be improved at no cost: replacing the ranking based on maximising $p_j$ with the ranking based on minimising $\sum_i p_i\,\mathrm{height}(\mathrm{LCA}(i,j))$ (conditional risk minimisation) reduces distance@$K$ without retraining. By contrast, a strictly proper scoring rule such as $S_{\mathrm{res}}$ cannot be manipulated or ``gamed'' in that way.

\paragraph{The Resistance score as a proper alternative.}

By Proposition~\ref{prop:resistance_equals_shortest_path_on_trees}, the resistance distance on any tree equals the shortest-path distance, so the Resistance score~\eqref{eqn:s_omega2} takes the explicit form
\be
    \label{eqn:resistance_score_tree}
    S_{\mathrm{res}}(\vec{p}, i)
    ~=~
    \sum_{j \in V} p_j\, d(i,j)
    ~-~
    \frac{1}{2}\sum_{j,k \in V} p_j\, d(j,k)\, p_k \,,
\ee
where $d$ denotes the shortest-path distance in the taxonomy tree. The first term measures the expected taxonomic distance of $\vec{p}$ from the true class $i$; the second term is the variance of $\vec{p}$ under the tree distance metric in the sense of~\citet{Devriendt:2022}.
 
Since $d(i,j) = \mathrm{depth}(i) + \mathrm{depth}(j) - 2\,\mathrm{depth}(\mathrm{LCA}(i,j))$ for any rooted tree, where $\mathrm{depth}(j)=d(r,j)$ denotes the distance from the root $r$ to $j$, substituting into~\eqref{eqn:resistance_score_tree} and expanding yields
\be
    \label{eqn:resistance_score_lca}
    S_{\mathrm{res}}(\vec{p}, i)
    ~=~
    \mathrm{depth}(i)
    ~-~ 2\,\E_{j \sim \vec{p}}\bigl[\mathrm{depth}(\mathrm{LCA}(i,j))\bigr]
    ~+~ \E_{j,k \sim \vec{p}}\bigl[\mathrm{depth}(\mathrm{LCA}(j,k))\bigr]\,.
\ee
Equation \ref{eqn:resistance_score_lca} expresses the Resistance score purely in terms of LCA depths: its calibration term rewards distributions that concentrate probability mass on classes sharing a deep LCA with the true class $i$, while the entropy term rewards a distribution if two independent draws from it tend to share a shallow LCA.
 
On a balanced tree (where all leaves have the same depth $\mathrm{d}$), the identity $\mathrm{height}(j) = \mathrm{d} - \mathrm{depth}(j)$ holds for all nodes $j$, so~\eqref{eqn:resistance_score_lca} further simplifies to
\be
    \label{eqn:resistance_score_balanced}
    S_{\mathrm{res}}(\vec{p}, i)
    ~=~
    2\,\E_{j \sim \vec{p}}\bigl[\mathrm{height}(\mathrm{LCA}(i,j))\bigr]
    ~-~ \E_{j,k \sim \vec{p}}\bigl[\mathrm{height}(\mathrm{LCA}(j,k))\bigr]
    ~-~\mathrm{height}(i)
    \,,
\ee
which is purely in terms of LCA heights. Given that $i = \mathrm{LCA}(i,i)$, this is exactly twice the kernel score of \eqref{eqn:kernel-score-finite-+} with kernel $k(i,j) = \mathrm{height}(\mathrm{LCA}(i,j))$. This observation makes it evident that the Resistance score is the natural way to turn the Hierarchical error distance into a proper scoring rule. It is obtained by replacing mode-evaluation with a carefully balanced distribution-averaging.

\paragraph{The log-score family.}

The resistance score is not the only strictly proper scoring rule for hierarchical classification in the MLP setting. \citet{Bertinetto:2020} introduce the \emph{hierarchical cross-entropy} (HXE), a hierarchy-aware training loss generalising the log-score, apparently unaware that it constitutes a proper scoring rule. As a consequence, neither \citet{Bertinetto:2020} nor \citet{Karthik:2021} apply HXE directly as an evaluation metric, which would have sidestepped the gameability of $S_{\mathrm{LCA}}$ entirely. \citet{Urbani:2024} generalise HXE and prove propriety. Below we extend their result to strict propriety.

For a node $j \in V$ denote by $\mathrm{par}(j)$ its parent in the rooted tree, by $\mathrm{Anc}(j)\subset V$ the closed set of its ancestors (i.e., including $j$ itself), by $\tilde{V}(j) = \{k \in \tilde{V} ~|~ j \in \mathrm{Anc}(k)\rbrace$ the set of leaves in its subtree, and by $P_j(\tilde{\vec{p}}) := \sum_{k \in \tilde{V}(j)} \tilde{p}_k$ the total probability mass on that subtree.

The \emph{hierarchical cross-entropy} (HXE) is defined for $i \in \tilde{V}$ by
\be
    \label{eqn:hxe}
    S_{\mathrm{HXE}}^\alpha(\tilde{\vec{p}},\, i)
    ~=~
    -\!\sum_{j \in \mathrm{Anc}(i)\setminus \lbrace r\rbrace}
    e^{-\alpha\,\cdot\,\mathrm{depth}(j)}
    \log\frac{P_j(\tilde{\vec{p}})}{P_{\mathrm{par}(j)}(\tilde{\vec{p}})}\,,
\ee
where $\alpha \geq 0$ is a free parameter. At $\alpha = 0$ all weights equal $1$ and the log-ratios telescope to $\log \tilde{p}_i$, recovering the standard log-score. For $\alpha > 0$, the exponential factor down-weights fine-grained (deep) terms and up-weights coarse (near-root) terms, so that the score penalises taxonomically distant errors more heavily. The parameter $\alpha$ plays a similar role as $\tau$ in the Diffusion score (Section~\ref{subsec:diffusion_score}): both interpolate between a score over independent classes (the log-score for $\alpha=0$, and the Brier score for $\tau=0$, respectively) and a progressively more coarse-grained, hierarchy- or graph-aware score.
 
\citet{Urbani:2024} generalise HXE to the family
\be
    \label{eqn:urbani_st}
    S_{\mathrm{Urb}}(\tilde{\vec{p}},\, i)
    ~=~
    -\!\sum_{j \in \mathrm{Anc}(i)\setminus \lbrace r\rbrace} w_j\, \log P_j(\tilde{\vec{p}})
\ee
with arbitrary positive weights $w_j > 0$, using log-marginals rather than log-ratios. Expanding the ratios in~\eqref{eqn:hxe} shows that $S_{\mathrm{HXE}}^\alpha$ is a special case of~\eqref{eqn:urbani_st} with weights $w_j = e^{-\alpha\,\cdot\,\mathrm{depth}(j)}(1-e^{-\alpha})$ for internal nodes $j \in \mathrm{Anc}(i) \setminus \{i, r\}$ and $w_i = e^{-\alpha\,\cdot\,\mathrm{depth}(i)}$ for the leaf $i$. \citet{Urbani:2024} prove that $S_{\mathrm{Urb}}$ is proper when the weights are \emph{balanced}, which means that $\sum_{j \in \mathrm{Anc}(k)\setminus \lbrace r\rbrace} w_j$ is the same constant $C$ for all leaves $k \in \tilde{V}$. In fact, their proof readily extends to strict propriety:
 
\begin{proposition}
\label{prop:lt_strictly_proper}
    $S_{\mathrm{Urb}}$ is strictly proper whenever the weights are balanced.
\end{proposition}
\begin{proof}
    Let $\tilde{\vec{q}}\in\Delta^{\tilde{N}-1}$ be a distribution over the leaf nodes and $\tilde{\vec{f}}\in\lbrace \tilde{\vec{v}}\in\R^{\tilde{N}} ~|~ P_j(\tilde{\vec{v}}) > 0 ~\forall j\in V\rbrace\supset \Delta^{\tilde{N}-1}$ a more general vector. The expectation
    \begin{align*}
        E(\tilde{\vec{f}}) := \mathbb{E}_{i\sim\tilde{\vec{q}}}[S_{\mathrm{Urb}}(\tilde{\vec{f}}, i)]
        &~=~ -\sum_{i\in\tilde{V}}\tilde{q}_i\sum_{j\in\mathrm{Anc}(i)\setminus\{r\}} w_j\log P_j(\tilde{\vec{f}}) \\
        &~=~ - \sum_{j\in V\setminus\{r\}}\sum_{i\in\tilde{V}(j)}\tilde{q}_i\, w_j\log P_j(\tilde{\vec{f}}) \\
        &~=~ - \sum_{j\in V\setminus\{r\}}P_j(\tilde{\vec{q}})\, w_j\log P_j(\tilde{\vec{f}})
    \end{align*}
    is strictly convex in $\tilde{\vec{f}}$, and so is its restriction to $\Delta^{\tilde{N}-1}$. It is not hard to see that the gradient $\nabla_{\tilde{\vec{f}}}E(\tilde{\vec{f}})$ at $\tilde{\vec{f}} = \tilde{\vec{q}}$ has $k$-th component $\sum_{j \in \mathrm{Anc}(k)\setminus \lbrace r\rbrace} w_j$, which is independent of $k$ under the balancedness condition. The gradient is therefore proportional to $\vec{1}$, which is orthogonal to $\Delta^{\tilde{N}-1}$. Its projection to $\Delta^{\tilde{N}-1}$ is thus zero, which, together with strict convexity, implies that $\tilde{\vec{q}}$ is the unique minimiser of $E$ in $\Delta^{\tilde{N}-1}$. But this is the same as strict propriety of $S_{\mathrm{Urb}}$.
\end{proof}
 
For HXE, the weights satisfy $\sum_{j \in \mathrm{Anc}(k)\setminus\{r\}} w_j = e^{-\alpha}$ for every leaf $k \in \tilde{V}$, independently of $\mathrm{depth}(k)$ (the sum telescopes), so the balancedness condition holds on any tree.
 
\begin{corollary}
\label{cor:hxe_strictly_proper}
    $S_{\mathrm{HXE}}^\alpha$ is strictly proper for every $\alpha \geq 0$.
\end{corollary}
 
Both $S_{\mathrm{Urb}}$ and $S_{\mathrm{HXE}}^\alpha$ are defined for MLP, and there is no obvious extension to NMLP. The balancedness condition cannot hold simultaneously for all nodes $k$ beyond just the leaves, and the gradient argument breaks down.

The following observation is trivial:

\begin{proposition}
\label{prop:star_graph_reduction}
    On a rooted tree of depth 1 (star graph) with $N$ nodes and $\tilde{N} = N-1$ leaves, the Resistance score in the MLP setting is strongly equivalent to the Brier score with $\tilde{N}$ classes, while $S_{\mathrm{Urb}}$ (and therefore also HXE) is strongly equivalent to the logarithmic score.
\end{proposition}

In summary, the Resistance score and the HXE are both strictly proper and hierarchy-aware, but generalise different hierarchy-blind scores: the Resistance score generalises the Brier score (and the RPS), while HXE and the Urbani score generalise the logarithmic score. The Resistance score additionally applies to both MLP and NMLP and requires no free parameter. We compare the two families empirically in Section~\ref{sec:imagenet}. Table~\ref{tab:score_comparison_tree} gives an overview of hierarchy-aware metrics for single-label hierarchical classification.

\begin{table}[t]
\centering
\caption{Hierarchy-aware evaluation metrics for (single-label) hierarchical classification. All metrics are applicable to the MLP setting, and the \emph{NMLP} column specifies whether each metric can be applied meaningfully to the NMLP setting as well.}
\label{tab:score_comparison_tree}
\small
\begin{tabular}{llcccc}
\toprule
\textbf{Score} & \textbf{Reference} & \textbf{Proper} & \textbf{\shortstack{Strictly\\proper}} & \textbf{NMLP} & \textbf{Parameters} \\
\midrule
Wu--Palmer similarity                                     & \citet{WuPalmer:1994}    & \xmark & \xmark & \cmark & — \\
Gen. tree distance loss                                  & \citet{Cao:2024}         & \xmark & \xmark & \cmark & $c \geq 0$ \\
\quad Tree distance loss ($c=0$)                          & \citet{Ramaswamy:2015}   & \xmark & \xmark & \cmark & — \\
distance@$K$                                              & \citet{Karthik:2021}     & \xmark & \xmark & \cmark & $K=1,2,3,\dots$ \\
\quad Hier.\ error dist. ($K=1$)                       & \citet{Russakovsky:2015} & \xmark & \xmark & \cmark & — \\
Urbani family                                             & \citet{Urbani:2024}      & \cmark & \cmark & \xmark      & $w_j > 0$ balanced \\
\quad HXE ($w_j \propto e^{-\alpha\,\mathrm{depth}(j)}$) & \citet{Bertinetto:2020}  & \cmark & \cmark & \xmark      & $\alpha \geq 0$ \\
\textbf{Resistance score}                          &        [ours]                   & \cmark & \cmark & \cmark & — \\
\bottomrule
\end{tabular}
\end{table}

\subsection{Efficient computation}
\label{subsec:efficient_computation}

Both the Resistance score and the Diffusion score are defined in terms of the graph Laplacian $L$, and their naive computation via dense matrix operations is prohibitively expensive for large graphs. In this section, we briefly describe efficient alternatives.

\paragraph{Resistance score.}
The naive computation of $S_{\mathrm{res}}$ requires the Moore-Penrose pseudoinverse $L^\dagger$, which costs $\mathcal{O}(N^3)$ via SVD (or $\mathcal{O}(N^\eta)$ with $\eta$ close to $3$ when using fast matrix multiplication). However, recall from~\eqref{eqn:s_omega1} the quadratic form of the Resistance score, $S_{\mathrm{res}}(\vec{p}, i) = (\vec{p}-\vec{e}_i)^\top L^\dagger (\vec{p}-\vec{e}_i)$. The key observation is that $\vec{v} = \vec{p} - \vec{e}_i \in \vec{1}^\perp$. For a connected graph, $L$ restricted to $\vec{1}^\perp$ is invertible, and its inverse coincides with the restriction of $L^\dagger$. Thus, $L^\dagger\vec{v}$ can be obtained by solving $L\vec{x}=\vec{v}$ on the subspace $\vec{1}^\perp$. The Resistance score can therefore be computed as the dot product $\vec{v}^\top\vec{x}$, without ever forming $L^\dagger$ explicitly. \emph{Laplacian solvers} \citep{Spielman:2004} solve such systems approximately in nearly linear time in $|E|$. Our fast implementation uses the \textsc{ApproxChol} solver from the \textsc{Laplacians.jl} Julia package \citep{Gao:2023}, called from Python via \textsc{juliacall}. \citet{Coscia:2023} demonstrate on a range of real-world networks and representative synthetic network models that the approximation error is negligible for practical purposes, while empirically the solver scales as $\mathcal{O}(N^\eta)$ with $\eta\in[1.1, 1.4]$, compared to $\eta\in [2.6,2.9]$ for the SVD-based baseline. This translates to a runtime advantage over the exact SVD baseline of several orders of magnitude for large and reasonably sparse graphs.

\paragraph{Diffusion score.}
The naive computation of $S_{\mathrm{diff}}^\tau$ requires forming the full $N\times N$ matrix exponential $e^{-\frac{\tau}{2}L}$, which costs $\mathcal{O}(N^3)$. Since $S_{\mathrm{diff}}^\tau(\vec{p},i) = (\vec{p}-\vec{e}_i)^\top e^{-\frac{\tau}{2}L} (\vec{p}-\vec{e}_i)$, it suffices to compute the matrix-exponential-vector product $e^{-\frac{\tau}{2}L}\vec{v}$ for $\vec{v} = \vec{p} - \vec{e}_i$. This can be done efficiently using Krylov subspace methods \citep{Saad:1992}, such as the \texttt{expm\_multiply} routine \citep{Al-Mohy:2011} of \textsc{SciPy}. It approximates $e^{-\frac{\tau}{2}L}\vec{v}$ by projecting onto a low-dimensional Krylov subspace $\mathrm{span}\{\vec{v}, L\vec{v}, L^2\vec{v}, \dots, L^{k-1}\vec{v}\}$, requiring only sparse matrix-vector products with $L$. The cost is $\mathcal{O}(|E| \cdot k)$ where $k$ is the Krylov dimension, which is chosen adaptively by the algorithm based on the spectral radius of $\frac{\tau}{2}L$ and a target accuracy. In practice, it tends to be small relative to $N$, although it grows with $\tau$.

\section{A canonical use case: source detection}
\label{sec:source_detection}

As a first concrete example of a node selection task, we consider the source detection problem. Consider a network $G = (V,E)$ whose nodes $V$ represent individuals and whose edges $E$ represent contacts between them. An epidemic outbreak takes place on this network, with the disease spreading along the edges. At any given time, nodes occupy different states with respect to the disease, for example \emph{susceptible} or \emph{infected}. Denote by $\mathcal{M}$ the stochastic process governing the disease dynamics, such as a simple Susceptible-Infected (SI) model characterised by a fixed transmission probability per unit time. For simplicity, we assume the outbreak began at time $t=0$ from a single unknown source node $i_0\in V$.

\begin{figure}[htbp]
\centering
\includegraphics[width=\textwidth]{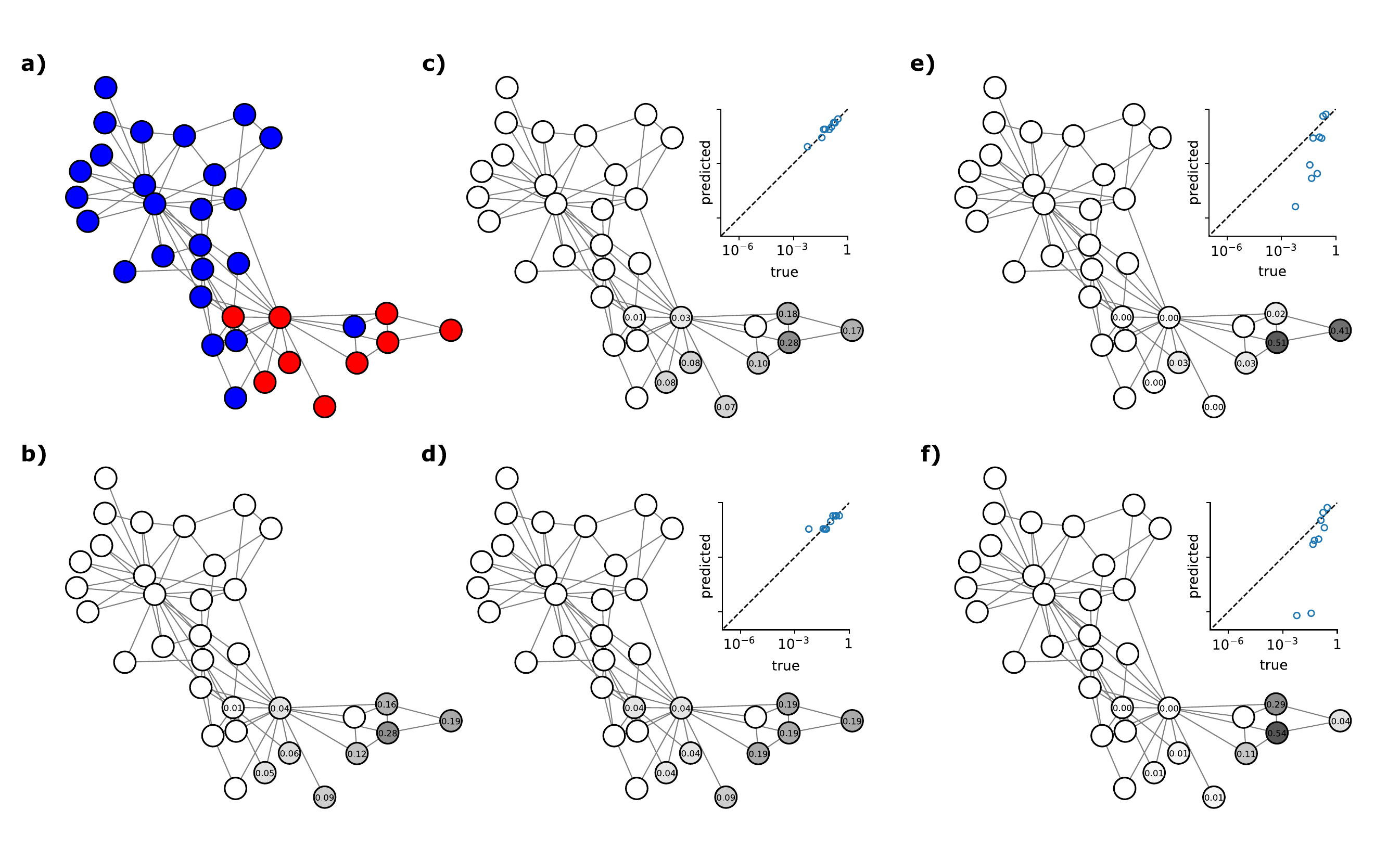}
\caption{An instance of the source detection problem on the Karate Club network.
\textbf{(a)}~Infection status at time $T$: red nodes are infected, blue nodes are susceptible.
\textbf{(b)}~True conditional source distribution $\vec{q}$, computed by ABC with zero-tolerance rejection sampling.
\textbf{(c)--(f)}~Four different probabilistic estimates $\vec{p}$ of the source node $i_0$. Node shading represents estimated probability. The scatter plots show the estimated probabilities against the true probabilities from (b), whereby each dot represents a node. A perfectly calibrated estimator would place all points on the diagonal.}
\label{fig:source_detection_snapshots}
\end{figure}

Suppose that we are given the vector of observed states of all nodes at time $t=T$, denoted by $\vec{O}_T$. Figure~\ref{fig:source_detection_snapshots}a shows an example of such a snapshot, taken from a simulated outbreak on the well-known Zachary's Karate Club network~\citep{Zachary:1977}. The task is to identify the most likely source of the outbreak from this snapshot alone. 

Naturally, the source is not fully determined by $\vec{O}_T$. Rather, the problem can be framed as estimating the true conditional distribution $q_i = \Pr_{\mathcal{M}}(i_0=i\mid\vec{O}_T)$ over potential source nodes $i\in V$, where the probability is taken over the joint distribution induced by a uniformly random source node $i_0 \in V$ and the subsequent stochastic dynamics $\mathcal{M}$. Due to the inverse nature of the problem and the inherent combinatorial complexity of stochastic processes on networks, this distribution $\vec{q}=(q_1,\dots, q_N)^\top$ is generally intractable. For the small network in our example, however, we can estimate it numerically almost exactly by Approximate Bayesian computation (ABC) with a zero-tolerance criterion in the rejection sampler: we accept only those simulation runs that produce exactly the observed snapshot $\vec{O}_T$ at time $t=T$, which are 852 in total. Figure~\ref{fig:source_detection_snapshots}b depicts the resulting distribution $\vec{q}$.

\begin{figure}[htbp]
\centering
\includegraphics[width=\textwidth]{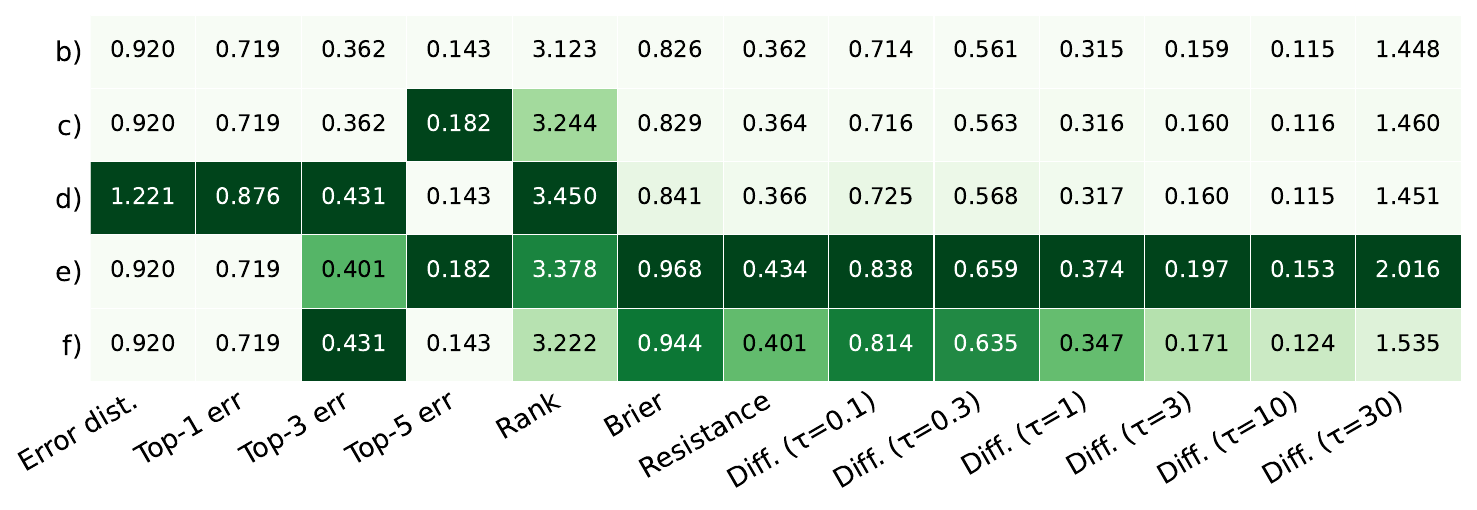}
\caption{Comparison of scores on the four source estimators (c)--(f) against the true distribution (b), averaged over realisations $i_0\sim\vec{q}$. Lower is better for all scores. Colour represents performance normalised within each column (darkest: worst model). The $\tau$-Diffusion score uses the regularisation with $\mu=0.8$.}
\label{fig:model_comparison}
\end{figure}

Suppose now that we have a model that issues a probabilistic estimate $\vec{p}$ of the source node $i_0$. Figures~\ref{fig:source_detection_snapshots}c--f show four such estimates. Three are derived from published source detection models (see Appendix~\ref{app:source_detection} for details); the fourth estimate is a piecewise-uniform heuristic added for illustrative purposes. The four estimators are not meant as a representative benchmark of source detection methods. Figure~\ref{fig:source_detection_snapshots} also shows calibration plots, which compare the estimated probabilities $\vec{p}$ to the true probabilities $\vec{q}$ for each model. These leave no doubt that the estimates in panels~(c) and~(d) are superior to those in~(e) and~(f), with~(c) likely the best.

However, in real scenarios without access to the entire distribution $\vec{q}$, the estimate $\vec{p}$ can only be judged on the basis of the single ground-truth source $i_0$ using a scoring rule. Figure~\ref{fig:model_comparison} shows how each probabilistic estimate stacks up under a range of scoring rules, averaged over all 852 realisations $i_0\sim \vec{q}$.

The results reveal a number of troubling patterns among the traditional scores. First and most importantly, they are sometimes rather off-mark: estimator~(f), for instance, consistently receives the same or better scores than~(d), even though visual inspection and the calibration plots confirm it is clearly inferior. Second, some scores show very limited ability to discriminate between the four estimates: for example, the Error distance rates estimators~(c), (e) and (f) identically. Third, the traditional scores disagree substantially with one another about the ranking of the estimates.

The main reason for this behaviour is that estimators~(e) and (f) are both overconfident. Their mode coincides with the mode of the true distribution~(b), but their distributions are significantly more peaked. All strictly proper scores (including the graph-unaware Brier score) detect and penalise this excess sharpness: (e) and~(f) score worse than (c) and~(d) on every strictly proper score. The top-$K$ error and Error distance, which depend only on the mode or top-$K$ of the distribution, are blind to this.

The graph-aware strictly proper scores detect additional structure beyond sharpness. For example, graph-aware scores clearly prefer~(f) over (e), while the difference in Brier score is less significant. Moreover, estimate~(d) achieves the best Diffusion score at $\tau = 30$ among all non-true distributions. This distribution was handcrafted to place exactly the right total mass on each of the three main clusters of the infected subgraph but distributes it uniformly within these clusters. The example shows the sensitivity of the Diffusion score at large $\tau$ to correct cluster-level mass assignment.

Our example illustrates the limitations of classical scores vis-a-vis the new graph-aware and strictly proper scores. Note again that all scores in Figure~\ref{fig:model_comparison} are averaged over the true distribution $\vec{q}$, which should already have washed out many of the more egregious artefacts that may appear when evaluating on a single realisation $i_0\sim\vec{q}$. Any real epidemic source detection scenario yields exactly one ground truth per outbreak, making the choice of a strictly proper, graph-aware score all the more consequential.

\section{Application to hierarchical classification: ImageNet}
\label{sec:imagenet}

As a second concrete illustration of the node selection framework, we apply our scoring rules to the evaluation of image classifiers on the ImageNet-1k benchmark~\citep{Russakovsky:2015}. In this setting, the graph is the WordNet taxonomy pruned to a tree $T$, the leaf nodes $\tilde{V}$ are the 1{,}000 fine-grained ImageNet classes, and the node selection task is: given an image, identify its correct leaf label. A classifier outputs a distribution $\tilde{\vec{p}}\in\Delta^{\tilde{N}-1}$ over $\tilde{V}$, and the true label is a single leaf $i\in\tilde{V}$. This is the mandatory leaf-node prediction (MLP) setting of Section~\ref{subsec:hierarchical_classification}.

\subsection{Setup}
\label{subsec:imagenet_setup}

\paragraph{Label graph.}
Starting from the WordNet hypernym graph, we take the subgraph made up of the 1{,}000 ImageNet leaf synsets and all their ancestors, yielding a directed acyclic graph (DAG) of 1{,}860 nodes and 1{,}937 edges. The DAG structure arises because a small percentage of classes are part of multiple superclasses. Following~\citet{Bertinetto:2020} we then prune the DAG to a tree: for each leaf, we retain the longest path to the root, breaking ties in favour of paths that introduce the fewest new nodes. This yields a tree of 1{,}816 nodes, on which the HXE of \citet{Bertinetto:2020} is well-defined and the Resistance score, according to Proposition~\ref{prop:resistance_equals_shortest_path_on_trees}, has the explicit form~\eqref{eqn:resistance_score_tree}.

\paragraph{Classifiers.}
We evaluate ten pretrained classifiers spanning a wide range of architectural families and accuracy levels. Four models represent architecturally distinct approaches at similar top-1 error rates (18--22\%): ConvNeXt-Tiny~\citep{Liu:2022} (modern CNN), Swin-Tiny~\citep{Liu:2021} (hierarchical Vision Transformer), ViT-S/16~\citep{Dosovitskiy:2021} (pure Vision Transformer), and EfficientNet-B2~\citep{Tan:2019} (compound-scaled CNN). Six further models of varying top-1 error rates (19--33\%) provide a broader comparison baseline: ResNet-50 and ResNet-18~\citep{He:2016}, EfficientNet-B0~\citep{Tan:2019}, MobileNetV3-Large and MobileNetV3-Small~\citep{Howard:2019}, and ShuffleNet-V2~\citep{Ma:2018}. All models use publicly available pretrained weights. We run inference on the full ImageNet-1k validation set (50{,}000 images) and save the softmax probability vectors.

\paragraph{Metrics.}
We report top-1 and top-5 error, Hierarchical error distance@1 and distance@5 (improper, mode-dependent), log-loss, HXE, Brier score and the Resistance score. For HXE we use $\alpha = 0.5$, the mid-range value that \citet{Bertinetto:2020} report as best-performing on ImageNet when using HXE as a training loss.

\subsection{Aggregate comparison}
\label{subsec:imagenet_aggregate}

\begin{figure}[htbp]
    \centering
    \includegraphics[width=\linewidth]{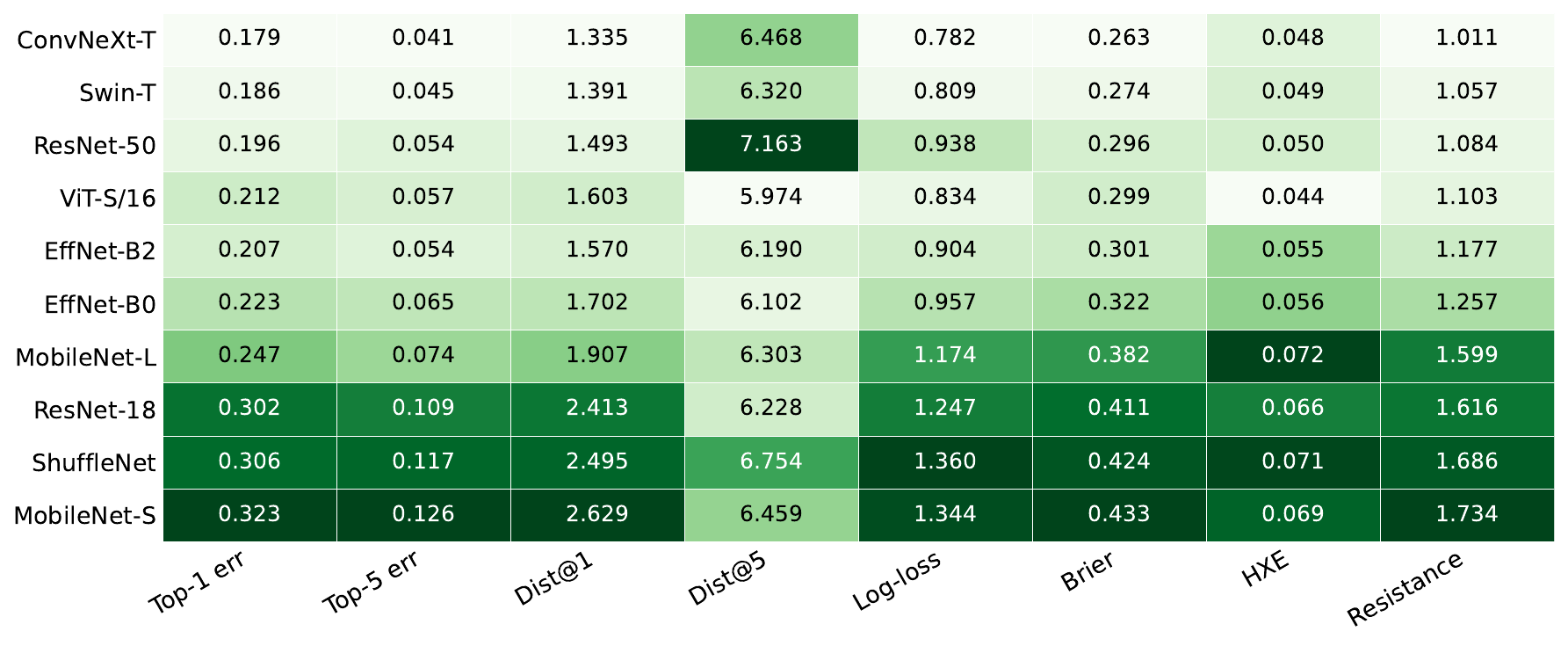}
    \caption{Comparison of ten ImageNet classifiers, ordered by their Resistance Score. All metrics are averaged over 50{,}000 validation images. Lower is better throughout (top-1 and top-5 reported as error rates).}
    \label{fig:imagenet_table}
\end{figure}

Figure~\ref{fig:imagenet_table} shows the full comparison. The aggregate model ranking produced by the Resistance score aligns closely with top-$1$ error rate, Brier score, and log-loss (Spearman rank correlations of 0.99, 0.99, and 0.94, respectively). Hierarchical error distance@1 agrees perfectly with top-$1$ error as well. The hierarchical error distance@5, by contrast, is essentially uncorrelated with top-1 error (Spearman rank correlation $-0.07$), underlining our doubts about distance@$K$ as a suitable hierarchy-aware metric. Meanwhile, HXE correlates reasonably well with top-$1$ error (Spearman rank correlation $0.79$) and the Resistance score, but shows some notable disagreements, indicating that it captures a subtly different notion of hierarchical quality than the Resistance score.

The most obvious anomaly is ResNet-50, which ranks third by top-1 error rate and Resistance score but last by distance@5, worse than models with top-$1$ error rates 12 percentage points higher. ResNet-type models trained with standard cross-entropy are known to be systematically overconfident \citep{Guo:2017}: they concentrate too much probability mass on the top-1 prediction. The remaining mass is spread thinly across the taxonomy with little consideration of the hierarchical structure, so the top-5 predictions land far from the true class on average. Proper scores are less affected because they weigh predictions by their probability: the low mass outside the top-1 class contributes little to either the calibration or the entropy term.

Another interesting comparison is ViT-S/16 versus EfficientNet-B2. EfficientNet-B2 has a marginally lower top-1 error rate and distance@1, yet all four proper scores favour ViT-S/16, indicating better distributions overall. Similarly, ConvNeXt-Tiny beats Swin-Tiny on all proper scores and on distance@1, but distance@5 gives the opposite verdict. This again underscores the fact that an improper score such as Hierarchical error distance@$K$ should not be trusted for evaluation.

\begin{figure}[htbp]
    \centering
    \includegraphics[width=\textwidth]{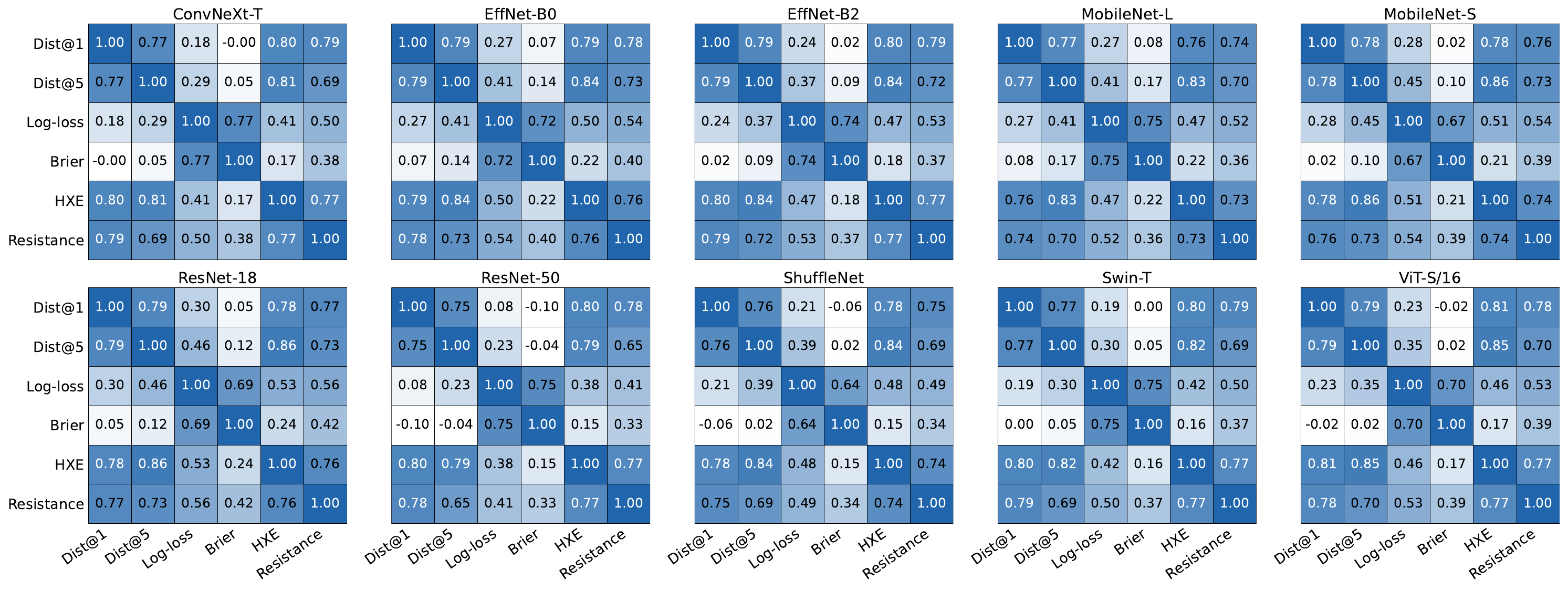}
    \caption{Spearman correlation of different scores, averaged over all images that are misclassified by the respective model.}
    \label{fig:spearman_per_model}
\end{figure}

At a more granular level, we examine how the metrics behave in cases where models struggle to identify the correct class. For each individual model, we compute Spearman correlations between metrics on all misclassified images (wrong top-1). Figure~\ref{fig:spearman_per_model} reports the results, with all models exhibiting very similar patterns. First, the Brier score is essentially uncorrelated with distance@1 and is only weakly correlated with distance@5 and HXE. This illustrates at the image level that the Brier score and the hierarchy-aware metrics measure orthogonal aspects of prediction quality: the Brier score prefers a uniform distribution over the wrong classes, while distance@$K$ and HXE favour hierarchical proximity to the true class. Second, log-loss sits somewhere in between: it correlates most clearly with Brier, since both are sensitive to the probability assigned to the true class, and more moderately with Resistance, HXE, distance@5 and distance@1. Third, the Resistance score correlates with HXE, but not perfectly. This shows that the two scores are complementary rather than redundant, even though both are hierarchy-aware and strictly proper.

\subsection{Qualitative examples}
\label{subsec:imagenet_qualitative}

\begin{figure}[tp]
    \begin{overpic}[width=\textwidth, percent]{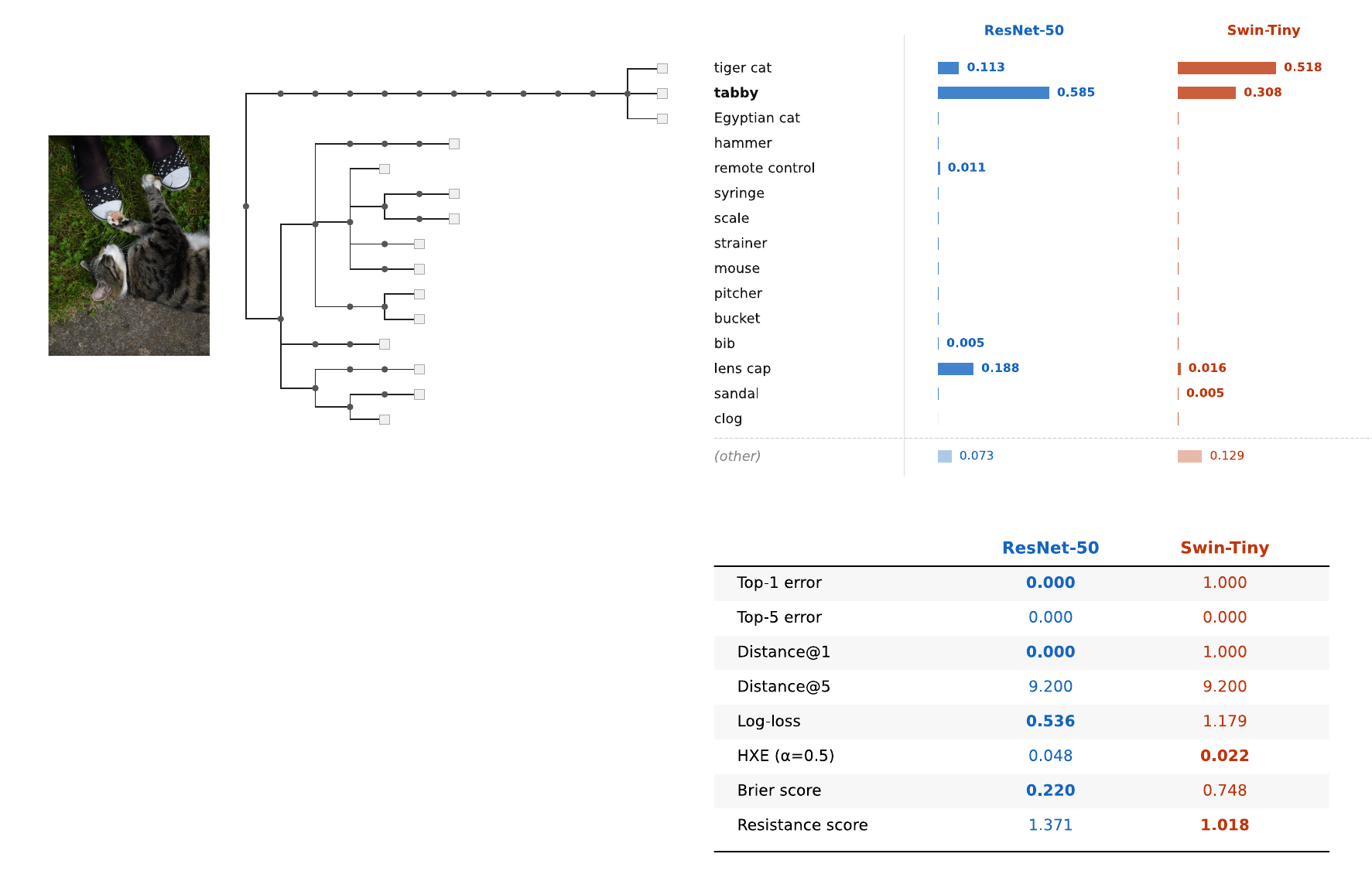}
        \put(0, 0){%
            \raisebox{0pt}[0pt][0pt]{%
                \parbox[b]{0.45\textwidth}{%
                    \captionsetup{type=figure, singlelinecheck=false}
                    \caption{%
                        ResNet-50 versus Swin-Tiny on image \#16763 (ground truth: \emph{tabby}). \textbf{Top:} Output distributions of both models on the 15 classes with the highest assigned probability, together with the induced subtree of the WordNet taxonomy. The remaining mass is attributed to `other'. \textbf{Right:} Comparisons of different scores. Hierarchy-aware scores favour Swin-Tiny's concentration of probability mass within the cat subtree.
                    }
                    \label{fig:dendro_hedging}
                }%
            }%
        }
    \end{overpic}

    \hspace{4cm}

    \begin{overpic}[width=\textwidth, percent]{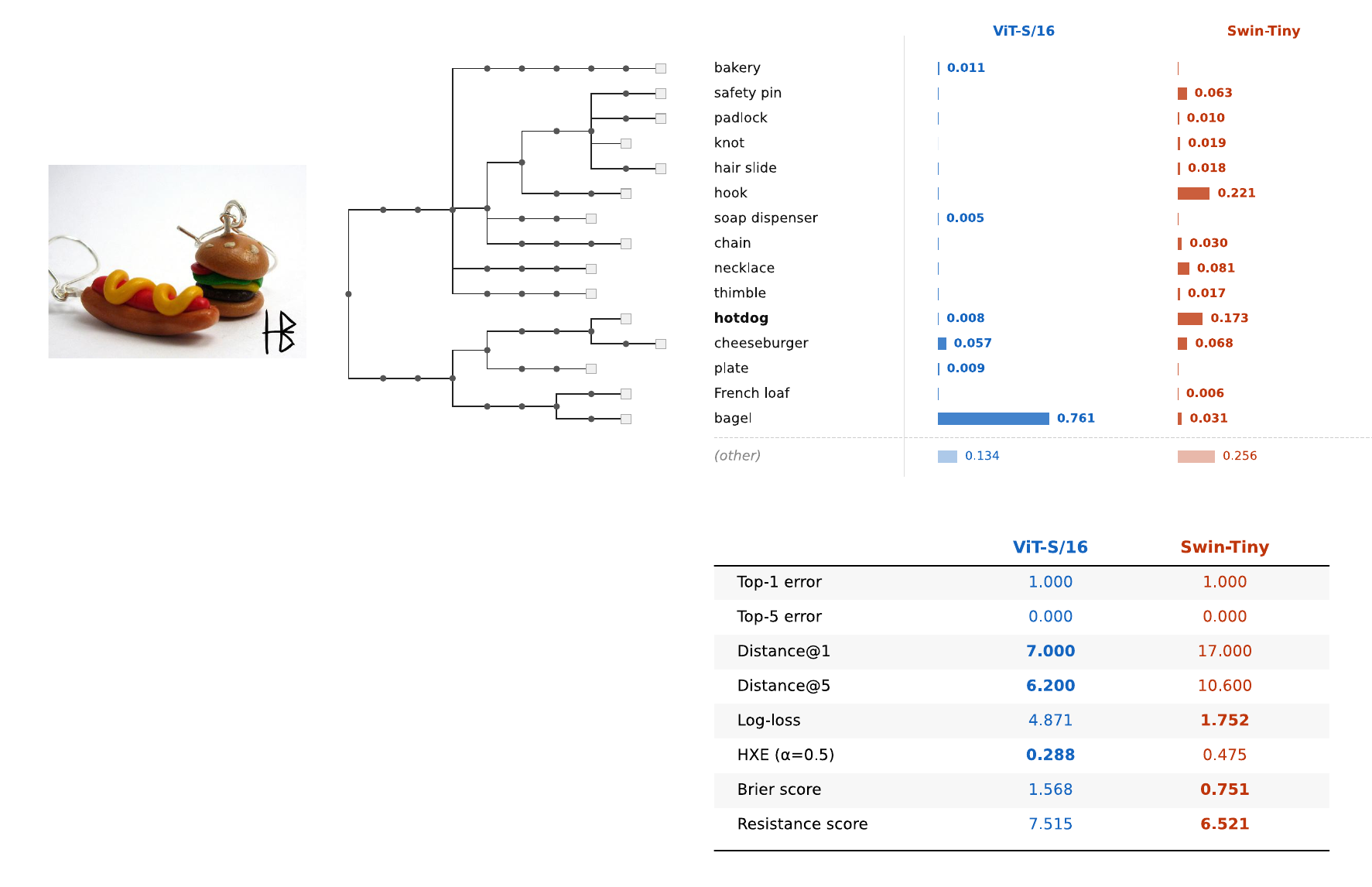}
        \put(0, 0){%
            \raisebox{0pt}[0pt][0pt]{%
                \parbox[b]{0.45\textwidth}{%
                    \captionsetup{type=figure, singlelinecheck=false}
                    \caption{%
                        ViT-S/16 versus Swin-Tiny on image \#22561 (ground truth: \emph{hotdog}). \textbf{Top:} Output distributions of both models on the 15 classes with the highest assigned probability, together with the induced subtree of the WordNet taxonomy. The remaining mass is attributed to `other'. \textbf{Right:} Comparisons of different scores. HXE favours ViT-S/16's confident prediction of the nearby \emph{bagel}, while the Resistance score favours Swin-Tiny for placing more mass on the true class.
                    }
                    \label{fig:dendro_hxe_vs_resistance}
                }%
            }%
        }
    \end{overpic}
\end{figure}

\begin{figure}[htb]
    \includegraphics[width=\textwidth]{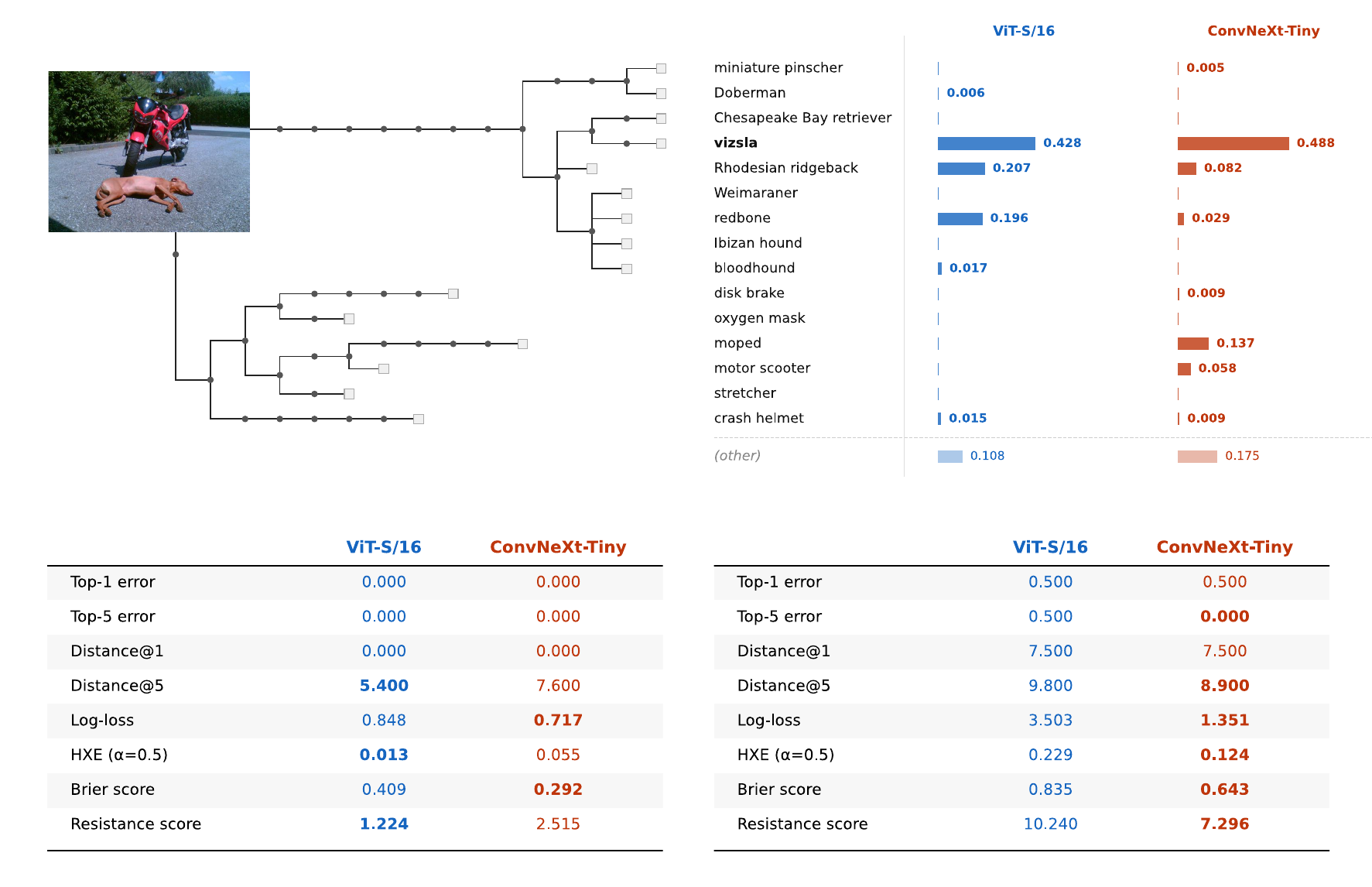}
    \caption{%
        ViT-S/16 versus ConvNeXt-Tiny on image \#44101 (ground truth: \emph{vizsla}). \textbf{Top:} Output distributions of both models on the 15 classes with the highest assigned probability, together with the induced subtree of the WordNet taxonomy. The remaining mass is attributed to `other'. \textbf{Bottom left:} Scores under the ground truth \emph{vizsla}. Hierarchy-aware scores penalise ConvNeXt-Tiny for placing a significant proportion of mass on \emph{moped} and \emph{motor scooter}. \textbf{Bottom right:} Expected scores under an ambiguous ground truth $q_{\mathrm{vizsla}}=q_{\mathrm{moped}}=0.5$. In this case, all scores reward ConvNeXt-Tiny for hedging.
    }
    \label{fig:dendro_ambiguous}
\end{figure}

Figures~\ref{fig:dendro_hedging}--\ref{fig:dendro_ambiguous} show head-to-head comparisons of selected model pairs on three individual images, illustrating characteristic behaviours of the different evaluation metrics. Each figure shows the relevant subtree of the WordNet hierarchy, the probability mass that the models place on the included classes, and a score comparison table for that image.

In Figure~\ref{fig:dendro_hedging} (Image \#16763), ResNet-50 correctly predicts a \emph{tabby} cat, while Swin-Tiny opts for a \emph{tiger cat} as top-1. All classical metrics are either unable to distinguish the two models or prefer ResNet-50. However, Swin-Tiny is actually significantly more confident at predicting a cat in general (assigning about 83\% of the total probability to cats versus 70\% for ResNet-50). As a consequence, the hierarchy-aware scores both prefer Swin-Tiny. This example illustrates that classical proper scores alone are insufficient to detect the hierarchical quality of a distribution.

Figure~\ref{fig:dendro_hxe_vs_resistance} (Image \#22561) shows an example where HXE and the Resistance score disagree. Both ViT-S/16 and Swin-Tiny misclassify the \emph{hotdog} in the picture: ViT-S/16 confidently predicts \emph{bagel} (with probability 0.761), placing only 0.008 probability on the true class, while Swin-Tiny’s top-1 is \emph{hook} (with probability 0.221). HXE favours ViT-S/16, because \emph{bagel} is close to \emph{hotdog} in the taxonomy and its hierarchical weighting rewards a confident prediction near the true class. In contrast, the Resistance score favours Swin-Tiny, because it assigns substantially more probability (0.173) to the true class and otherwise hedges its prediction quite broadly. This illustrates that HXE, like the Diffusion score with large $\tau$, rewards proximity at the cluster level: getting the broad category right matters more than placing mass on the exact true leaf. Whether this is desirable depends on the application.
 
Finally, Figure~\ref{fig:dendro_ambiguous} (Image \#44101) illustrates a potential limitation of hierarchy-aware scores in the standard ImageNet evaluation setting. The image contains a dog of breed \emph{vizsla} (ground-truth label) and a \emph{moped} with roughly equal visual prominence. ViT-S/16 concentrates $0.428$ mass on vizsla and most remaining mass on related hunting dogs, while ConvNeXt-Tiny places $0.488$ on vizsla but $0.137$ on moped and $0.058$ on motor scooter. Under the official ground truth, the hierarchy-aware scores penalise ConvNeXt-Tiny for this hedging. However, under a ground truth distribution of $q_{\mathrm{vizsla}}=q_{\mathrm{moped}}=0.5$, the verdict reverses: ConvNeXt-Tiny's hedging toward moped is now rewarded, outperforming ViT-S/16 in expectation on all scores. We return to the implications of such label ambiguities in the Discussion.

\section{Discussion}
\label{sec:discussion}

\paragraph{Scope and limitations.}
The scoring rules introduced in this paper are designed for \emph{(single) node selection tasks}: settings in which the ground truth is a single node $i \in V$ and the quality of a probabilistic prediction $\vec{p}$ is judged by the proximity of its mass to $i$ in the graph topology. This is the natural setting for source detection and related localisation problems, where graph proximity is directly meaningful: a predicted source one hop away from the true source is a qualitatively different error from one in a far away corner of the network. For hierarchical classification it is the right framework when the taxonomy genuinely encodes semantic proximity and when the ground truth label is unambiguous.

The ImageNet experiments (Section~\ref{sec:imagenet}) illustrate both the strength and an important limitation of our scores. When the uncertainty of the annotator (i.e., the true data distribution $\vec{q}$) is concentrated within a closely connected region of the graph, proper hierarchy-aware scores reward well-calibrated distributional predictions in ways that classical metrics cannot detect. When the true distribution $\vec{q}$ is itself spread across distant parts of the taxonomy (as in the vizsla/moped example of Figure~\ref{fig:dendro_ambiguous}), these scores conflate what are effectively independent prediction problems. For large data sets such as ImageNet with relatively few ambiguous labels, such ambiguities likely average out and the aggregate rankings remain trustworthy. However, if label-ambiguities are more abundant, the problem should be treated as a \emph{multi-label} (or \emph{multi-node selection}) problem instead.

\paragraph{Choice of scoring rule.} Which of the two proposed scoring rules should be used depends both on the graph in question and on the relevant notion of proximity. On small graphs and on sufficiently sparse graphs (in particular trees, which includes the case of hierarchical classification) the Resistance score is the natural choice. It is parameter-free and fully graph-aware. On large and denser graphs, the asymptotic behaviour of the resistance distance raises questions about the sensitivity of the Resistance score to the global graph topology. The Diffusion score provides a more interpretable alternative for this case at the cost of introducing the diffusion time $\tau$ as a free parameter. Moreover, if proximity to the true node matters mainly at the cluster level, the Diffusion score with a suitably large value of $\tau$ is the score of choice. Likewise, in hierarchical classification, HXE may be preferred over the Resistance score when cluster-level accuracy of a classifier is the primary concern.

\paragraph{Extensions and future work.}
Several extensions to more general graphs are of interest. The extension to weighted graphs is immediate: replace the adjacency matrix $A$ with a weighted version $W = (w_{jj'})$, set $L = D_W - W$ where $D_W$ is the diagonal matrix of weighted degrees, and all results carry over without modification. The extension to directed graphs is more subtle. The Laplacian of a directed graph is not symmetric, so its pseudoinverse $L^\dagger$ is not symmetric either, and the effective resistance formula $\omega_{jj'} = L^\dagger_{jj} + L^\dagger_{j'j'} - 2L^\dagger_{jj'}$ no longer defines a symmetric kernel. Symmetrising the Laplacian before taking the pseudoinverse is one workaround, but this is tantamount to discarding all directional information. However, \citet{Young:2016} propose a different generalisation of effective resistance to directed graphs. It reduces to the standard effective resistance on undirected graphs and its square root defines a valid metric on the nodes. Whether this generalisation constitutes the right notion of proximity for a given directed-graph application would require careful examination of the score's properties. We leave this for future work. Two concrete applications motivate this extension: localisation tasks (e.g., source detection) on temporal networks, where the lossless temporal event graph representation of \citet{Mellor:2018} represents a temporal network as a DAG whose nodes are events and whose edges encode temporal precedence; and hierarchical classification on DAG-structured taxonomies (e.g., WordNet), where pruning to a tree (as we do in Section~\ref{sec:imagenet}) discards some structural information that a directed analogue of the Resistance score could exploit.

As hinted at above, a practically important extension is to \emph{multi-node selection}, where the ground truth is a set of nodes rather than a single one. Multi-node selection is subtly different from simple binary node classification as it constitutes a structured prediction problem rather than treating all nodes independently. It has many applications both on general graphs (e.g., multi-source detection, multi-seed influence maximisation, multi-fault detection) as well as in hierarchical classification (e.g., document labelling, protein function prediction, multi-object image classification). Extending strictly proper scoring rules to multi-node selection tasks is a non-trivial open problem, as it would have to deal with distributions on the power set of $V$ rather than $V$ itself.

Strictly proper scoring rules also make natural training losses, since minimising their empirical average encourages the model distribution to match the true data distribution. The Resistance score and Diffusion score can in principle serve as loss functions for node selection problems. The gradient of a kernel score with respect to $\vec{p}$ is straightforward: $\nabla_{\vec{p}} S_k^+(\vec{p}, i) = K(\vec{e}_i - \vec{p})$. For the Resistance and Diffusion score, this is computable via the same Laplacian solver or Krylov subspace methods, respectively, as described in Section~\ref{subsec:efficient_computation}. One practical challenge is that the gradient vanishes in the corners of the simplex. However, this issue could be bypassed using a Fenchel-Young regularisation of the score as described in \citep{Blondel:2020}. A systematic investigation of graph-aware scoring rules as training losses is left for future work.

\section{Conclusion}
\label{sec:conclusion}

In many node selection tasks, proximity in the network matters: a node estimate one hop away from the true node is different from a node estimate in an entirely different region of the network. We argue that evaluation metrics should reflect this distinction and be graph-aware. Yet, as we have shown, the graph-aware metrics in common use, such as Error distance on general graphs and Hierarchical error distance in hierarchical classification, are statistically ill-founded: they are improper scoring rules and therefore produce misleading evaluations.

We have introduced two graph-aware strictly proper scoring rules for probabilistic node selection tasks. The Resistance score is a parameter-free score which reduces to the Brier score on the complete graph and to the RPS on the path graph, thus revealing it as a natural generalisation of these widely used scoring rules to arbitrary graphs. The $\tau$-Diffusion score is a one-parameter family of scores where the diffusion time $\tau$ controls the coarse-graining scale of evaluation, from node-level accuracy at $\tau=0$ to cluster-level accuracy at large $\tau$. Both scores can be computed in nearly linear time on large and reasonably sparse graphs.

Specialising to hierarchical classification on rooted trees, the Resistance score can be seen as both the hierarchy-aware generalisation of the Brier score as well as the strict ``proprification'' of the improper Hierarchical error distance. As such, it is orthogonal to the HXE family of \citet{Bertinetto:2020} and \citet{Urbani:2024} that generalises the logarithmic score instead. We have also shown that the recently established propriety of the HXE family does, in fact, hold strictly.

The generalisation of our scores to weighted graphs is straightforward. Extensions to directed graphs, directed acyclic graphs, temporal networks, multi-node ground truths, and the use of these scores as loss functions for training remain open directions for future work.

\subsubsection*{Code availability}

Code implementing the proposed scoring rules and reproducing all experiments and figures is available at \url{https://github.com/lo-hfk/proper-graph-scoring-rules}.

\subsubsection*{Acknowledgments}

This work was supported by the Swiss National Science Foundation (SNSF), grant number 228527.

\newcommand*{\doi}[1]{\href{https://doi.org/#1}{doi:#1}}
\bibliographystyle{tmlr}
\bibliography{main}

\appendix

\section{RPS as a kernel score}
\label{app:rps}

We show that the Ranked probability score is a kernel score~\eqref{eqn:kernel-score-finite-+} with kernel $k(j,j') = |i-j|$. Indeed, for $j, j' \in \{1,\dots,N\}$, the identity
\[
    |j - j'| = \sum_{k=1}^{N-1} \left(\boldone_{j \leq k} - \boldone_{j' \leq k}\right)^2
\]
holds by direct inspection. Taking $J \sim \vec{p}$ and computing the expectation over $J$ gives
\begin{align*}
    \E_{J \sim \vec{p}}|J - i|
    &= \sum_{k=1}^{N-1} \E_{J \sim \vec{p}}\!\left[\left(\boldone_{J \leq k} - \boldone_{i \leq k}\right)^2\right] \nonumber\\
    &= \sum_{k=1}^{N-1} \left(\Var_{J \sim \vec{p}}[\boldone_{J \leq k}] + \left(F_{\vec{p}}(k) - \boldone_{i \leq k}\right)^2\right) \nonumber\\
    &= \sum_{k=1}^{N-1} F_{\vec{p}}(k)\!\left(1 - F_{\vec{p}}(k)\right) + \mathrm{RPS}(\vec{p}, i),
\end{align*}
where $F_{\vec{p}}(k) = \sum_{j=1}^k p_j$ is the CDF of $\vec{p}$ and we used
$\Var[\boldone_{J \leq k}] = F_{\vec{p}}(k)(1 - F_{\vec{p}}(k))$.
Taking two independent copies $J, J' \sim \vec{p}$ and computing the double expectation gives
\begin{align*}
    \E_{J,J' \sim \vec{p}}|J - J'|
    &= \sum_{k=1}^{N-1} \E_{J,J' \sim \vec{p}}\!\left[\left(\boldone_{J \leq k} - \boldone_{J' \leq k}\right)^2\right] \nonumber\\
    &= 2\sum_{k=1}^{N-1} \Var_{J \sim \vec{p}}[\boldone_{J \leq k}] \nonumber\\
    &= 2\sum_{k=1}^{N-1} F_{\vec{p}}(k)\!\left(1 - F_{\vec{p}}(k)\right).
\end{align*}
Combining the two expressions and solving for $\mathrm{RPS}(\vec{p}, i)$ yields
\[
    \mathrm{RPS}(\vec{p}, i)
    ~=~ \E_{J \sim \vec{p}}|J - i| ~-~ \frac{1}{2}\,\E_{J,J' \sim \vec{p}}|J - J'|,
\]
which is precisely~\eqref{eqn:kernel-score-finite-+} with kernel $k(i,j) = |i-j|$.

\section{Propriety of traditional scores}
\label{app:impropriety_proofs}

\begin{proof}[Proof of Proposition~\ref{prop:rank_based_scores_are_proper}]
    Let $\vec{q},\vec{p}\in\Delta^{N-1}$. Assume w.l.o.g. that the nodes are ordered in such a way that $q_0\leq q_1\leq \dots \leq q_{N-1}$, then $\sigma_{\vec{q}}(N-1) \leq \dots \leq \sigma_{\vec{q}}(1)\leq\sigma_{\vec{q}}(0)$. For the rank score, propriety is a simple consequence of the rearrangement inequality, which says that for increasing sequences $x_0,\dots, x_{N-1}$ and $y_0,\dots, y_{N-1}$ and for all permutations $\sigma\in S^N$, the expression $\sum_{i=0}^{N-1} x_i y_{N-1-i}$ is smaller or equal than $\sum_{i=0}^{N-1} x_i y_{\sigma(i)}$. Specifically, the inequality implies that $\sum_{i=0}^{N-1} q_i \sigma_{\vec{q}}(i) \leq \sum_{i=0}^{N-1} q_i \sigma_{\vec{p}}(i)$, which proves propriety.
    Meanwhile, propriety of the top-$K$ error can be seen as follows:
    \begin{align*}
        \sum_{j=0}^{N-1} q_j S_{\mathrm{top}K}(\vec{q}, j)  &= \sum_{j=0}^{N-1}  q_j\boldone_{\sigma_{\vec{q}}(j)> K} = \sum_{j:\sigma_{\vec{q}}(j)> K} q_j = \sum_{j=0}^{N-K-1} q_j\\
        &\leq \sum_{j:\sigma_{\vec{p}}(j)> K} q_j = \sum_{j=0}^{N-1}  q_j\boldone_{\sigma_{\vec{p}}(j)> K} = \sum_{j=0}^{N-1} q_j S_{\mathrm{top}K}(\vec{p}, j)  
    \end{align*}
\end{proof}

\begin{proof}[Proof of Proposition~\ref{prop:error_distance_is_improper}]
    Consider the tree graph $T=(V,E)$ with vertex set $V=\{r,v,a,b,c\}$, edge set $E=\{(r,v),(v,a),(v,b),(r,c)\}$, and leaf set $\tilde{V}=\{a,b,c\}$. The shortest-path distances between the leaf nodes are $d(a,b)=2$, $d(a,c)=d(b,c)=3$. Let $\delta\in\bigl(0,\tfrac{1}{18}\bigr)$ and define distributions over $V$,
    \be
    \label{eqn:p_q_for_proof_err_dist_is_improper}
        \vec{p}
        = \Bigl(0,0,\tfrac{1}{3},\;\tfrac{1}{3}+\delta,\;\tfrac{1}{3}-\delta\Bigr)
        \qquad\text{and}\qquad
        \vec{q}
        = \Bigl(0,0,\tfrac{1}{3},\;\tfrac{1}{3}-\delta,\;\tfrac{1}{3}+\delta\Bigr),
    \ee
    where entries correspond to $r$, $v$, $a$, $b$ and $c$, respectively (see Figure~\ref{fig:p_q_for_proof_err_dist_is_improper}).

    \begin{figure}[htbp]
    \centering
    \begin{tikzpicture}[
        inode/.style={circle, draw=black, thick,
                      inner sep=2pt, minimum size=18pt, font=\small},
        edge/.style={thick},
        prob/.style={draw=none, fill=none, font=\footnotesize, inner sep=1pt}
    ]
    
    \begin{scope}[xshift=0cm]
      \node[inode] (o_p) at (1.0, 2.8) {$r$};
      \node[inode] (v_p) at (0.0, 1.4) {$v$};
      \node[inode] (a_p) at (-0.8, 0.0) {$a$};
      \node[inode] (b_p) at ( 0.8, 0.0) {$b$};
      \node[inode] (c_p) at ( 2.2, 1.4) {$c$};
      \draw[edge] (o_p)--(v_p);
      \draw[edge] (o_p)--(c_p);
      \draw[edge] (v_p)--(a_p);
      \draw[edge] (v_p)--(b_p);
      \node[prob] at ( 0.45, 2.8)  {$0$};
      \node[prob] at (-0.55, 1.4)  {$0$};
      \node[prob] at (-0.8, -0.55) {$\tfrac{1}{3}$};
      \node[prob] at ( 0.8, -0.55) {$\tfrac{1}{3}+\delta$};
      \node[prob] at ( 2.2,  0.85) {$\tfrac{1}{3}-\delta$};
      \node[draw=none, fill=none, font=\small] at (0.7, 3.6) {$\mathbf{p}$};
    \end{scope}
    
    \begin{scope}[xshift=4.8cm]
      \node[inode] (o_q) at (1.0, 2.8) {$r$};
      \node[inode] (v_q) at (0.0, 1.4) {$v$};
      \node[inode] (a_q) at (-0.8, 0.0) {$a$};
      \node[inode] (b_q) at ( 0.8, 0.0) {$b$};
      \node[inode] (c_q) at ( 2.2, 1.4) {$c$};
      \draw[edge] (o_q)--(v_q);
      \draw[edge] (o_q)--(c_q);
      \draw[edge] (v_q)--(a_q);
      \draw[edge] (v_q)--(b_q);
      \node[prob] at ( 0.45, 2.8)  {$0$};
      \node[prob] at (-0.55, 1.4)  {$0$};
      \node[prob] at (-0.8, -0.55) {$\tfrac{1}{3}$};
      \node[prob] at ( 0.8, -0.55) {$\tfrac{1}{3}-\delta$};
      \node[prob] at ( 2.2,  0.85) {$\tfrac{1}{3}+\delta$};
      \node[draw=none, fill=none, font=\small] at (0.7, 3.6) {$\mathbf{q}$};
    \end{scope}
    \end{tikzpicture}
    
    \caption{Tree $T$ with distributions $\vec{p}$ and $\vec{q}$ as defined in \eqref{eqn:p_q_for_proof_err_dist_is_improper}.}
    \label{fig:p_q_for_proof_err_dist_is_improper}
    \end{figure}
    
    For $\delta>0$ we have $m_{\vec{p}}=b$ and $m_{\vec{q}}=c$. Therefore,
    \begin{align*}
        \E_{i\sim\vec{q}}\,S_{\mathrm{ed}}(\vec{p},i)
        &= \tfrac{1}{3}\cdot d(b,a)
        + \bigl(\tfrac{1}{3}-\delta\bigr)\cdot d(b,b)
        + \bigl(\tfrac{1}{3}+\delta\bigr)\cdot d(b,c)
        = \tfrac{1}{3}\cdot 2 + 0 + \bigl(\tfrac{1}{3}+\delta\bigr)\cdot 3
        = \tfrac{5}{3}+3\delta
    \end{align*}
    and
    \begin{align*}
        \E_{i\sim\vec{q}}\,S_{\mathrm{ed}}(\vec{q},i)
        &= \tfrac{1}{3}\cdot d(c,a)
        + \bigl(\tfrac{1}{3}-\delta\bigr)\cdot d(c,b)
        + \bigl(\tfrac{1}{3}+\delta\bigr)\cdot d(c,c)
        = \tfrac{1}{3}\cdot 3 + \bigl(\tfrac{1}{3}-\delta\bigr)\cdot 3 + 0
        = 2-3\delta \, .
    \end{align*}
    Since $\delta<\tfrac{1}{18}$ implies $6\delta<\tfrac{1}{3}$, we have $\tfrac{5}{3}+3\delta < 2-3\delta$, so $\E_{i\sim\vec{q}}S_{\mathrm{ed}}(\vec{p},i) < \E_{i\sim\vec{q}}S_{\mathrm{ed}}(\vec{q},i)$ and the Error distance is improper.
\end{proof}

\begin{proof}[Proof of Proposition~\ref{prop:integrated_error_distance_is_improper}]
    We use the same tree $T$, leaf nodes, distributions $\vec{p}$, $\vec{q}$, and $\delta\in\bigl(0,\tfrac{1}{18}\bigr)$ as above. The matrix of shortest-path distances between leaf nodes is
    \[
        D_{\mathrm{sp}}
        =
        \begin{bmatrix}
            0 & 2 & 3 \\
            2 & 0 & 3 \\
            3 & 3 & 0
        \end{bmatrix},
    \]
    with rows and columns indexed by $a$, $b$, $c$. Denote by $\tilde{\vec{p}}$ and $\tilde{\vec{q}}$ the three-dimensional restrictions of $\vec{p}$ and $\vec{q}$ to the leaf nodes. Note that $\E_{i\sim\vec{q}}S_{\mathrm{aed}}(\vec{p},i) = \tilde{\vec{q}}^\top D_{\mathrm{sp}}\tilde{\vec{p}}$ and $\E_{i\sim\vec{q}}S_{\mathrm{aed}}(\vec{q},i) = \tilde{\vec{q}}^\top D_{\mathrm{sp}}\tilde{\vec{q}}$. A direct computation yields the difference
    \[
        \tilde{\vec{q}}^\top D_{\mathrm{sp}}(\tilde{\vec{q}}-\tilde{\vec{p}})
        =
        \tilde{\vec{q}}^\top D_{\mathrm{sp}}
        \begin{pmatrix}0\\-2\delta\\2\delta\end{pmatrix}
        \;=\;
        2\delta\Bigl(\tfrac{1}{3}-6\delta\Bigr).
    \]
    For $\delta<\tfrac{1}{18}$ we have $\tfrac{1}{3}-6\delta>0$, so $\tilde{\vec{q}}^\top D_{\mathrm{sp}}(\tilde{\vec{q}}-\tilde{\vec{p}})>0$, giving $\E_{i\sim\vec{q}}S_{\mathrm{aed}}(\vec{p},i) < \E_{i\sim\vec{q}}S_{\mathrm{aed}}(\vec{q},i)$. Hence, the average Error distance is improper.
\end{proof}

\begin{proof}[Proof of Proposition~\ref{prop:lca_height_is_improper}]
    We use the same tree $T$ and distributions $\vec{p}$, $\vec{q}$ which depend on $\delta$ as above. This time, let $\delta\in (0,\frac{1}{12})$. Furthermore, we explicitly mark $r$ as the root node. The lowest common ancestors are then given by $\mathrm{LCA}(a,b)=v$, $\mathrm{LCA}(a,c)=\mathrm{LCA}(b,c)=r$, and $\mathrm{LCA}(x,x)=x$ for any leaf $x\in \tilde{V}$. 
    Recall that the LCA-height of two nodes is the height of their lowest common ancestor, where the height of a node $u$ is the length of the longest downward path from $u$ to any leaf in its subtree (so all leaves have height $0$). In $T$, the node $v$ has height $1$ and the root $r$ has height $2$. Given that $m_{\vec{p}}=b$ and $m_{\vec{q}}=c$, we find
    \begin{align*}
        \E_{i\sim\vec{q}}\,S_{\mathrm{LCA}}(\vec{p},i)
        &= \tfrac{1}{3}\cdot 1
        + \bigl(\tfrac{1}{3}-\delta\bigr)\cdot 0
        + \bigl(\tfrac{1}{3}+\delta\bigr)\cdot 2
        = 1+2\delta
    \end{align*}
    and
    \begin{align*}
        \E_{i\sim\vec{q}}\,S_{\mathrm{LCA}}(\vec{q},i)
        &= \tfrac{1}{3}\cdot 2
        + \bigl(\tfrac{1}{3}-\delta\bigr)\cdot 2
        + \bigl(\tfrac{1}{3}+\delta\bigr)\cdot 0
        = \tfrac{4}{3}-2\delta.
    \end{align*}
    Since $\delta<\tfrac{1}{12}$ implies  $1+2\delta < \tfrac{4}{3}-2\delta$, we have $\E_{i\sim\vec{q}}S_{\mathrm{LCA}}(\vec{p},i) < \E_{i\sim\vec{q}}S_{\mathrm{LCA}}(\vec{q},i)$ and thus $S_{\mathrm{LCA}}$ is improper.
\end{proof}

\section{Source detection: supplementary information}
\label{app:source_detection}

This section contains supplementary information and additional results for the example presented in Section~\ref{sec:source_detection}.

\subsection{Network and observation state selection}
\label{app:source-detection:network}

The Karate Club network ($N=34$) serves as the basis for this example because it sits at a convenient middle ground: large enough to produce non-trivial outbreak patterns, yet small enough that the true source distribution is computationally accessible via Monte Carlo enumeration for at least a subset of observation states.

From every node in the network, we simulate $10^7$ SI outbreaks ($3.4 \times 10^8$ simulations in total) with transmission rate $\beta = 0.03$ and a fixed observation horizon of $T=25$ time steps, yielding an average outbreak size of $11.5$ of nodes.

\begin{figure}[p]
  \centering
  \begin{tabular}{cc}
    \includegraphics[width=0.5\linewidth]{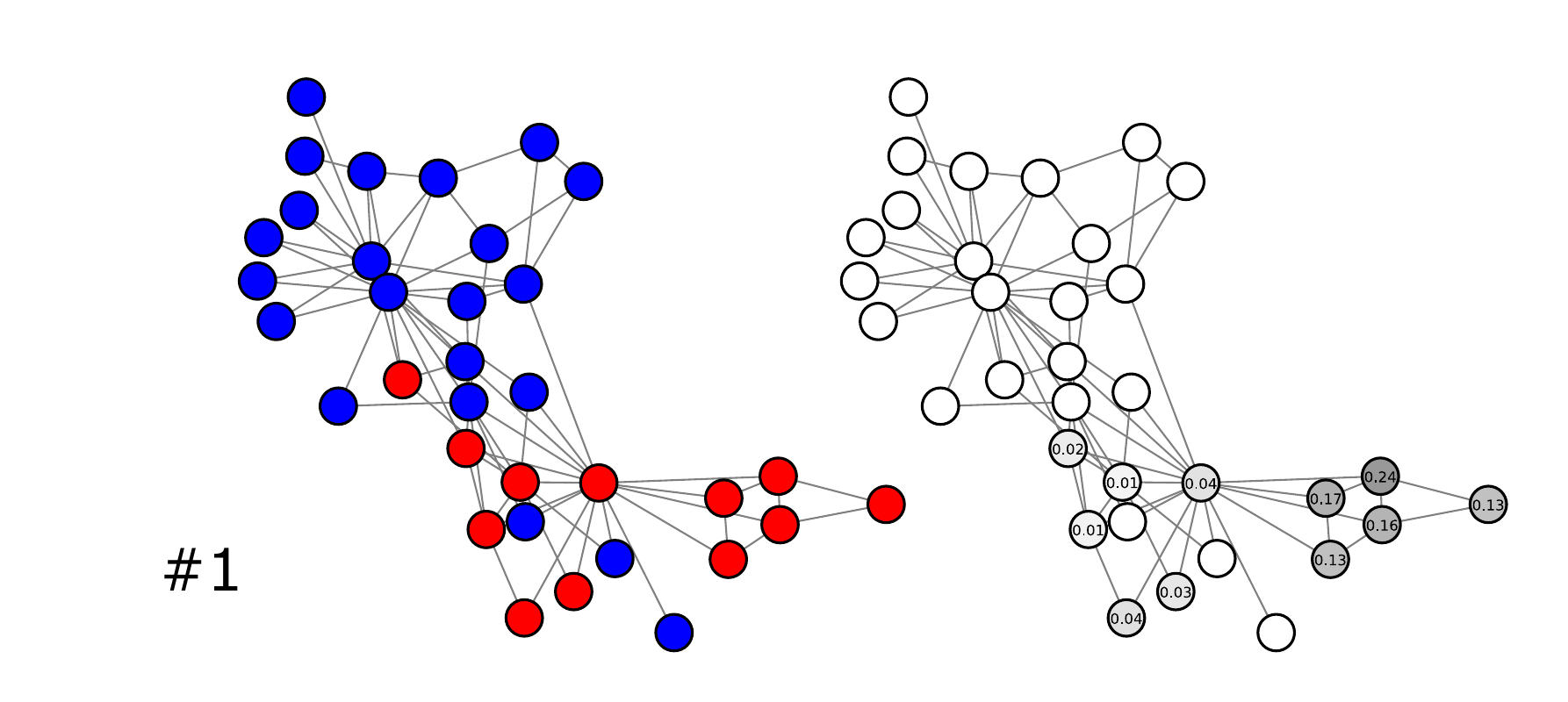} &
    \includegraphics[width=0.5\linewidth]{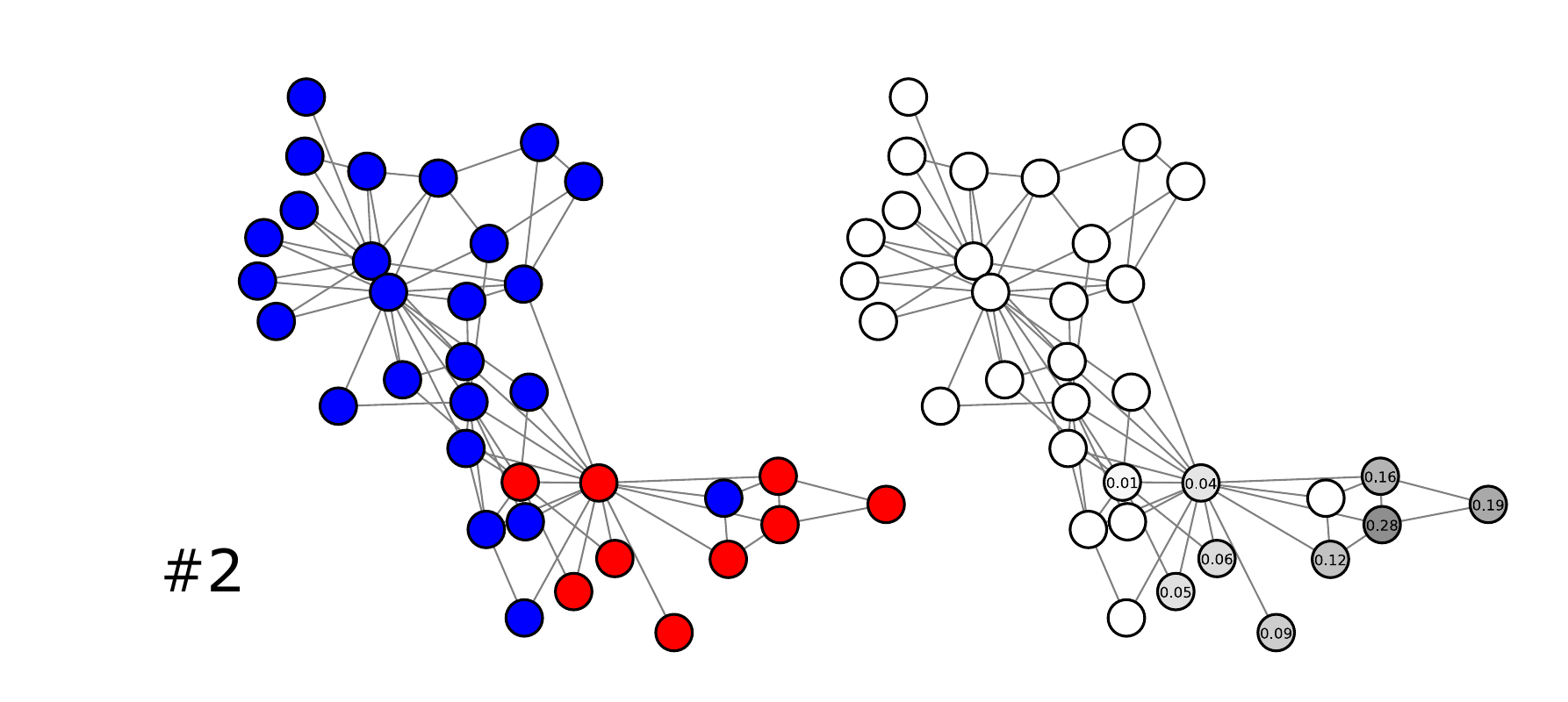} \\
    \includegraphics[width=0.5\linewidth]{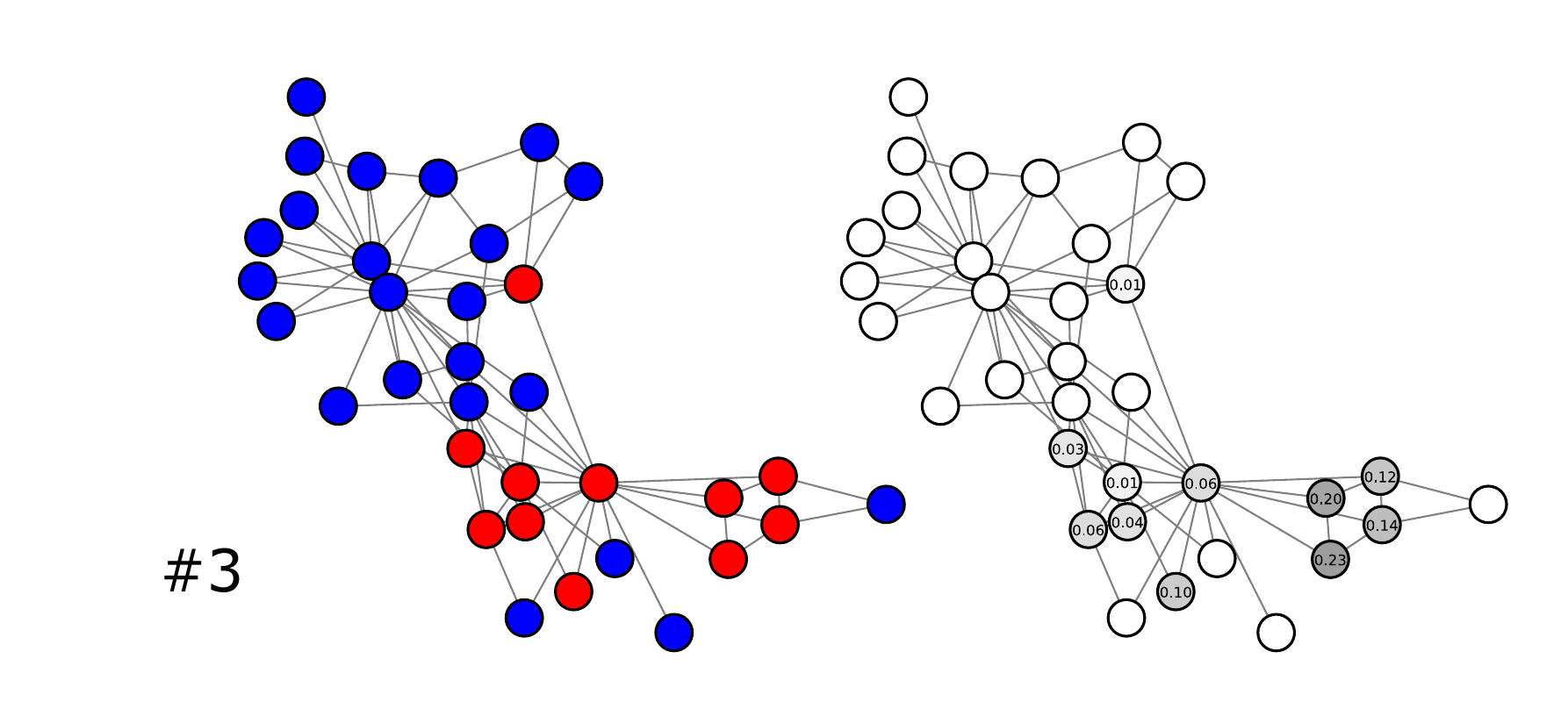} &
    \includegraphics[width=0.5\linewidth]{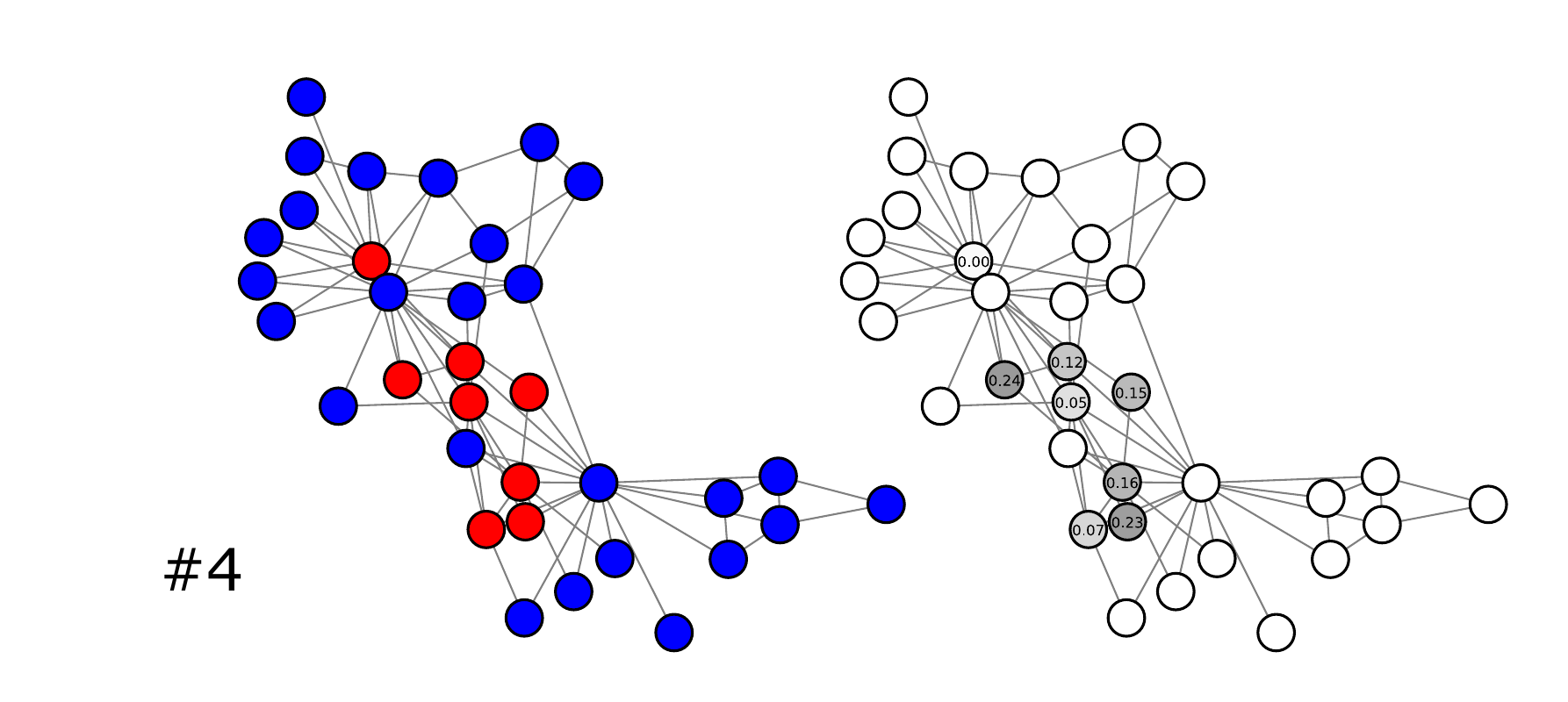} \\
    \includegraphics[width=0.5\linewidth]{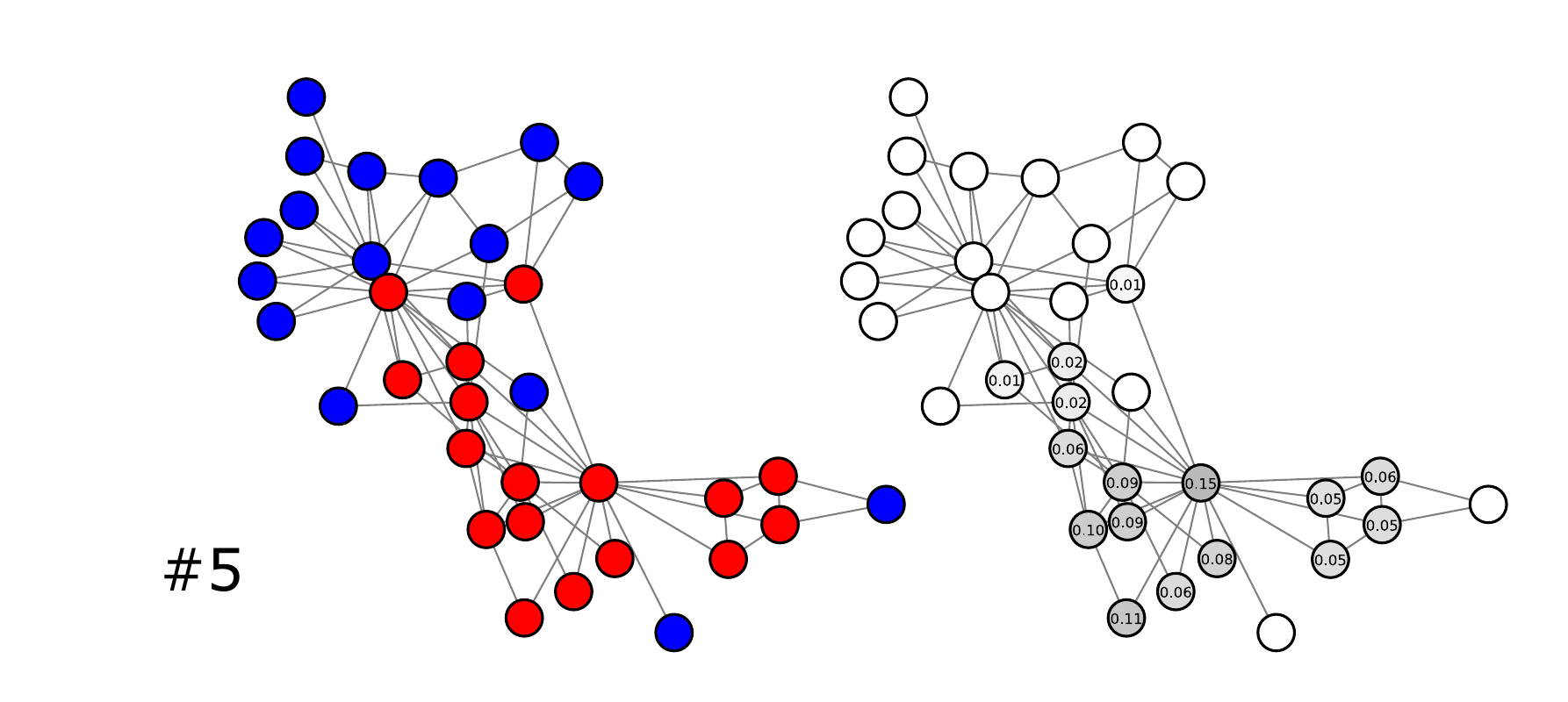} &
    \includegraphics[width=0.5\linewidth]{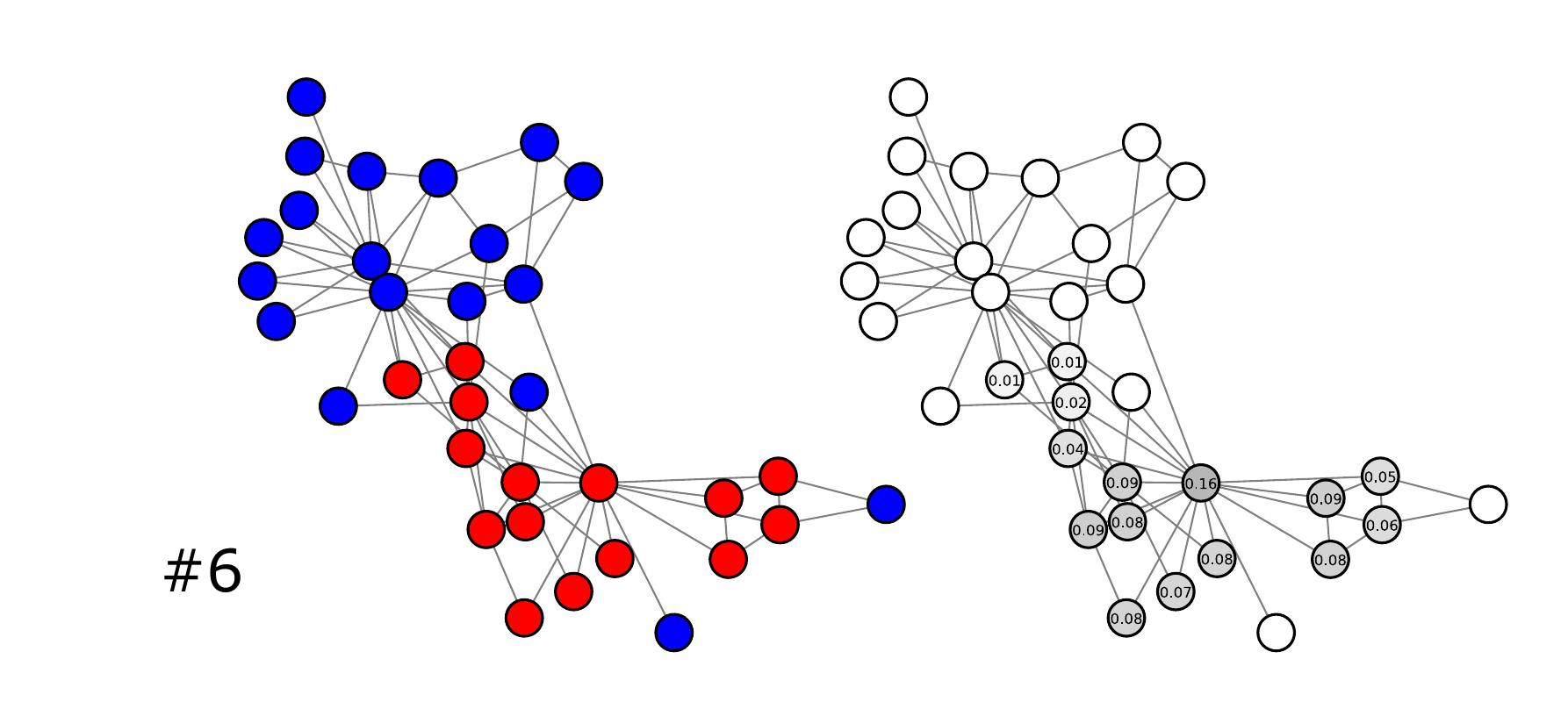} \\
    \includegraphics[width=0.5\linewidth]{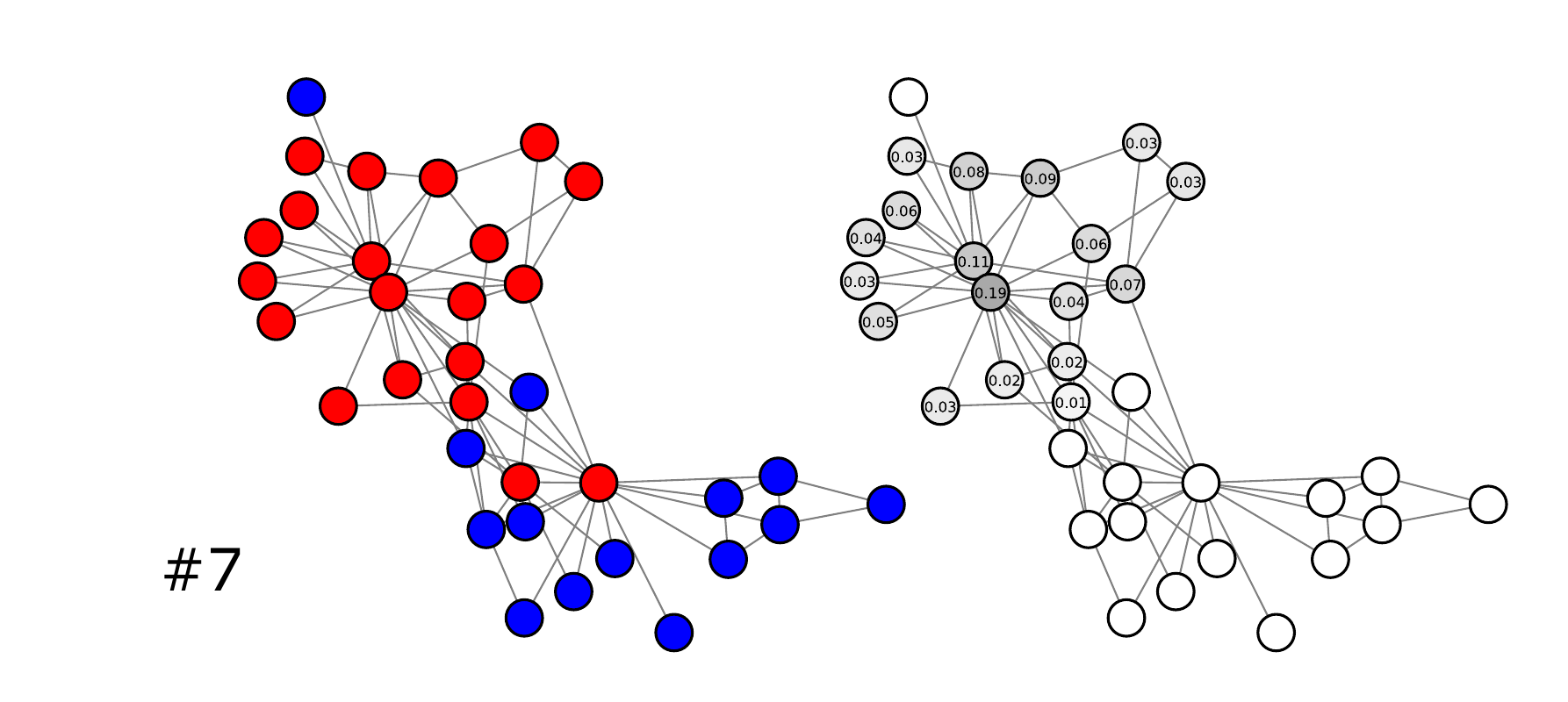} &
    \includegraphics[width=0.5\linewidth]{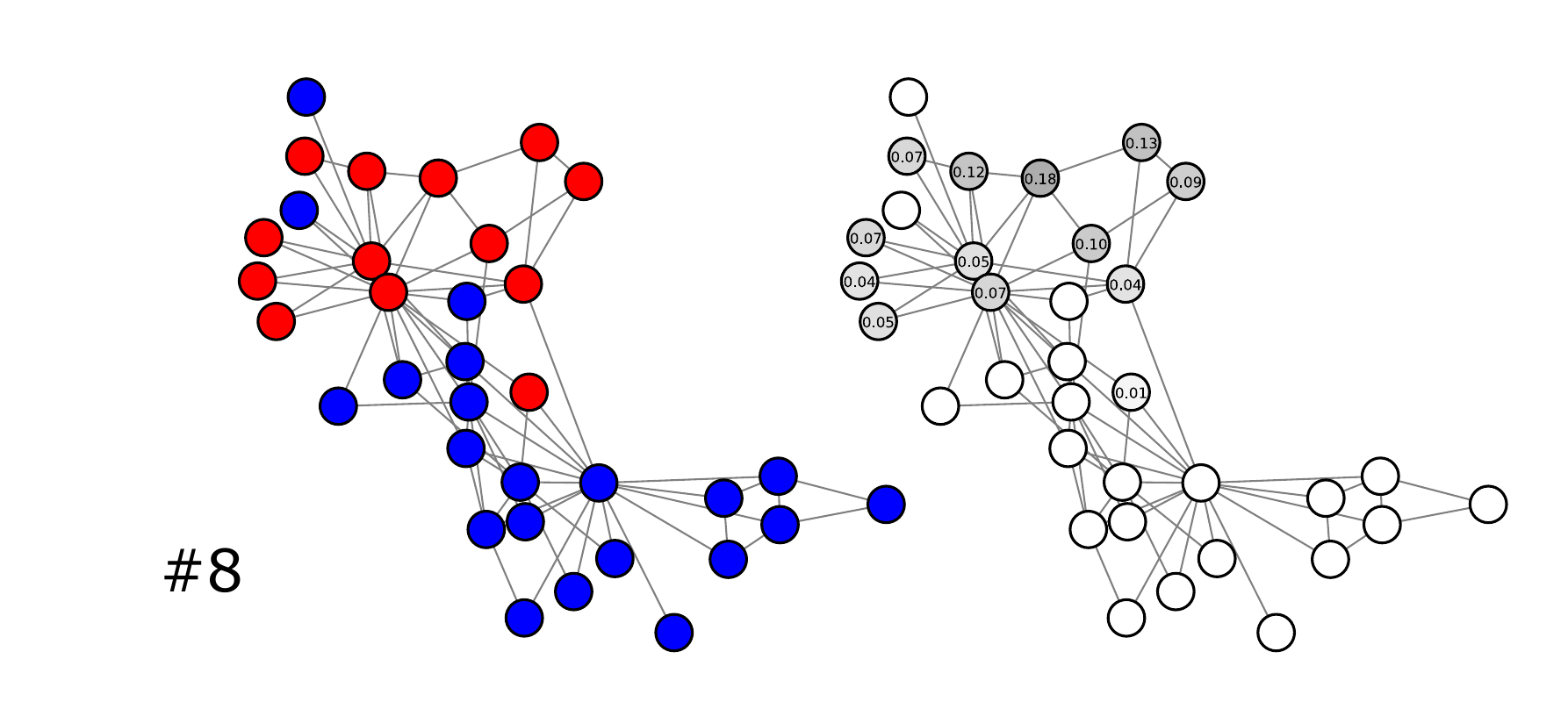} \\
    \includegraphics[width=0.5\linewidth]{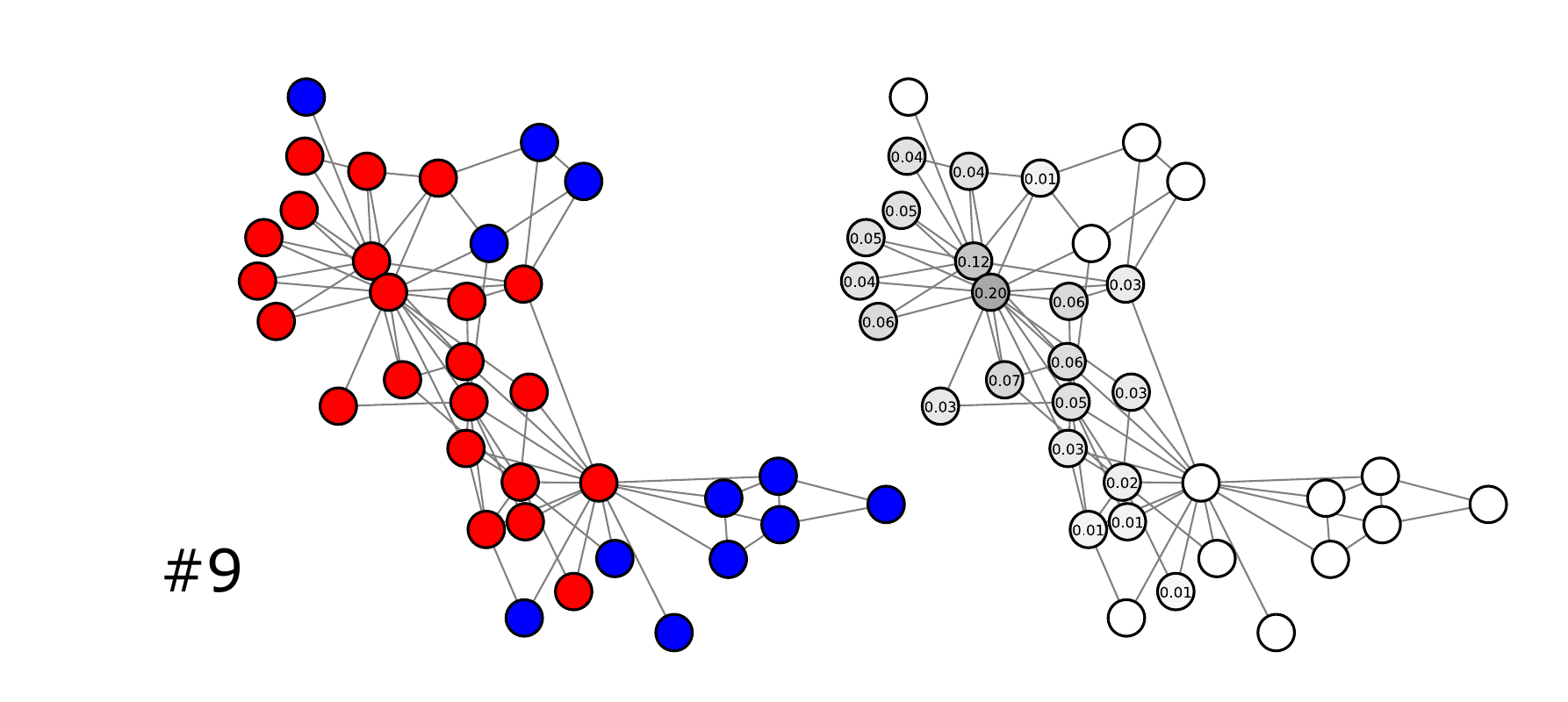} &
    \includegraphics[width=0.5\linewidth]{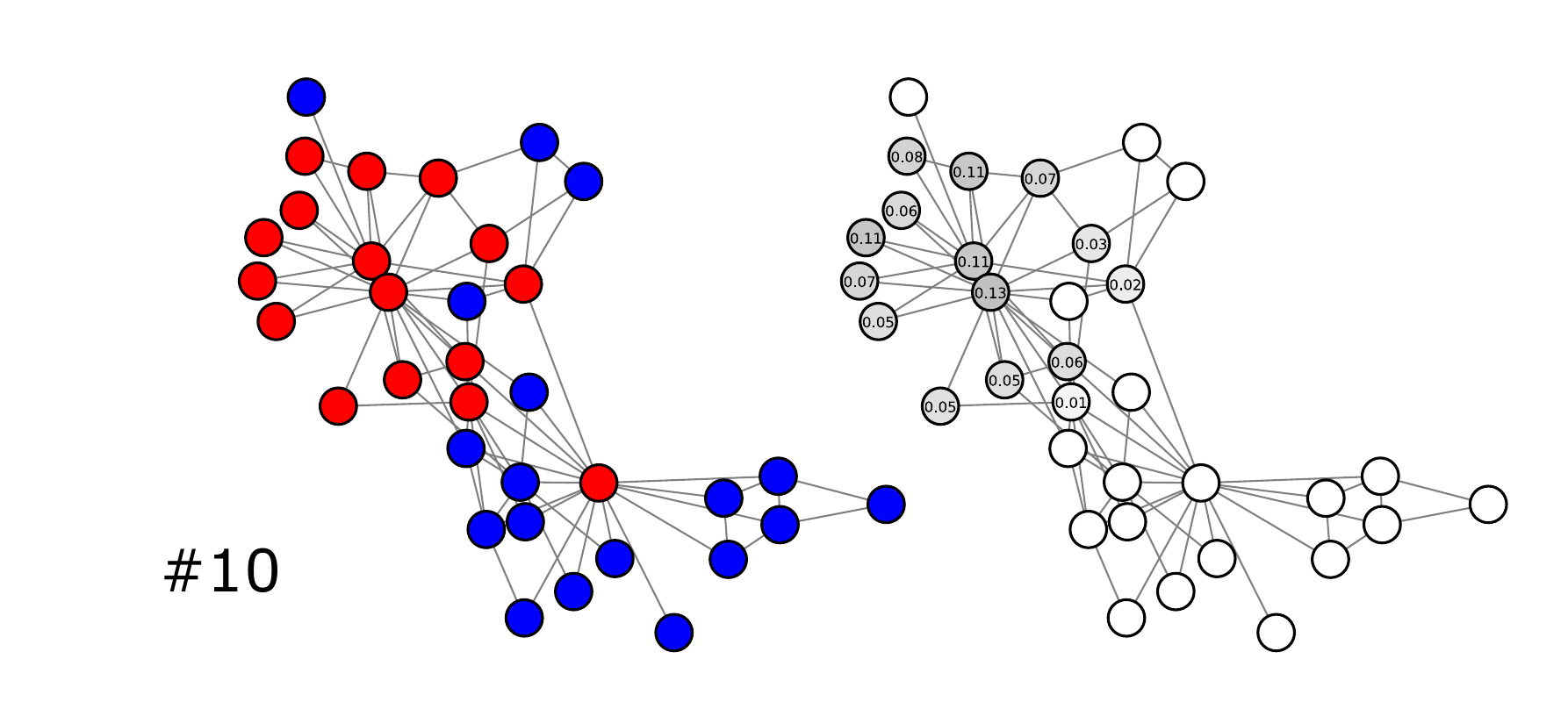} \\
  \end{tabular}
  \caption{All ten observed states together with their true source distribution. Red nodes are infected, blue nodes susceptible.}
  \label{fig:app:states}
\end{figure}

From among the $2^{34}$ possible observation states at time $T$, we select ten states $\vec{O}_T$ satisfying two criteria: (a) sufficiently many simulations resulted in $\vec{O}_T$ for the true distribution $\vec{q}$ to be estimated with reasonable precision, and (b) the true distribution is neither too concentrated nor too uniform. For each selected $\vec{O}_T$, the true distribution is estimated as the empirical distribution over all simulated outbreaks that resulted in that particular state. Figure \ref{fig:app:states} shows the ten selected observed states together with their true source distributions $\vec{q}$. The example shown in Section \ref{sec:source_detection} is state \#2.

\subsection{Estimators}
\label{app:source-detection:estimators}

All estimators are trained on balanced sets of simulated outbreaks with $n$ outbreaks per source node, $n \in \{10, 100, 1{,}000, 10{,}000\}$ ($34n$ outbreaks in total).

\paragraph{GNN.}
The GNN estimator is a variant of the best-performing graph neural network architecture identified in the benchmark study of \citet{Sterchi:2026}. It takes the binary infection-status vector and the graph structure as input and outputs a probability distribution over source nodes. We train separate models for training-set sizes $n \in \{10, 100, 1{,}000, 10{,}000\}$.

\paragraph{MCMF.}
The Monte Carlo Mean-Field (MCMF) estimator of \citet{Sterchi:2023} estimates the posterior source distribution by approximating the likelihood of the observed infection state given each candidate source under a node-independence assumption, with marginal node-state probabilities estimated by Monte Carlo simulation. Here, $n$ refers to the number of simulations per candidate source node. 

\paragraph{SME.}
The Soft-Margin Estimator (SME) of \citet{Antulov:2015} is another Monte Carlo method based on Approximate Bayesian computation (ABC) with a soft acceptance criterium, whose tolerance is determined by a parameter $a$. The original paper prescribes a method to fit $a$ to the data. We instead evaluate the method at five fixed values $a \in \{0.1, 0.15, 0.2, 0.3, 0.4\}$.

\subsection{Selected states and score tables}
\label{app:source-detection:results}

\begin{figure}[p]
  \centering
  \includegraphics[width=0.9\linewidth]{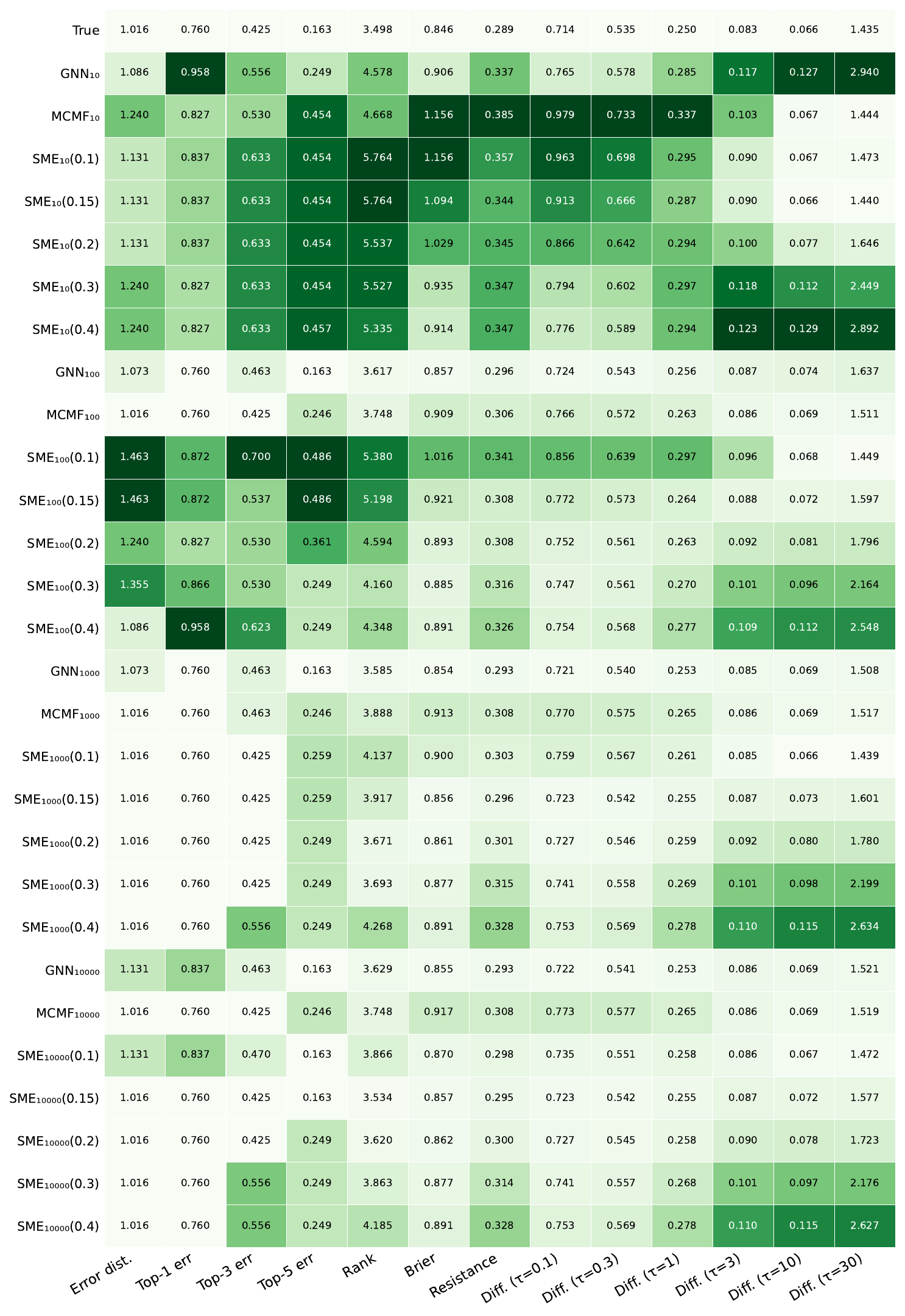}
  \caption{Observed state \#1 -- expected scores under the true distribution for all estimators (darker: worse).}
  \label{fig:app:state1}
\end{figure}

\begin{figure}[p]
  \centering
  \includegraphics[width=0.9\linewidth]{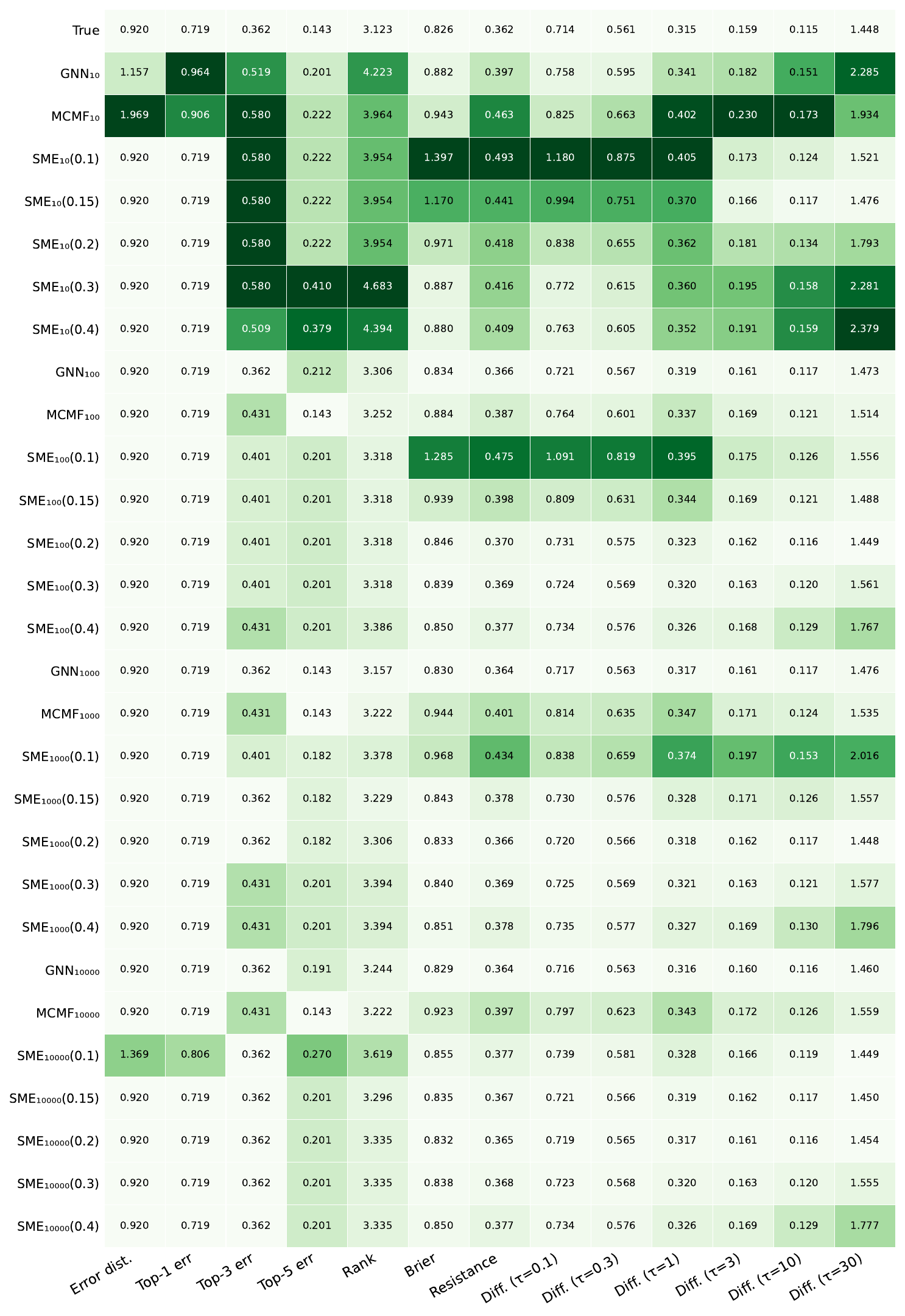}
  \caption{Observed state \#2 -- expected scores under the true distribution for all estimators (darker: worse).}
  \label{fig:app:state2}
\end{figure}

\begin{figure}[p]
  \centering
  \includegraphics[width=0.9\linewidth]{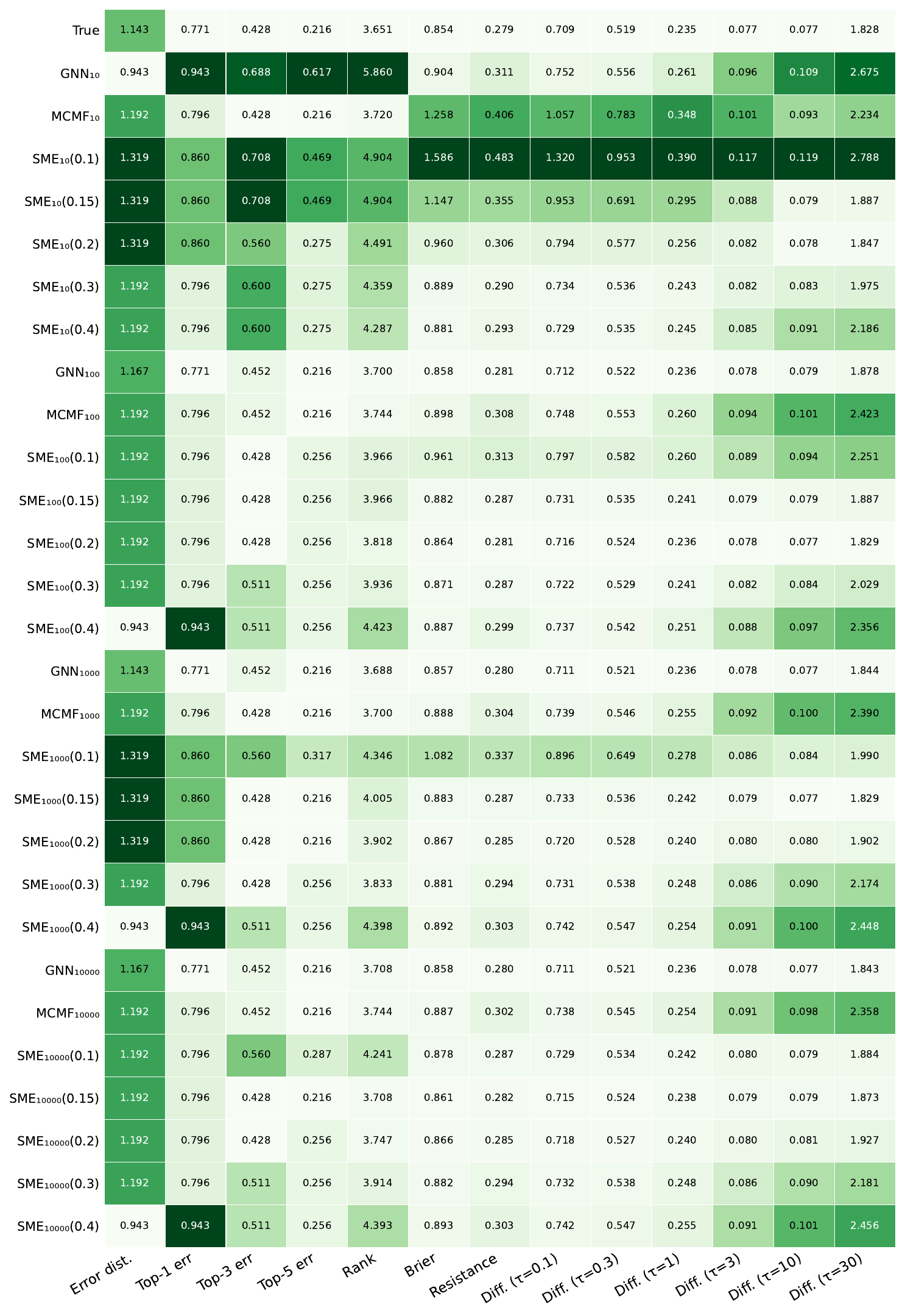}
  \caption{Observed state \#3 -- expected scores under the true distribution for all estimators (darker: worse).}
  \label{fig:app:state3}
\end{figure}

\begin{figure}[p]
  \centering
  \includegraphics[width=0.9\linewidth]{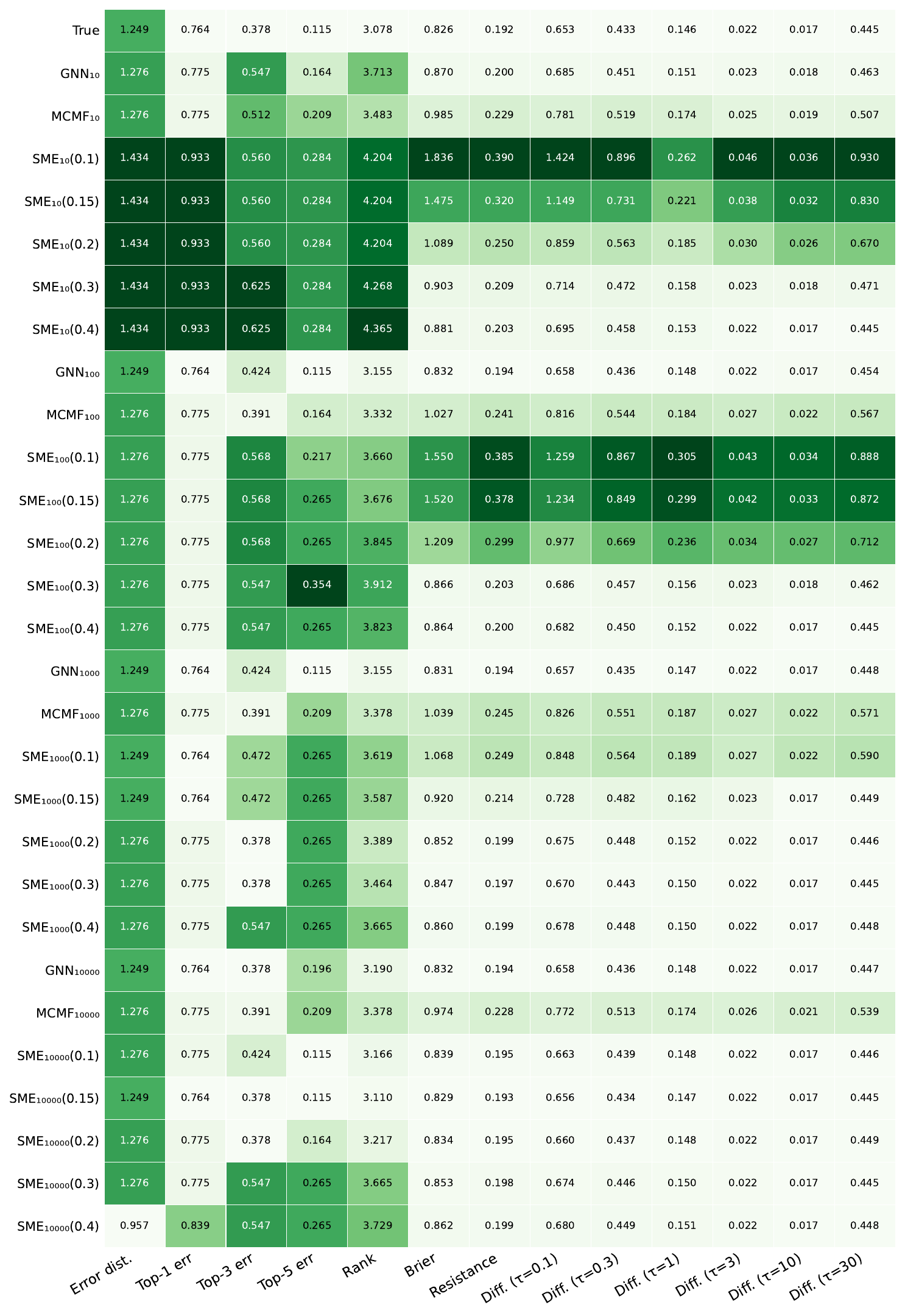}
  \caption{Observed state \#4 -- expected scores under the true distribution for all estimators (darker: worse).}
  \label{fig:app:state4}
\end{figure}

\begin{figure}[p]
  \centering
  \includegraphics[width=0.9\linewidth]{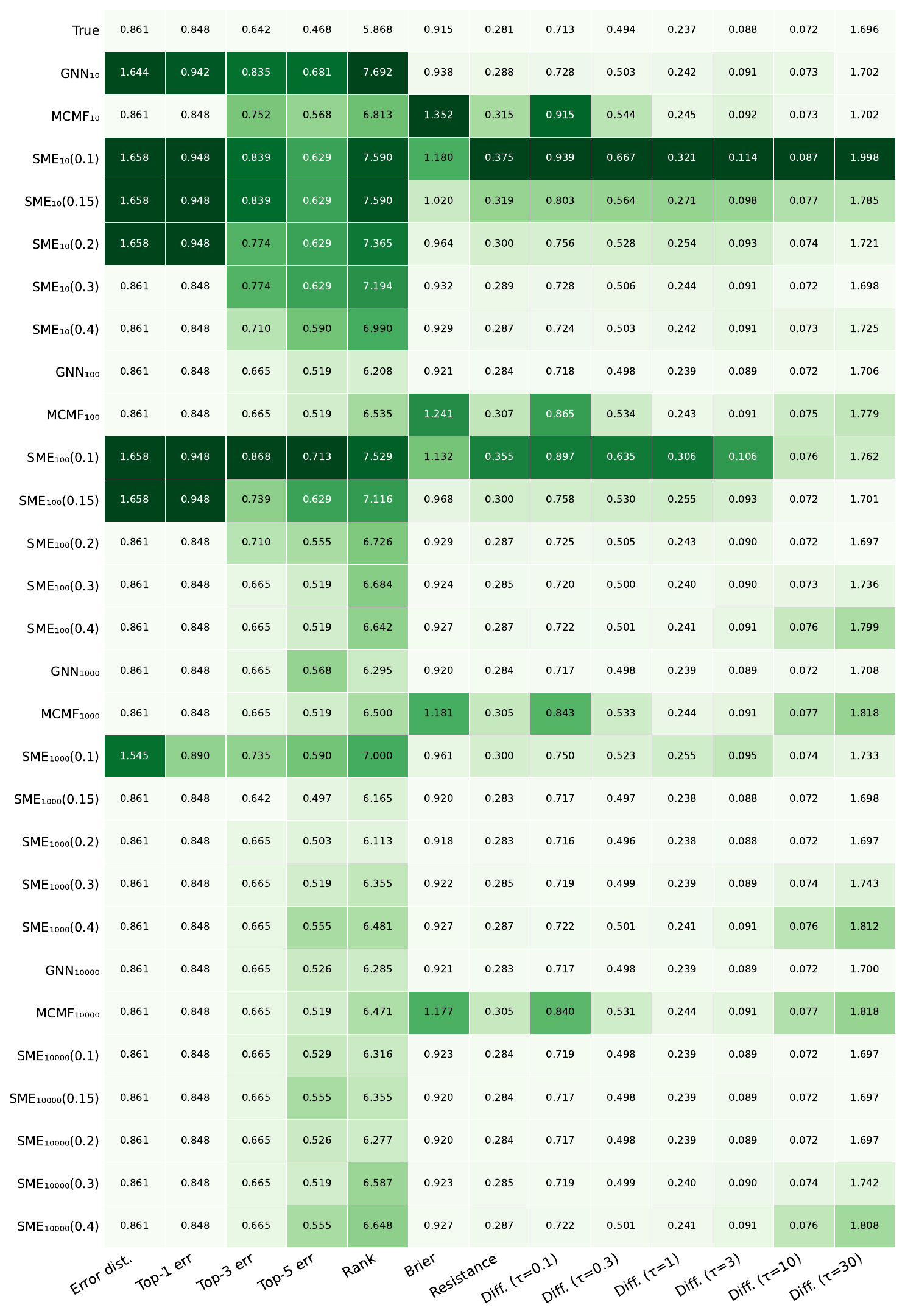}
  \caption{Observed state \#5 -- expected scores under the true distribution for all estimators (darker: worse).}
  \label{fig:app:state5}
\end{figure}

\begin{figure}[p]
  \centering
  \includegraphics[width=0.9\linewidth]{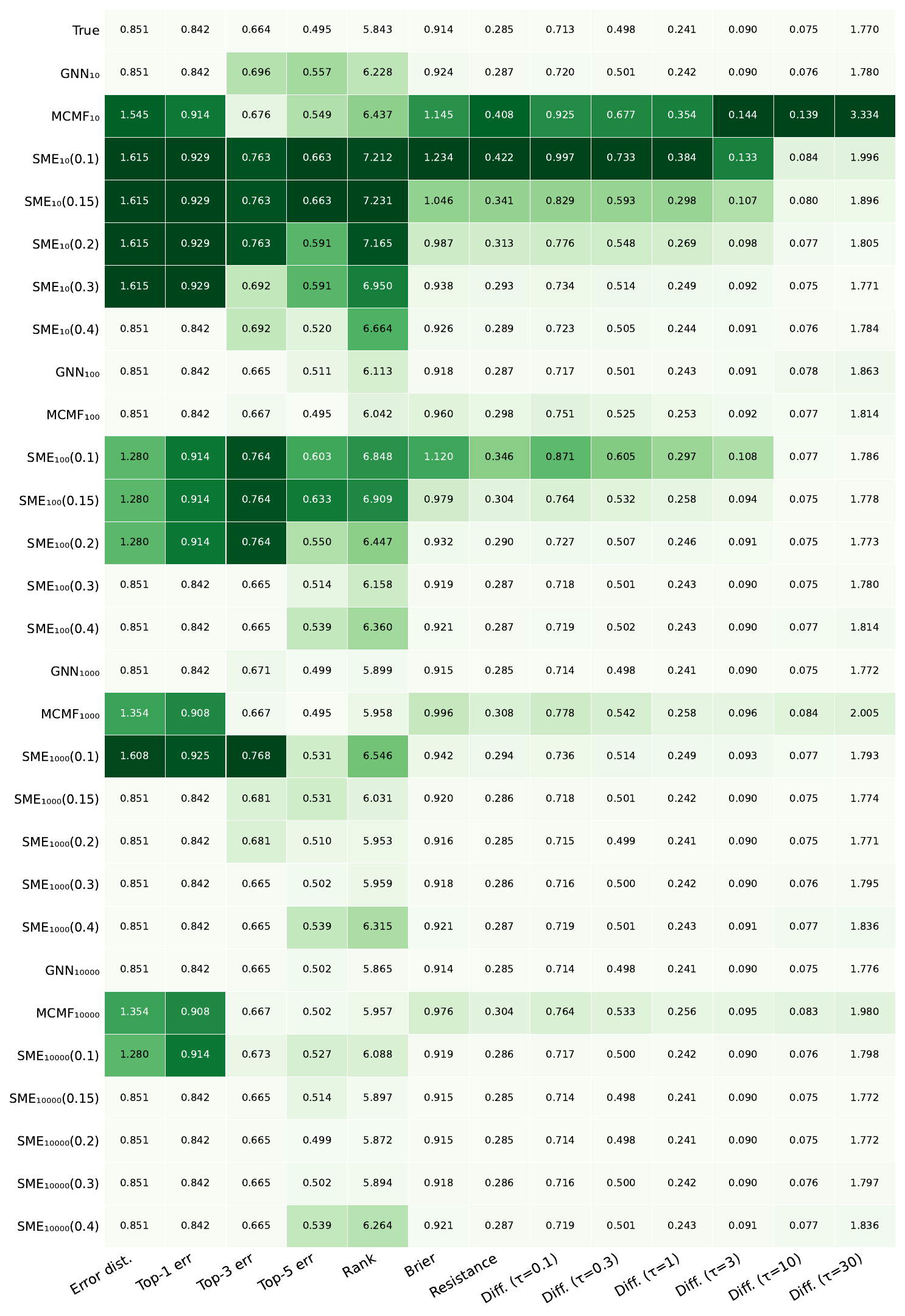}
  \caption{Observed state \#6 -- expected scores under the true distribution for all estimators (darker: worse).}
  \label{fig:app:state6}
\end{figure}

\begin{figure}[p]
  \centering
  \includegraphics[width=0.9\linewidth]{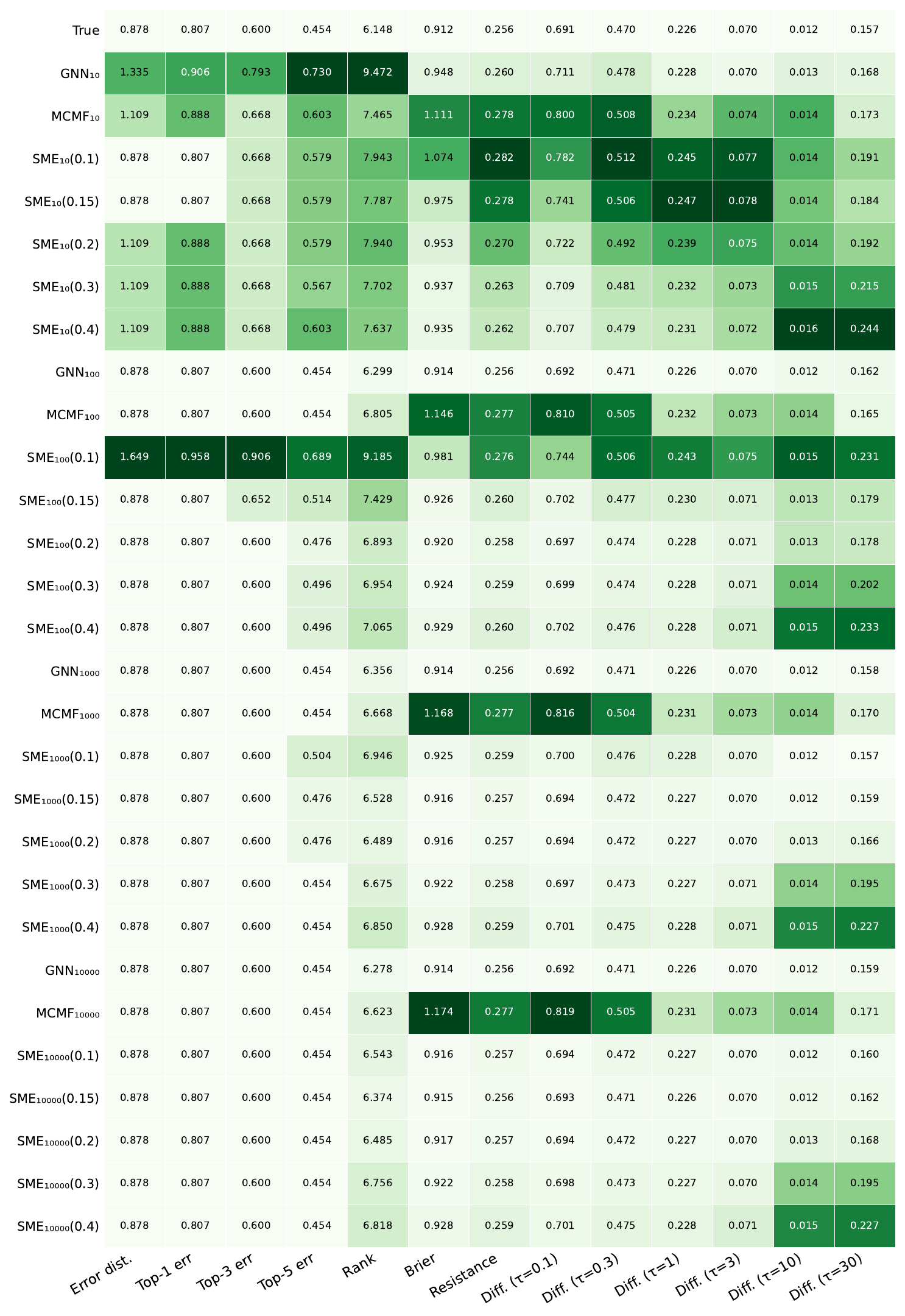}
  \caption{Observed state \#7 -- expected scores under the true distribution for all estimators (darker: worse).}
  \label{fig:app:state7}
\end{figure}

\begin{figure}[p]
  \centering
  \includegraphics[width=0.9\linewidth]{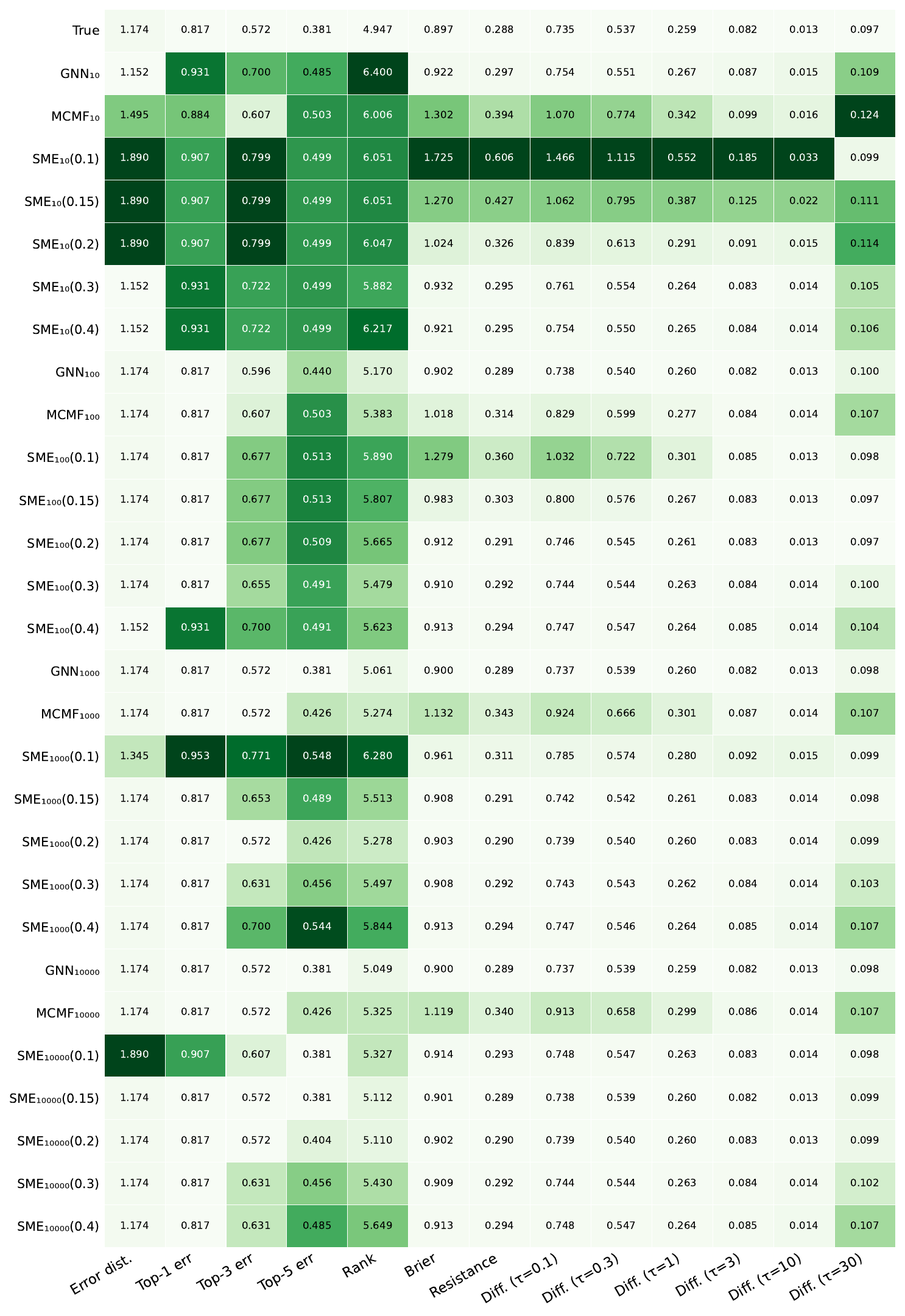}
  \caption{Observed state \#8 -- expected scores under the true distribution for all estimators (darker: worse).}
  \label{fig:app:state8}
\end{figure}

\begin{figure}[p]
  \centering
  \includegraphics[width=0.9\linewidth]{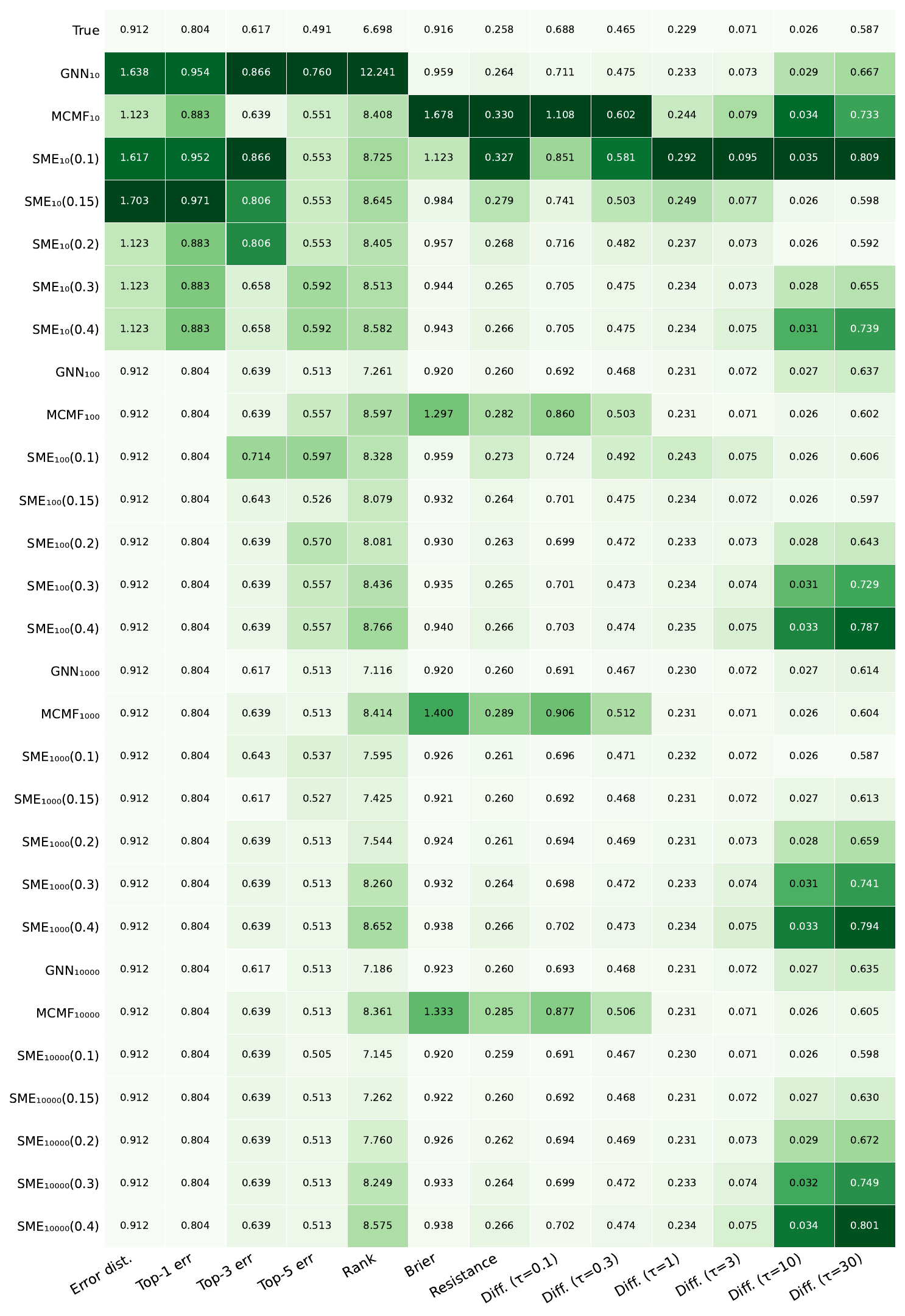}
  \caption{Observed state \#9 -- expected scores under the true distribution for all estimators (darker: worse).}
  \label{fig:app:state9}
\end{figure}

\begin{figure}[p]
  \centering
  \includegraphics[width=0.9\linewidth]{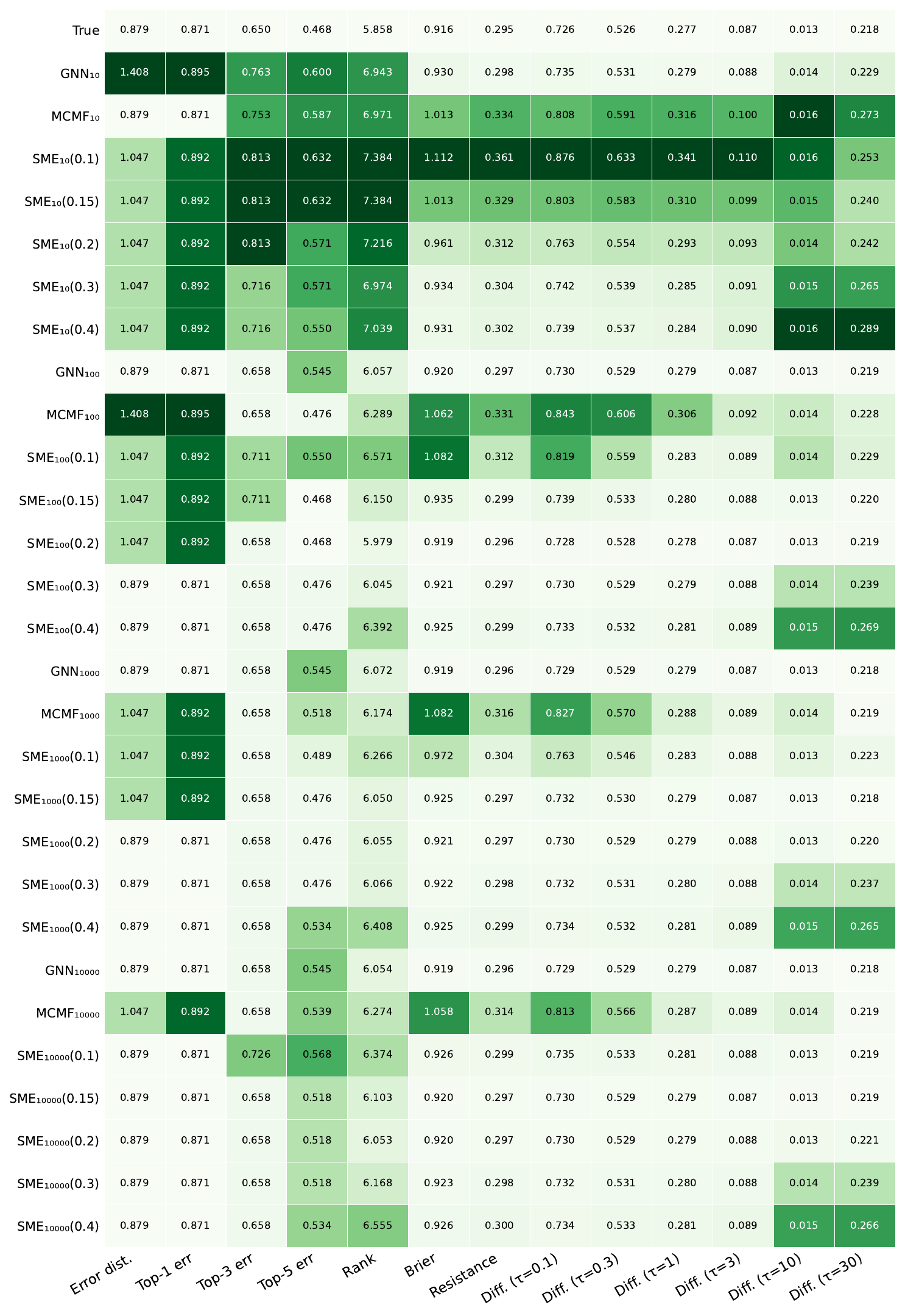}
  \caption{Observed state \#10 -- expected scores under the true distribution for all estimators (darker: worse).}
  \label{fig:app:state10}
\end{figure}

Figures~\ref{fig:app:state1}--\ref{fig:app:state10} show, for each of the ten selected observed states, the network coloured by infection status alongside the empirical true-source distribution, and the full score table comparing all estimators. The observed state and models in the main text are selected to jointly illustrate several interesting features.

\end{document}